\documentclass[twoside,11pt]{article}

\usepackage{blindtext}

\usepackage{jmlr2e}

\usepackage[utf8]{inputenc} % allow utf-8 input
\usepackage[T1]{fontenc}    % use 8-bit T1 fonts
\usepackage{hyperref}       % hyperlinks
\hypersetup{
    pdftitle={NEVIS'22},
}

\usepackage{url}            % simple URL typesetting
\usepackage{booktabs}       % professional-quality tables
\usepackage{amsfonts}       % blackboard math symbols
\usepackage{nicefrac}       % compact symbols for 1/2, etc.
\usepackage{microtype}      % microtypography
\usepackage{xcolor}         % colors
\usepackage{graphicx}
\usepackage{caption}
\usepackage{subcaption}
\usepackage[fleqn]{amsmath}
\usepackage{algorithm}
\usepackage{algpseudocode}
\usepackage{multirow}
\usepackage{soul}
\usepackage[normalem]{ulem}
\usepackage{longtable}
\usepackage{wrapfig}
\usepackage[export]{adjustbox}

\newcommand{\minerva}{\textsc{Nevis'22}}
\newcommand{\mr}[1]{#1}%{{\bf MR:}#1}}
\newcommand{\yutian}[1]{#1}%{{\bf Yutian: }#1}}
\newcommand{\jb}[1]{#1}%{JB: #1}}
\newcommand{\ag}[1]{#1}%{{\bf Alex: }#1}}
\newcommand{\art}[1]{#1}%{{\bf Amal: }#1}}

\newcommand{\mAP}{\operatorname{mAP}}
\newcommand{\acc}{\operatorname{acc}}
\newcommand{\task}{\mathcal{T}}
\newcommand{\SHORT}{\textsc{short} }

\newcommand{\defined}{:=}
\newcommand{\codelink}{\url{https://github.com/deepmind/dm_nevis}}

\newcommand{\metatrainstream}{\mathcal{S}^{\text{Tr}}}
\newcommand{\metateststream}{\mathcal{S}^{\text{Ts}}}
\newcommand{\numtasks}{$106$}
\usepackage{lastpage}
\jmlrheading{**}{2022}{1-\pageref{LastPage}}{mm/yy; Revised mm/yy}{mm/yy}{**-****}{Bornschein, Galashov, Hemsley, Rannen-Triki et al.}

\ShortHeadings{\minerva}{Bornschein, Galashov, Hemsley, Rannen-Triki et al.}
\firstpageno{1}

\begin{document}

\title{\minerva: A Stream of 100 Tasks Sampled \\ from 30 Years of Computer Vision Research}

\author{\name J\"org Bornschein\thanks{Equal contribution} \email bornschein@deepmind.com \\
       \name Alexandre Galashov\footnotemark[1] \email agalashov@deepmind.com \\
       \name Ross Hemsley\footnotemark[1] \email rhemsley@deepmind.com \\
       \name Amal Rannen-Triki\footnotemark[1] \email arannen@deepmind.com \\
       \name Yutian Chen \email yutianc@deepmind.com\\
       \name Arslan Chaudhry \email arslanch@deepmind.com\\
       \name Xu Owen He \email hexu@deepmind.com \\
       \name Arthur Douillard \email douillard@deepmind.com\\ 
       \name Massimo Caccia\thanks{Current affiliation: Mila - Quebec AI Institute. Work done while interning at DeepMind.} \email massimo.p.caccia@gmail.com\\ 
       \name Qixuan Feng \email qixuan@deepmind.com\\
       \name Jiajun Shen \email jiajuns@deepmind.com\\ 
       \name Sylvestre-Alvise Rebuffi\email sylvestre@deepmind.com \\ 
       \name Kitty Stacpoole \email kstacpoole@deepmind.com\\ 
       \name Diego de las Casas\email diegolascasas@deepmind.com \\ 
       \name Will Hawkins\email willhawkins@deepmind.com \\
       \name Angeliki Lazaridou \email angeliki@deepmind.com\\ 
       \name Yee Whye Teh \email ywteh@deepmind.com\\ 
       \name Andrei A. Rusu \email andrei@deepmind.com\\ 
       \name Razvan Pascanu \email razp@deepmind.com\\
       \name  Marc’Aurelio Ranzato \email ranzato@deepmind.com\\
       \addr DeepMind \thanks{For questions about the benchmark, please email us at \href{mailto:nevis@deepmind.com}{nevis@deepmind.com} or write to us at 14-18 Handyside Street, King’s Cross, London, N1C 4DN.  More details are provided in Appendix~\ref{app:authors}.} \\
       }

\editor{n/a}

\maketitle

\begin{abstract}%   <- trailing '%' for backward compatibility of .sty file
\mr{A shared goal of several machine learning communities like continual learning, meta-learning and transfer learning, is to design algorithms and models that efficiently and robustly adapt to unseen tasks. An even more ambitious goal is to build models that never stop adapting, and that become increasingly more efficient through time by suitably transferring the accrued knowledge.
Beyond the study of the actual learning algorithm and model architecture, there are several hurdles towards our quest to build such models, such as the choice of learning protocol, metric of success and data needed to validate research hypotheses. In this work, we}
 introduce the {\bf N}ever-{\bf E}nding {\bf VI}sual-classification {\bf S}tream (\minerva), a benchmark consisting of a stream of over $100$ visual classification tasks, sorted chronologically and extracted from papers sampled uniformly from computer vision proceedings spanning the last three decades. The resulting stream 
reflects what the research community thought was meaningful at any point in time\mr{, and it serves as an ideal test bed to assess how well models can adapt to new tasks, and do so better and more efficiently as time goes by}. Despite being limited to classification, the resulting stream has a rich diversity of tasks from OCR, to texture analysis, scene recognition, and so forth. The diversity is also reflected in the wide range of dataset sizes, spanning over four orders of magnitude. Overall, \minerva{} poses an unprecedented challenge for current sequential learning approaches due to the scale and diversity of tasks, yet with a low entry barrier as it is limited to a single modality and well understood supervised learning problems. Moreover, we provide a reference implementation including strong baselines and an evaluation protocol to compare methods in terms of their trade-off between accuracy and compute. We hope that \minerva{} can be useful to researchers working on continual learning, meta-learning, AutoML and more generally sequential learning, and help these communities join forces towards more robust 
%\mr{\sout{and efficient}} 
models that efficiently adapt to a never ending stream of data\footnote{Implementations have been made available at \codelink.}.  
\end{abstract}

\section{Introduction}
The machine learning community has focused extensively on the {\it stationary} batch setting, for which there exists a static and unchanging data distribution used to sample fixed training and test sets from~\citep{erm}. This has enabled the rigorous evaluation of learning systems, and driven unprecedented progress over the last four decades, across a wide range of domains \cite[e.g.][]{lecun2015deep, Jumper2021HighlyAP, GPT3, Flamingo}. Throughout this journey, researchers have spent a considerable amount of time and compute developing algorithmic and architectural improvements, adapting methods to new application domains, and developing insights into how to transfer their knowledge and know-how to new and more challenging settings. 

% shift over the last decade
% look at how to generalize to held out tasks, from held out examples
% automate process of designing architectures, algorithms, optimizers, etc.
% goal: more efficient and more automatic adaptation to a new task or data distribution
% CL, meta-learning, etc look at this problem under slightly different assumptions and using different metrics.
% ultimately, we'd like the system not to adapt only once after training, but to be able to be deployed and keep adapting over time. In fact, as more things are learned over time and more knowledge is somehow accumulated, we'd like such system to keep adapting and to learn more and more efficiently over time.
% Since adapting over a sequence of tasks is challenging (lots of open questions), we focus on supervised classification tasks, because they are well understood.
% In this work we focus on a benchmark, as current benchmark are too small, homogenous or artificial. With the objective to eventually deploy a model and let it learn over time with the tasks that our community generates, we take a hindsight approach and build a stream out of the tasks that our community has generated so far, and ask the q of whether a model that has learned on the tasks of the first few years, can generalize well to the tasks of the more recent years.

\mr{Over the past decade, there has been a surge of interest in the design of learning algorithms that generalize not only to novel examples, but also to entirely new tasks} \art{~\citep{vtab, wald2021calibration, gulrajani2020search, Meta-Dataset}. This line of research relates to efforts to automate the design process of architectures~\citep{bai2021ood,ardywibowo2020nads} and improve learning algorithms~\citep{arjovsky2019invariant, triantafillou2021learning}.}
\mr{Broadly speaking, the goal of this new  endeavors is to understand the principles and to design learning algorithms that enable further adaptation  after training. In fact, training never really ends. The system observes a never-ending sequence of tasks and the question is how it can adapt faster and better over time. Can it succeed by leveraging its ever increasing knowledge of the world acquired through its past experiences, while limiting as much as possible human intervention?}  

\mr{
There are several open questions in this research area, from how to represent knowledge, to how to accrue knowledge over time, how to retain computational efficiency, etc. In this work we focus on the methodology and data used as playground for advancing research in this area. First, we consider a stream of vision classification tasks. Each such task is very well understood, the only remaining challenge is how to automatically and efficiently learn such tasks in  sequence. 
Second, we take a \textit{hindsight} approach to the benchmark construction process. Our objective is to eventually deploy a system that is capable of automatically learning whatever task the research community comes up with and, by doing so, to become more apt at solving any other future task. We therefore build a stream by sorting chronologically the tasks that the research community has introduced and used over the last three decades. We then assess  whether models that have learned on all the tasks up to time $t$,  can better learn the next task, and whether learning becomes more effective and efficient for larger values of $t$.}

\mr{This construction process stands in stark contrast to how current benchmarks are built. These are often very small scale which prevents the assessment of efficiency of learning, they are very homogeneous which prevents the assessment of robustness of learning, and they are built out of a small number of hand picked tasks which might poorly represent the task distribution  of interest to our community.
}

%\mr{\sout{In this work, we consider whether this traditional procedure for developing models can be made more efficient and further automated. This is important to enable our community to make faster progress and to further scale up our models, a process that has been so critical to the recent success of the field.
%Taking the perspective that the machine learning community itself can be seen as an agent interacting with data to produce artifacts, we explore approaches that can robustly and automatically develop models for new tasks, while accruing knowledge over time to further accelerate learning in the future. }}

%\mr{ \sout{Unfortunately, there is no consensus on how to measure the speed of adaptation or knowledge accrual.
% \citep{Parisi18review, survey_cl20, mundt2022clevacompass}.    % JB: striking out citations causes compile errors.
%}  }

%\mr{\sout{Moreover, current benchmarks often focus on other problems, such as catastrophic forgetting 
%\citep{catastrophic_french}  % JB: striking out citations causes compile errors.
% , are too small, or lack sufficient diversity.
%\citep{chaudhry2020continual}.   % JB: striking out citations causes compile errors. 
% }}

This motivates us to introduce \minerva, a challenging stream which comprises \numtasks{} tasks, all representing publicly available datasets from the last 30 years of computer vision research. By construction, \minerva{} tracks what the vision community has deemed interesting over time, since tasks are sorted according to the year in which they appeared in publications. Over time, new and more challenging domains are considered, datasets get larger, and overall there are more opportunities to transfer knowledge from an ever growing set of related tasks.

%\mr{ \sout{ Since each task in isolation is very well understood, the only research question remaining is how to robustly and effectively adapt to tasks over time.} } 
As an indirect measure of whether a system is capable of accruing knowledge over time, we assess performance not just in terms of final error rate, but also compute needed to reach such performance. The assumption is that, if a method can transfer knowledge from related past tasks, then it can quickly learn the next task using less compute.

We believe \minerva{} should appeal to and challenge researchers in several communities. It should attract researchers in continual learning~\citep{Ring1994, thrun-iros94} because the stream is non-stationary. Some of the tasks are repeated over time, providing a natural opportunity to measure forgetting and forward transfer~\citep{DBLP:conf/nips/Lopez-PazR17, ChaudhryDAT18, Schwarz0LGTPH18,survey_cl20, Parisi18review}. It should empower researchers in meta-learning~\citep{DBLP:books/sp/98/ThrunP98} because there is a rich structure across tasks, which should enable the study of learning-to-learn. Finally, it should be useful to researchers in AutoML~\citep{autoweka} as each task has to be solved in a black box manner, without humans in the loop. Since our metrics include the compute used during hyper-parameter search, \minerva{} incentivizes the development of efficient approaches for algorithm, architecture and hyper-parameter search. For the very same reasons, however, \minerva{} also constitutes a challenge, as it requires tools from each of these communities. Moreover, \minerva{} is the first benchmark simulating supervised never-ending learning at this scale, and with such rich and diverse set of realistic tasks.
\minerva{} is accompanied by code to reproduce the stream, the training and evaluation protocols, and representative baselines we have considered. \ag{We summarize the main findings in Table~\ref{tab:main_findings} with more detailed discussion in Section~\ref{sec:main_results} and in the Appendix. }

\begin{table}[tb]
    \begin{center}
        \begin{tabular}{p{0.7\linewidth}||p{0.25\linewidth}}
            {\bf Findings} & {\bf References} \\
            \hline
            \hline
            \minerva{} enables the comparison of methods in terms of their compute-performance trade-off & Fig.~\ref{fig:full_stream-pareto_fronts}\\
            % Methods need to be compared on a pareto front, measuring performance for the same compute budget; otherwise, we may favor more expensive approaches. & Fig.~\ref{fig:full_stream-pareto_fronts}\\
            \hline
            \minerva{} favors methods that leverage knowledge transfer across tasks (e.g., various forms of fine-tuning)  & Section~\ref{sec:main_results}, Figure~\ref{fig:full_stream-pareto_fronts}, Figure~\ref{fig:regret_plots}, Section~\ref{apx:finetuning}, Figure~\ref{fig:ft-d}, Figure~\ref{fig:pareto_finetuning}, Figure~\ref{fig:full_stream-per-domain-ablation}  \\
            \hline
            \minerva{} enables the study of how to use and adapt pretrained representations  & Section~\ref{sec:main_results}, Figure~\ref{fig:full_stream-pareto_fronts}, Figure~\ref{fig:regret_plots}, Section~\ref{apx:pretraining} \\
            \hline
            \minerva{} shows that current methods are not capable of transferring from a large number of smaller datasets & Tab.\ref{tab:cripple_stream_ablation} \\
            \hline
      \minerva{} supports fine-grained analysis (domain, forward-transfer, etc.) & Fig.~\ref{fig:full_stream-per-domain-ablation}, \ref{fig:forward_transfer}  \\
            \hline

        \end{tabular}
    \end{center}
\caption{Summary of main findings. For more information, please refer to Section~\ref{sec:main_results}. Additional results are given in Appendix.}
\label{tab:main_findings}
\end{table}
\section{Related Work}
\begin{table}[tb]
\begin{center}
\begin{tabular}{lcccc}
\toprule
 & sequential & causal & memory restrictions & compute in the metric \\
\midrule
continual learning & yes & no & yes & no \\
meta-learning & no & - & no & no \\
auto-ml & no & - & no & no \\
lifelong auto-ml & yes & no & no & yes \\ 
\bf{\minerva{}} & \bf{yes} & \bf{yes} & \bf{no} & \bf{yes} \\
\bottomrule
\end{tabular}
\end{center}
\caption{\mr{Comparing learning frameworks across several axes, namely whether the learner observes tasks in sequence, it has access to future task while doing task specific hyper-parameter search (i.e., the model is allowed to do several passes over the stream), it has memory restrictions when accessing data and model parameters of past tasks, and whether compute is accounted in the evaluation. Note that this is an oversimplification, as often papers use intermediate setups.} }
\label{tab:comparing_protocols}
\end{table}

\begin{table}[tb]
\begin{center}
\begin{tabular}{lcccc}
\toprule
 & sequential & large-scale & diversity & compute in metric \\
\midrule
MNIST~\citep{mnist} variants & yes & no & no & no \\
CIFAR~\citep{cifar_100_introducing} variants & yes & no & no & no \\
CTrL~\citep{veniat2021efficient} & yes & yes & no & yes \\
CLOC~\citep{cloc} & yes & yes & no & no \\
CLEAR~\citep{CLEAR} & yes & yes & no & no \\
VTAB~\citep{vtab} & no & yes & yes & no \\
\art{Meta-Dataset~\citep{Meta-Dataset}} & no & yes & yes & no \\ 
\bf{\minerva{}} & \bf{yes} & \bf{yes} & \bf{yes} & \bf{yes} \\
\bottomrule
\end{tabular}
\end{center}
\caption{\mr{Comparing benchmarks made of several classification tasks.}}
\label{tab:comparing_benchmarks}
\end{table}

In this section we put our work in the broader context of the literature with a twofold goal. First, we relate the \minerva{} learning setting with existing learning frameworks such as continual learning, meta-learning and AutoML. Second, we contrast \minerva{} with existing benchmarks and highlight its unique features. \mr{Tables~\ref{tab:comparing_protocols} and \ref{tab:comparing_benchmarks} provide a high level overview.}

Continual (or Lifelong) Learning studies the question of learning under a non-stationary data distribution~\citep{silver2013lifelong, chen2018lifelong, survey_cl20, Parisi18review}. It typically assumes a series of tasks. The objective is to learn sequentially while achieving a list of desiderata ranging from avoiding catastrophic forgetting~\citep{catastrophic_f}, to leveraging forward or backward transfer\footnote{\mr{A model has positive backward (forward) transfer when performance on a past (future) task improves upon learning a new (previous) task. }}~\citep{DBLP:conf/nips/Lopez-PazR17}. Additional restrictions are typically considered, such as preventing access to previous data, limiting or accounting for the use of compute or memory. Given the multitude of potential desiderata and trade-offs of interest \citep{survey_cl20}, the literature has flourished with a considerable number of specialized benchmarks, each targeting a different scenario. In this work, we make the following central assumption: The learner cannot access data from novel future tasks. \mr{However, } 
%\sout{In particular},} 
accessing past data (or even past models) is permitted, although the compute cost of doing so will be taken into account in the final reporting. Our rationale is that, in modern applications of machine learning, memory \jb{for storing training data} is cheap relative to compute and time. We therefore focus on leveraging forward transfer for future tasks of interest to the community, rather than avoiding catastrophic forgetting or \mr{imposing } data storage limitations.

In continual Reinforcement Learning (RL)~\citep{Ring1994, khetarpal2020continual}, the non-stationarity is either imposed by changing environments, or it emerges from the interaction with the environment. While RL provides a natural test-bed for continual learning, it also makes it difficult to separate the challenges raised by exploration and learning with sparse rewards from the core continual learning problem of accruing knowledge over time. Attempts to studying the question of forward transfer have nonetheless been made~\citep{ContinualWorld}.
Similarly in the language domain, 
there have been studies on continuous training of language models with related benchmarks defined on sorted streams of text~\citep{streamingqa, temporalwiki}. Although very interesting, the lack of clear task structure or distinctions makes it difficult to assess when new concepts are introduced in data streams, and when the system is expected to learn new capabilities or skills.

In vision research, most existing benchmarks focus on measuring catastrophic forgetting and are derived from popular datasets such as MNIST~\citep{mnist}, CIFAR-10~\citep{cifar10}, ImageNet~\citep{imagenet} or Omniglot~\citep{omniglot}, whereby a stream is created by partitioning the data into disjoint subsets. This construction however greatly limits the diversity of the resulting stream. There are however exceptions. The Core50 dataset~\citep{CORE50} specifically collected realistic images of objects under different poses, to test continual learning capabilities in a setting most relevant for  robotics. The CLEAR benchmark~\citep{CLEAR} looks at temporal evolution of a set of visual concepts. CLOC~\citep{cloc} is a geo-localization task with a large collection of chronologically ordered images. Once again, these benchmarks target the setting of a single non-stationary task, as opposed to a sequence of a diverse set of tasks.
Forward transfer has become a more prominent goal of CL through benchmarks such as CTrL~\citep{veniat2021efficient}, which is however limited in its scale and diversity,  because it is entirely derived from just a handful of small datasets. 

Meta-learning assumes access to a distribution of tasks \citep{learning_to_learn, finn2018learning, metalearning_survey}. The goal is to learn from the observed tasks a mechanism that allows efficient learning on hold-out tasks from the same distribution.  Most popular benchmarks like VTAB~\citep{vtab} and Meta-Dataset~\citep{Meta-Dataset} focus on few-shot learning, while \minerva{} considers a variety of dataset sizes (including some with a handful of examples) and goes beyond a few steps of adaptation to characterize performance efficiency trade-offs. For example, \minerva{} accounts for compute in addition to error rate. 

Much of the field of AutoML is \mr{concerned with the \textit{automatic} discovery of algorithms, architectures and optimisers for a given new task. Auto-ML is mostly }focused on tabular tasks and shallow predictors. Benchmarks are often derived from the OpenML platform~\citep{OpenML2013}.  %\footnote{\url{https://www.openml.org/}}. 
The major limitation of current AutoML approaches is their computational cost, since one evaluation requires a full training run with a particular hyper-parameter setting. For instance, na\"ive neural architecture search would be too costly if applied within the inner loop of \minerva{}, as there are over 100 tasks. More recently, there has been excitement around lifelong AutoML~\citep{init_automl15, warmstarting-automl}, but again, evaluation has been limited to very few and very similar tasks. \minerva{} gives a more natural playground to explore ways to transfer knowledge at the level of the meta-learner thanks to shared structure across tasks. Moreover, \minerva{} sets incentives towards the development of more efficient AutoML methods because the evaluation accounts for compute spent during hyper-parameter search and prizes more parsimonious meta-learners.

Transfer learning \citep{transfer_learning, Bengio12transfer, weiss2016survey, tan2018survey, zhuang2020comprehensive} is a general and well studied paradigm for addressing the problem of leveraging one or a few related source tasks to improve the learning of a known target task. Techniques such as self-supervised learning \citep{chen2020big, grill2020bootstrap} and large-scale pretraining \citep{jia2021scaling, radford2021learning} are recent examples of successful approaches in computer vision, see \citep{Jaiswal2020asurvey} for a survey. The sequential aspect of data acquisition is often neglected in the transfer learning setting. \minerva{} provides a good test bed for testing transfer learning ideas at scale, and in a more realistic setting, where the system needs to keep adapting over time, as opposed to only once. In this work, we do consider several variants of pretraining among our baselines, and demonstrate that enabling continuous adaptation further improves their performance.

\section{The \minerva{} Benchmark}
The {\bf N}ever-{\bf E}nding {\bf VI}sual-classification {\bf S}tream benchmark, dubbed \minerva{}, is a playground for research in \mr{never-ending } %\sout{sequential}} 
learning. We start by \mr{summarizing its motivation and }discussing how it was built and conclude with an analysis highlighting its key features.

\subsection{\mr{Motivation}}
\mr{Our ultimate goal is to build a robust, efficient and autonomous never-ending learning system. For instance, we would like to provide the machine learning community with  a never-ending learning model which can learn and integrate knowledge from whatever tasks the community considers at any point in time, and be used as a baseline for new methods being published. We do not make further assumptions on the task distribution, as this is generated by the community. We would like such never-ending learning system to never stop adapting, and to become more efficient over time, despite being exposed to more and more data and more and more complex tasks. Of course, there are many practical applications of such never-ending learning system, from auto-ml applications to robotics. In this work, we do not investigate what such a model could be, but focus on a benchmark which can be useful to develop such a model.}

\subsection{Stream Construction} \label{sec:stream_construction}
\begin{figure}[t]
  \centering
  \includegraphics[width=.9\textwidth]{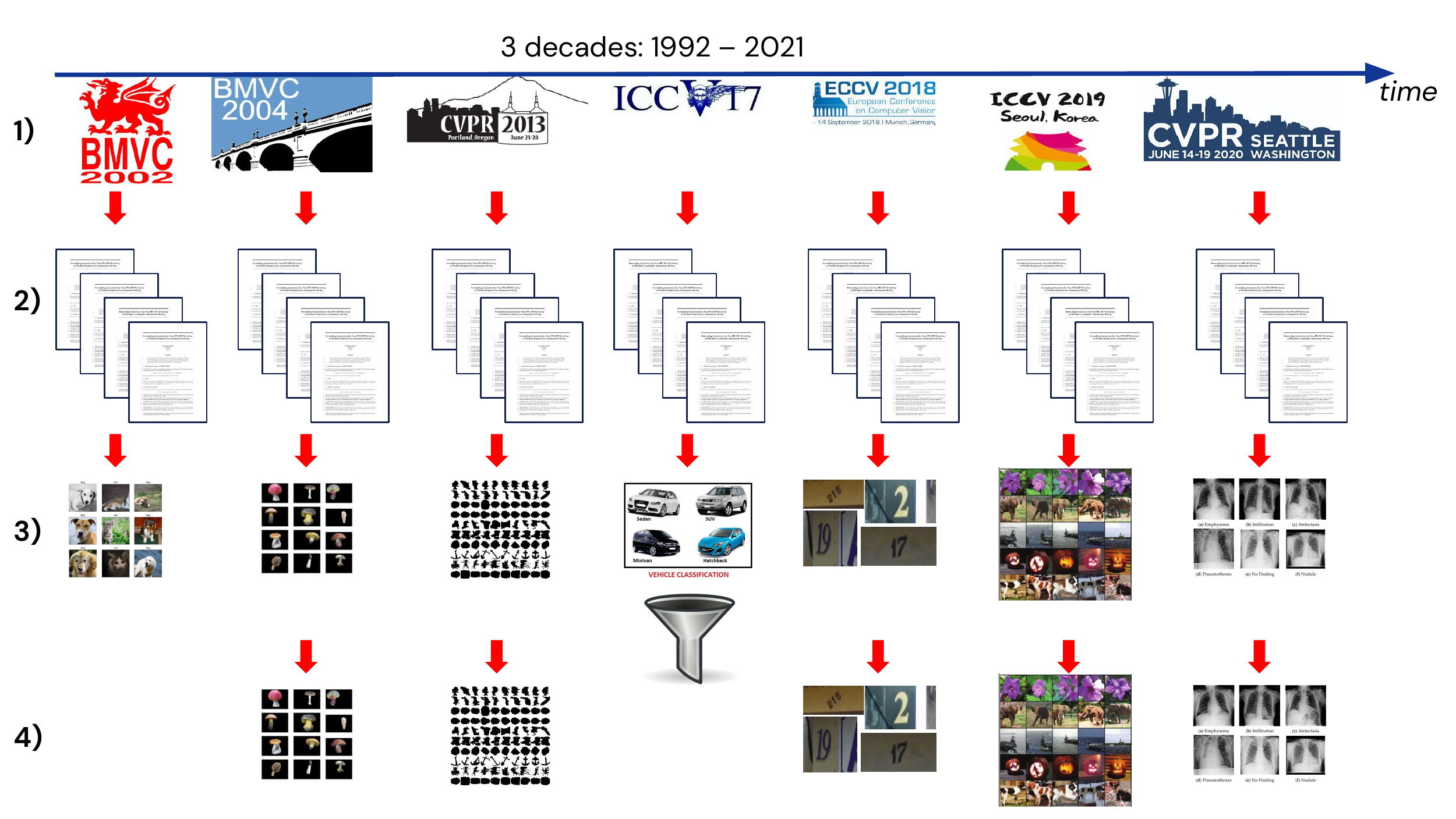}
\caption{\mr{Illustration of the steps used to construct \minerva. First, we gathered and sorted chronologically proceedings of various computer vision conferences. Second, we sampled papers at random from each year. Third, for each paper we extracted the tasks used in the empirical validation. Finally, we filtered tasks. For instance, we retained only classification tasks that used publicly available data (see text for more detail).}
}
  \label{fig:stram_construction}
\end{figure}

The \minerva{} benchmark is constructed according to four principles. 
1) {\bf Reproducibility:} This is an artifact for the research community, and therefore, data needs to be publicly available under permissive licenses. 
2) {\bf Simplicity:} The focus is on effective learning of {\em a sequence} of tasks, and therefore, each task must be well understood when taken in isolation.
3) {\bf Agnostic task selection:} The selection of which tasks to include in the stream should not aim at favoring any particular approach. 
4) {\bf Scale:} The benchmark has an intermediate scale, useful for research in sequential learning. It is 
 sufficiently large-scale to rule out approaches that do not scale gently with the amount of data. It is not too large to impede fast iteration of research ideas.

The protocol used in building \minerva{} is \mr{illustrated in Fig.~\ref{fig:stram_construction}}. 
%\sout{the following}}. 
We first gathered papers from leading computer vision conferences and workshops that host their proceedings publicly. We considered the British Machine Vision Conference (BMVC), the European Conference in Computer Vision (ECCV), the Computer Vision and Pattern Recognition conference (CVPR), the International Conference in Computer Vision (ICCV), ML4Health and Medical Imaging Workshops at NeurIPS. We sampled {\em uniformly at random} $90$ papers each year (if available), from 1989 until 2021. Secondly, we manually extracted the tasks used for empirical validation. We filtered these tasks, retaining only i) classification tasks or tasks that can be mapped to classification, ii) tasks for which the corresponding dataset is publicly available, it is not deprecated and it has a permissive license for research purposes. Thirdly, we removed any duplicates that appear within a window of 10 years, retaining only the first instance. The rationale was to make the stream not too long or redundant, yet enabling the assessment of whether learning on subsequent instances of a dataset is faster or better. For example, using the heuristic above, we kept only the first instance of ImageNet from a paper published in 2011, removed all duplicate instances from years 2012 until 2020 and only retained a second instance from a paper published in 2021. 
Finally, the stream is the sequence of these tasks presented in the order in which they appeared in publications.
We partition each dataset into three splits, namely training, validation, and test (see Sec.~\ref{sec:protocol} for details). We remove any duplicate example to make splits and tasks fully disjoint. The full list of tasks is reported in Appendix~\ref{app:ds_list}.

\subsection{Stream Statistics} \label{sec:stream_stats}

\begin{figure}[t]
  \centering
  \includegraphics[width=.45\textwidth, trim={0 0 3cm 0}, clip]{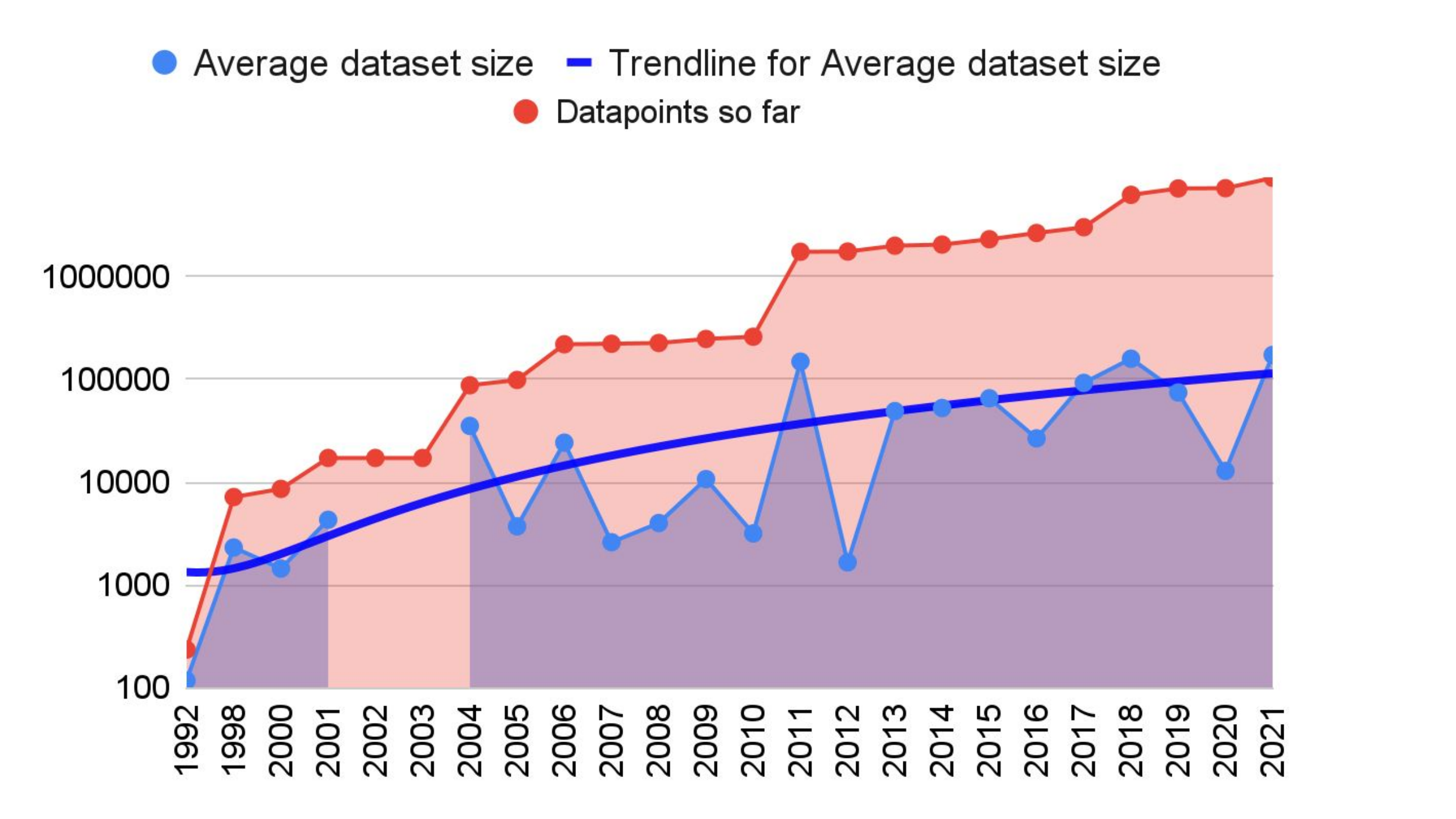}
  \includegraphics[width=.45\textwidth]{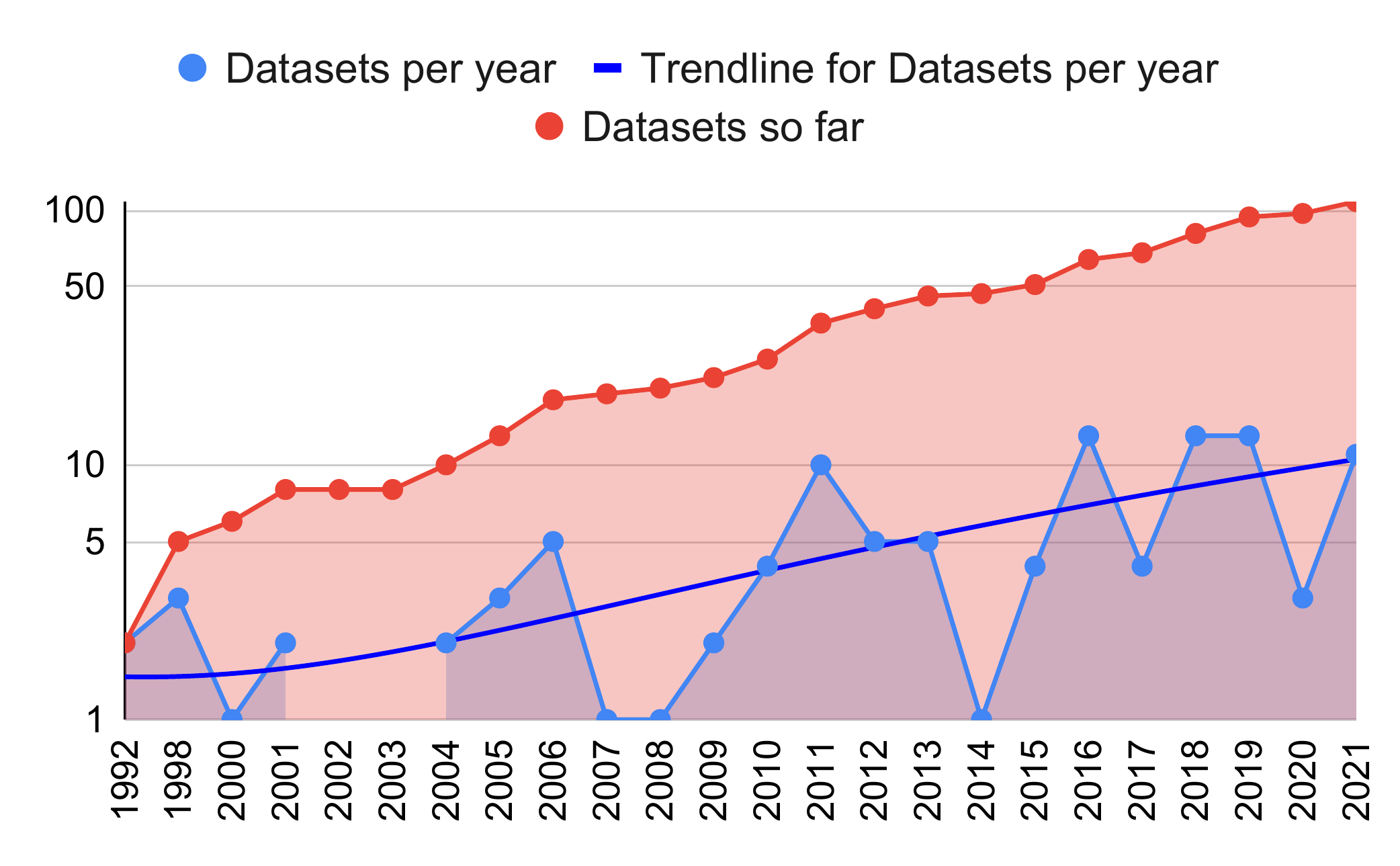}
\caption{
Left: Histogram of average dataset size each year, and cumulative number of datapoints in the stream. Right:  Histogram of datasets per year. Most datasets have between 1000 and 20000 examples. \mr{The gap between 2001 and 2004 is due to duplicate removal, see section~\ref{sec:stream_construction} for details.}
As expected, dataset sizes tends to increase over time. Note the log-scale of the plot. 
%\art{\sout{We remove duplicates in a ten-year window, which results in a gap between 2001 and 2004.}}
}
  \label{fig:stats_datasetsize}
\end{figure}

\begin{figure}[t]
  \centering
  \includegraphics[width=.55\textwidth]{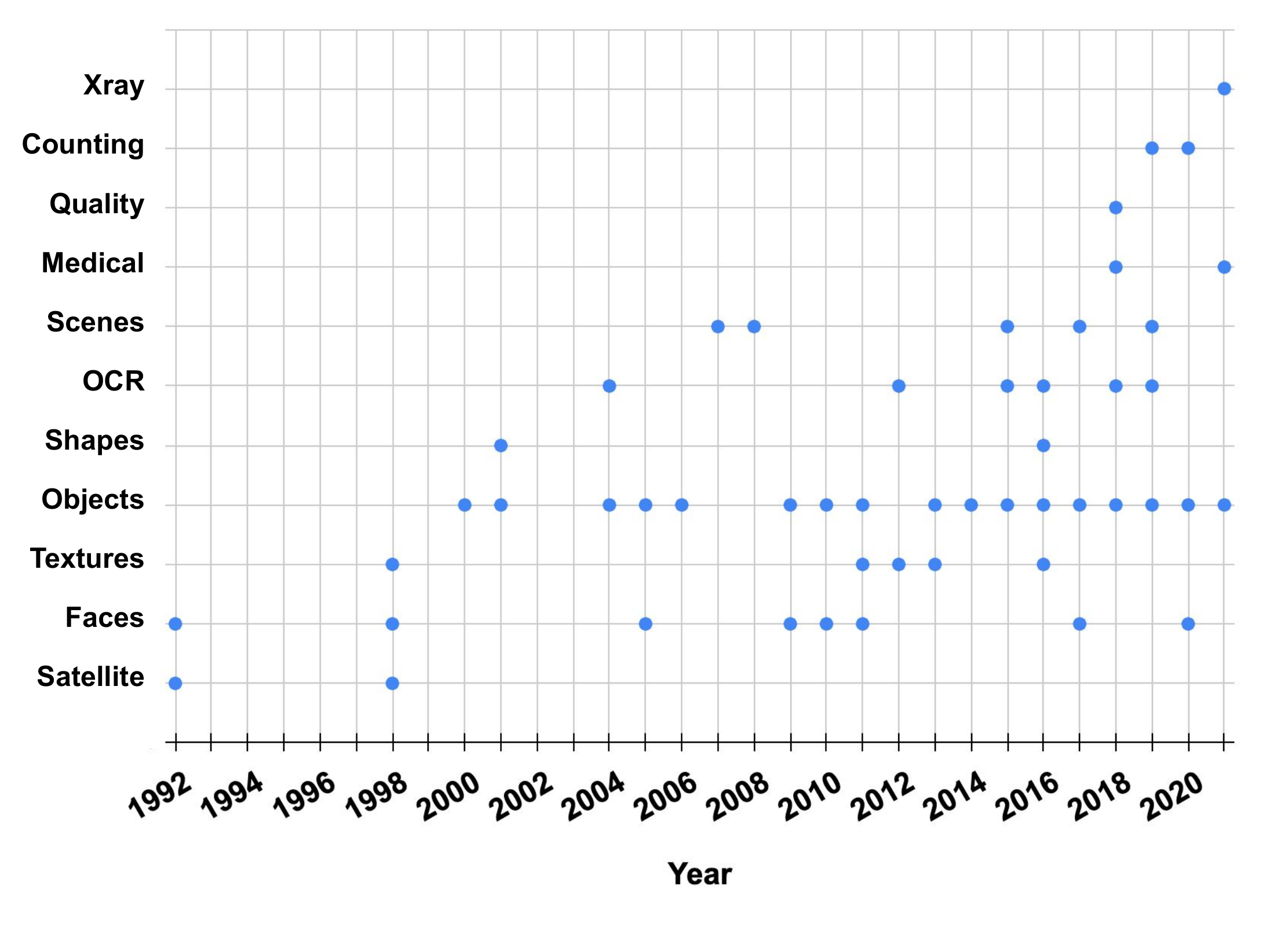}
  \includegraphics[width=.4\textwidth]{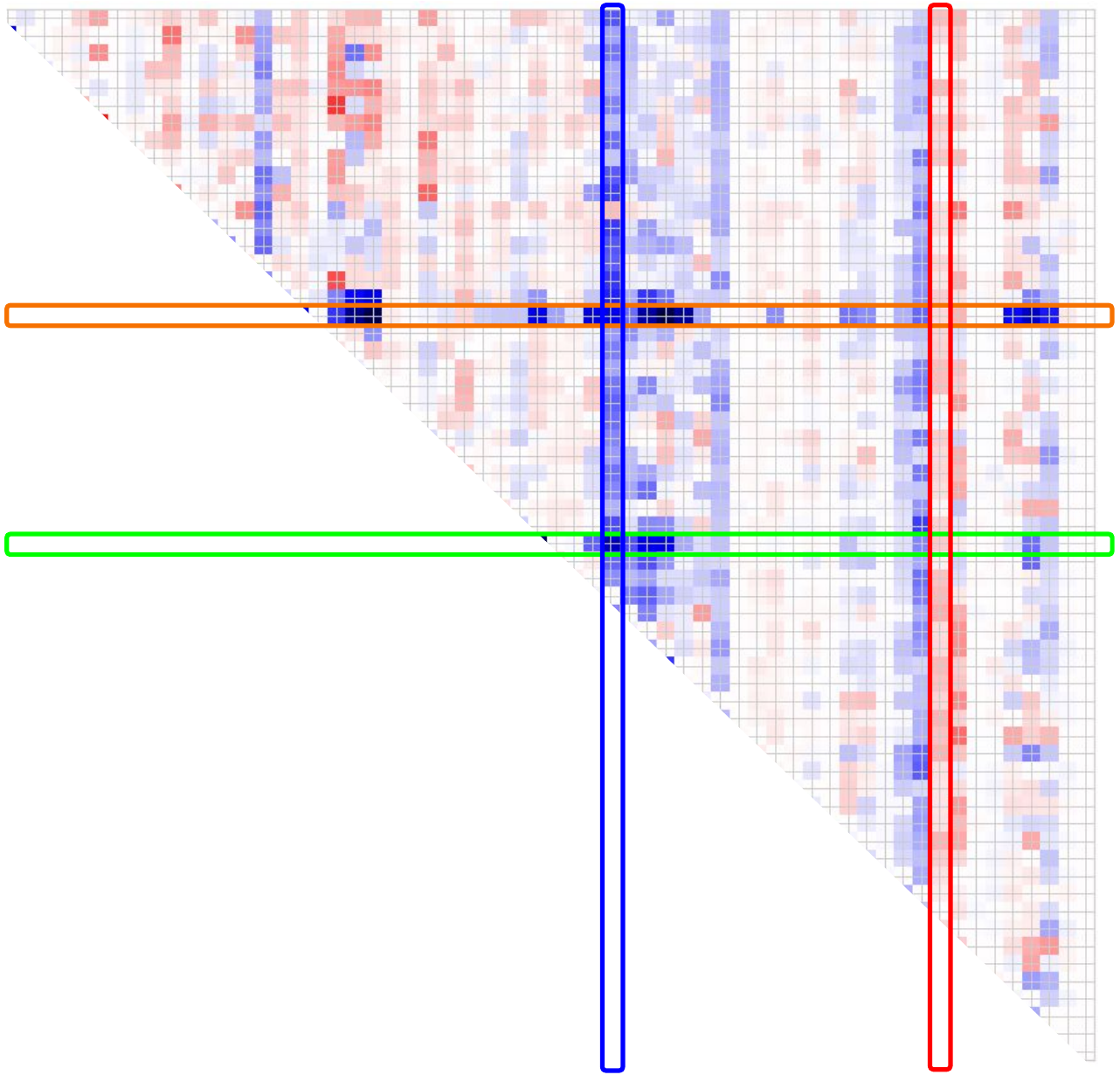}
\caption{Left: Scatter plot showing the domains present in each year of the \minerva{} stream. Each dot represents the presence of at least one dataset of a given domain in a certain year. The number of domains increases over time, and the popularity of domains vary with time.
Right: Upper-triangular transfer matrix on a subset of tasks extracted from \minerva. The figure shows at position $(i,j)$ the advantage of pretraining on task $i$ before finetuning on task $j$, compared to learning task $j$ from scratch. The upper triangular section shows only transfer from \emph{tasks that had already occurred in the sequence}. Shade of blue indicate positive transfer (i.e. pretraining was useful), while shade of red indicate negative transfer.  We notice that there exists tasks that are good for pretraining, leading to positive transfer to most future task (e.g., \art{orange and green} rectangles which correspond to ImageNet and SUN397 \art{respectively}), tasks that can transfer well from any other tasks (e.g., the blue rectangle corresponding to the Stanford Cars dataset) and tasks that do not transfer well from any other task (e.g., the red rectangle corresponding to the Mall dataset).
\jb{See Appendix \ref{app:transfer} for details.}
  }
  \label{fig:stats_domain}
\end{figure}

\minerva{} is a stream composed of \numtasks{} image classification datasets totalling approximately $8$ million images.
There is a large diversity in data.  For instance, the input {\em resolution} goes from $3\times3$ all the way to $2000\times3000$. Some datasets have fixed resolution, while for others each example has its own resolution.

The number of examples in each dataset also varies considerably, spanning four orders of magnitude, as it can be seen in Figure~\ref{fig:stats_datasetsize} left panel. Dataset size tend to increase over the years. 
Perhaps most importantly, \minerva{} contains a large variety of domains, and yet within each domain there are enough datasets to support potentially beneficial transfer.
The left plot of Figure~\ref{fig:stats_domain} shows the major families of domain and their evolution over time. There are interesting patterns of non-stationarity, for instance with some domains appearing throughout the stream (e.g., object), while others being popular only in short time windows (e.g., satellite). 

\mr{Note that such non-stationarity is a natural and desirable feature of \minerva, as it enables the development of models in a condition similar to deployment. Recall that at deployment time, the never-ending learning system might encounter tasks from entirely novel domains (for instance, around 2015 when the community started working on crowd counting or in 2020 when it started working on COVID-19 related tasks), as well as tasks with much more data than previously encountered (for instance, in 2009  when the ImageNet was introduced for the first time). \minerva{} reproduces such natural stream of tasks, without making assumptions on the inner working of the never-ending learners.}\footnote{\mr{Notice that chronological order and uniform sampling of tasks might be undesirable If the objective were to merely  maximize accuracy on a particular domain, for instance.}}

On the right side of Figure~\ref{fig:stats_domain} we can also see how datasets relate to each other. The transfer matrix shows interesting structure, with both positive and negative transfer. 

The scale, diversity and agnostic selection used in the construction process are the defining elements of \minerva{}, compared to existing benchmarks that might satisfy some of these desirable axes but, to the best of our knowledge, not all of them.

\begin{algorithm}[t]
    \centering
    \caption{Training \& Evaluation Protocol in \minerva}\label{alg:protocol}
    \begin{algorithmic}[1]
        \State \emph{\textbf{\# Initialization.}}
        \State \text{Meta-train stream: $\metatrainstream = (\task_1, \dots, \task_n)$}
        \State \text{Meta-test stream: $\metateststream = (\task_{n+1}, \dots, \task_{n+m})$}
        \State \text{Entire stream: $\mathcal{S} =  \metatrainstream +  \metateststream$}
        \State \text{$i$-th task: $\mathcal{T}_i = (\mathcal{D}_i^{\mbox{tr}},  \mathcal{D}_i^{\mbox{val}}, \mathcal{D}_i^{\mbox{ts}} )$.}
        \State \text{Meta-learner: $M$.}
        \State \emph{\textbf{\# Meta-training phase:} Tuning $M$'s hyper-parameters $\lambda^M$.}
        \Repeat
        \State \text{Designer chooses M's hyper-parameters $\lambda^M$.}
        \State \text{Initialize meta-learner state $s_0$ based on $\lambda^M$.}
        \For{$\task_i \in \metatrainstream$}
        \State \text{$\mathcal{D}_i^{\mbox{tr}},  \mathcal{D}_i^{\mbox{val}} \leftarrow \task_i$}
        \State \text{$P_i, \mbox{FLOP}_i \leftarrow \text{Using } s_{i-1}, \mathcal{D}_i^{\mbox{tr}} \text{ and } \lambda^M: M \text{ performs h.p. search and trains } P_i$}.
        \State \text{$s_i \leftarrow M \text{ updates its state using } \mathcal{D}_i^{\mbox{tr}}, s_{i-1} \text{ and } \lambda^M$}.
        \State \text{$e_i \leftarrow \text{error rate of } P_i \text{ on } \mathcal{D}_i^{\mbox{val}}$}
        \EndFor
        \State \text{$\mathcal{E}(\metatrainstream) = \sum_{i=1}^n e_i$}
        \State \text{$\mbox{cFLOP}(\metatrainstream) = \sum_{i=1}^n \mbox{FLOP}_i $}
        \Until {Designer is happy with choice of $\lambda^M$ based on $M$'s performance.}
        \State \emph{\textbf{\# Meta-test phase:} Evaluating $M$.}
        \State \text{Initialize meta-learner state $s_0$ based on $\lambda^M$.}
        \For{$\task_j \in \metatrainstream + \metateststream$}
        \State \text{$\mathcal{D}_i^{\mbox{tr}},  \mathcal{D}_i^{\mbox{ts}} \leftarrow \task_j$}
        \State \text{$P_j, \mbox{FLOP}_j \leftarrow \text{Using } s_{j-1}, \mathcal{D}_j^{\mbox{tr}} \text{ and } \lambda^M: M \text{ performs h.p. search and trains } P_j$}.
        \State \text{$s_j \leftarrow M \text{ updates its state using } \mathcal{D}_j^{\mbox{tr}}, s_{j-1} \text{ and } \lambda^M$}.
        \State \text{$e_j \leftarrow \text{error rate of } P_j \text{ on } \mathcal{D}_j^{\mbox{ts}}$}
        \EndFor
        \State  \text{$\mathcal{E}(\metateststream) = \sum_{i=n+1}^{n+m} e_i$.}
        \State  \text{$\mbox{cFLOP}(\metatrainstream + \metateststream) = \sum_{i=1}^{n+m} \mbox{FLOP}_i $.}
        \State \textbf{Return and report}
        \art{
        \State
        \text{Average error rate of meta-test stream: $\mathcal{E}(\metateststream)$}
        \State
        \text{Cumulative FLOPs of entire stream: $\mbox{cFLOP}(\metatrainstream + \metateststream)$}
        }
    \end{algorithmic}
\end{algorithm}

\section{Learning \& Evaluation Protocol} \label{sec:protocol}

We recall that the \minerva{} benchmark operates on a sequential stream of diverse tasks; and that learners are evaluated on their ability to efficiently generalize what they have learned early on in the stream to tasks that appear later. An important aspect of this setting is that learners are not permitted to use future observations to influence their behavior on a given task. Most notably, this condition is also intended to apply to the selection of the hyper-parameters used by learners. In particular, learners are not permitted to select hyper-parameters based on metrics computed on runs over the full stream, since this is equivalent to using information about future tasks in the stream to influence the present. This decision has been made to encourage algorithm designers to build robust learning systems that can truly adapt to changing distributions automatically, rather than through careful human-driven hyper-parameter tuning. These requirements necessitate a rigorous evaluation protocol that supports both the iteration and development of learners (including their meta-learning components), and also meaningful comparisons of learner implementations once they have been tuned. In this section, we outline the strategies we have adopted to support these requirements.

In all generality, we assume that there exists a ``meta-learner'' $M$ that is in charge of instantiating a predictor $P_i$ for the $i$-th task. $M$ is responsible for determining any hyper-parameters (such as learning rate and label smoothing values) needed to construct $P_i$, and also for possibly leveraging the results of previously observed tasks, if doing so could enable the learner to more efficiently solve the current task. For instance, $M$ could initialize the parameters of $P_i$ from the parameters of a network trained on a related previous task. Furthermore, $M$ itself might have hyper-parameters; for instance, they could be the range of values considered by random hyper-parameter search, or the choice of transfer learning method. How shall we cross-validate the meta-learner $M$ and each predictor $P_i$? How should we account for the compute used by both $M$ and $P_i$? To answer these questions, we propose the 
 training and evaluation protocol described in Algorithm~\ref{alg:protocol}. 

Since we are interested in assessing generalization to future tasks, we divide \minerva{} into two sub-streams (line 2 and 3): the ``meta-train stream'', denoted by $\metatrainstream$, and ``meta-test stream'', denoted by $\metateststream$. $\metatrainstream$ comprises the tasks contained in the first $27$ years for a total of $79$ tasks, $\metateststream$ contains the $27$ tasks from the last $3$ years. \mr{The choice of how many tasks to include in $\metateststream$ versus $\metatrainstream$ strikes a good trade-off between: a) having a sufficient number of tasks that can be used for development and b) having enough tasks to assess generalization at meta-testing. The last $27$ tasks which make $\metateststream$, are listed in tab.~\ref{app:ds_list}. Among the 27 datasets, there are some duplicate from metatrain (e.g., ImageNet, Oxford Flowers 102) and datasets from various domains like OCR, object, counting, scene understanding, medical, etc. Four datasets are from a new sub-domain, medical COVID-19 x-ray. We therefore believe $\metateststream$ provides a nice coverage of the scenarios encountered by a never-ending learning system.} 

Finally, each task $\task_i$ consists of three datasets, namely training $\mathcal{D}_i^{\mbox{tr}}$, validation $\mathcal{D}_i^{\mbox{val}}$, and test $\mathcal{D}_i^{\mbox{ts}}$. Next we explain how these streams and datasets splits are used.

\subsection{\mr{Meta-training Phase}} \label{sec:meta-training}
During the meta-training phase (lines 8 to 19), a designer (effectively, a \emph{meta} meta-learner) might run the meta-learner $M$ multiple  times over  $\metatrainstream$ to tune $M$'s hyper-parameters $\lambda^M$; for instance, this might include choosing neural network architectures, optimizers, data-augmentation strategies and intialization parameters. It is up to the designer to decide which configurations to try next (line 9), and when to stop the search (line 19). 
At every such iteration, $M$ sweeps over the tasks (or any subset thereof) of $\metatrainstream$. It 
first extracts the task specific training and validation set (line 12). Then it uses the training set $\mathcal{D}_i^{\mbox{tr}}$ to produce a predictor $P_i$ for the $i$-th task. This process typically involves some form of task-level search over $P_i$'s hyper-parameters. In order to support this, \minerva{} provides a default decomposition of $\mathcal{D}_i^{\mbox{tr}}$ into two sets, one used for actual training of $P_i$ and the other used for task-level cross validation. However, it is up to the meta-learner to decide whether to use this or other ways to partition  $\mathcal{D}_i^{\mbox{tr}}$ to better support $P_i$'s hyper-parameter search. The result of this step is not only $P_i$, a predictor for task $i$, but also the total number of floating point operations used during this training and hyper-parameter search process, denoted by $\text{FLOP}_i$. 

Notice that $M$ uses a certain configuration of its own hyper-parameters $\lambda^M$ and an internal state $s_{i-1}$ to find $P_i$. Examples of $\lambda^M$ could be the range of learning rate values used during the actual random search of $P_i$'s learning rate. 
The state $s_i$ instead is what represents the knowledge accrued up to the $i$-th step. This can be an empty set if $M$ instantiates independently learners to tasks. It could also consists of the set of parameters used by $P_j$ for $j < i$, supporting various kinds of finetuning strategies from models trained on previously encountered tasks. This state is updated by $M$ in line 14. In the previous example, this merely consists of adding an additional parameter vector to a pre-existing set of parameter vectors, one for each previous task. Finally, $P_i$ is evaluated on $\mathcal{D}_i^{\mbox{val}}$ (line 15) in terms of error, as follows:
\begin{gather}
    \label{eq:task_error}
    e_{i}
    \defined
    \begin{cases}
        1 - \acc_{i} & \text{if i is a single-label task} \\
        1 - \mAP_{i} & \text{if i is a multi-label task},
    \end{cases}
\end{gather}
where $\acc_{i}$ is the average accuracy on the task $i$; $\mAP_{i}$ is the mean average precision on the task $i$.

Ultimately, these task level metrics, $(e_i, \text{FLOP}_i)$ are aggregated at the stream level via averaging or sum. Given a stream $\mathcal{S}$ with $K$ tasks, we define:
\begin{eqnarray}
    \label{eq:stream_metrics}
    \mathcal{E}(\mathcal{S}) & \defined & \frac{1}{K} \sum_{i=1}^K e_{i}\\
    \mbox{cFLOP}(\mathcal{S}) & \defined &  \sum_{i=1}^K \mbox{FLOP}_i
\end{eqnarray}
where we denote the average error rate with $\mathcal{E}$ and the cumulative floating point operations with $\mbox{cFLOP}$. By varying hyper-parameters such as the number of gradient steps used, the size of the architecture or the number of trials used during hyper-parameter search, $M$ will strike different trade-offs between  average error rate and total compute.  The search process over $\lambda^M$ will aim at pushing the Pareto front of these points. In our work (as it is standard practice), the process of searching over $\lambda^M$ requires a human in the loop to determine the best meta-learner's configuration  to explore next
\footnote{\jb{
We focus on the Pareto front because we are faced with multi-objective optimization: for each meta-learner $M$ and (meta-learner) hyperparameter
setting we obtain a predictive performance $\mathcal{E}$ and $\mbox{cFLOP}$. The Pareto front illustrates 
the best attainable performance for each compute budget.
}}.

Note that one may choose to train on all tasks at once, or to pick only the largest one. Yet the meta-training loop is still used to find $\lambda^M$. Researchers are however free to explore other ways to cross-validate $M$, for instance by introducing a meta-validation stream. In this work, we opted for simplicity, but we leave open the question of how to best cross-validate $M$ and hope that \minerva{} can be useful also to address this research question.

\subsection{\mr{Meta-testing Phase}} \label{sec:meta-test}
Once $\lambda^M$ is determined, we evaluate $M$ (lines 21 to 30). The process follows the same steps discussed previously, with a few exceptions. First, $M$ must do a pass over the \emph{entire} stream, $\mathcal{S} = \metatrainstream + \metateststream$. Notice that at step $i$ $M$ cannot access any task $\task_j$ with $j > i$; in particular, since there is no outer loop over $M$'s hyper-parameters, tasks in $\metateststream$ are observed once and only once in sequence. During training on task $\task_i$, $P_i$ can do several epochs over $\mathcal{D}_i^{\mbox{tr}}$, but $M$ cannot revisit $\task_i$ twice because otherwise we would not be able to assess generalization to unseen tasks of $\metateststream$.
The second difference is that $P_i$ is evaluated on  $\mathcal{D}_i^{\mbox{ts}}$. 
Finally, the average error rate is calculated only using tasks belonging to $\metateststream$. 
The rationale is to remove tasks over which we did any kind of cross-validation to prevent overfitting. However, we still account for the cost of development of $M$ by including the FLOPS used while learning on $\metatrainstream$ (although for the sake of simplicity, we do not consider how many times  $\metatrainstream$ was visited during the meta-training phase).\footnote{Using a separate set of test tasks, $\metateststream$, and test splits for the datasets, $\mathcal{D}^{\mbox{ts}}$, might seem overly zealous. This was however useful to assess potential overfitting to $\metatrainstream$, and for some of our ablations which analyzed the full stream $\mathcal{S} = \metatrainstream + \metateststream$. For instance, the ablation by domain of Sec.~\ref{sec:ablation} required the analysis on the full stream.}

Ultimately a given method will yield a tuple, $(\mathcal{E}, \mbox{cFLOP})$, where $\mathcal{E}$ is the average error over $\metateststream$ and $\mbox{cFLOP}$ is the cumulative compute over the whole stream $\mathcal{S}$. 
%By varying hyper-parameters like the number of gradient update steps or the size of the network, the model will strike different trade-offs. 
The best methods will be the ones at the Pareto front, delivering the lowest $\mathcal{E}$ for a given total compute budget. A clever method could early stop on tasks whenever its gains in terms of error rate are too marginal, and use the saved compute on future tasks that do require more compute to achieve lower error rate.\footnote{\mr{While we recommend to rank methods by reporting  pareto-fronts of compute versus error rate, section~\ref{sec:ablation} reports additional  metrics which provide finer grained understanding of strengths and weakness of different methods; for instance, we consider slicing results by domain, dataset size and we also report forward transfer on datasets appearing more than once.}}

\subsection{Computational requirements and the \SHORT \; stream}

\jb{
Depending on algorithm, hyperparameters and available hardware, a single run on the benchmark can often take 
multiple days even on machines equipped with 16 A100 NVIDIA GPUs.} \mr{In Appendix~\ref{app:cheapo} we report experiments with cheaper computational budgets that could run for a few days on a single GPU device.}

\jb{To further facilitate quick experimentation and to ensure researchers with limited access to compute resources can contribute, 
we derive a \SHORT\; stream by randomly selecting only two datasets per year from the original list of publications;
but otherwise following the same stream creation procedure. We obtain a stream with 24 tasks in total. 
Table \ref{app:ds_short_list} lists the datasets and their chronological order. The majority of learning 
algorithms described in this study finish in under 24h on this stream when running on 16 A100 GPUs.} \mr{We used \SHORT for all our initial experiments.}

\subsection{Codebase} \label{sec:codebase}
An open source implementation of the \minerva{} benchmark has been made available at \codelink.
This repository implements the training and evaluation protocol described in this section, downloads and prepares the data, and enables other researchers to reproduce the main results we obtained with our baselines.

The implementation has been designed to be modular, compartmentalizing data processing, handling of the stream, learners and metrics.
In particular, no knowledge of the particular stream or metrics is needed in order to implement a new learner. Moreover, the learner interface is minimal, requiring only the implementation of functions that initialize, train, and compute predictions. This may be implemented using any appropriate Python based machine learning library such as JAX or PyTorch.

\section{Experiments}
In this section we describe the baseline approaches and results we obtained on the \minerva{} stream. In Sec.~\ref{sec:ablation} we conclude  with ablations showing what factors contribute to the performance of current approaches, and how the diversity and scale of \minerva{} enable better assessment of life-long learners. 

\paragraph{Preprocessing and Data Augmentation}
The tasks in \minerva{} have images spanning a wide range of different resolutions. Even within a task the resolution may vary from image to image.
For our baselines, we adopted a two-part strategy which favored simplicity over performance. During training, images are randomly resized and cropped to a fixed resolution of $64\times64$ pixels, and then  left-right flipped with probability $0.5$.
During evaluation, we take the central square crop of $\min(w,h)\times \min(w,h)$, where $w$ is the width of the image and $h$ is its height, and resize it to $64\times64$ pixels with no additional augmentation.

Note that this strategy is clearly sub-optimal for tasks involving fine details, such as crowd counting. Nonetheless, this choice simplifies the design of architectures. More details can be found in the sensitivity analysis of Fig.~\ref{fig:indep_resolution_sensitivity}.

\paragraph{Architecture}
Unless otherwise stated, our baselines use the ResNet34 backbone tailored for low resolution images~\citep[Sec. 4.2]{he2016deep}, since all images are resized to $64\times64$ pixels. Each task is assigned a task-specific {\it head} mapping backbone features to output logits. 

\paragraph{Meta-Learning}
Each of our baselines  includes a {\it meta-learner} that selects the task-specific hyper-parameters used during training of the actual predictor. For the sake of speeding up tuning of the meta-learner, we performed stream-level cross-validation on \jb{the \SHORT stream, containing $24$ tasks only. We }
tuned the choice of the architecture and the ranges used by the hyper-parameter search. 
In particular, we identified a set of hyper-parameters that are robust enough to be kept fixed throughout the learning experience on the stream: 1) cosine learning rate scheduling with warm-up phase proportional to the number of gradients updates in conjunction with SGD with Nesterov momentum (set to $0.9$, with a weight decay of $0.0001$), 2) data augmentation consisting of random cropping and flipping, and 3) a heuristic to set the batch size as a function of the dataset size,
\begin{equation}
    \label{eq:adaptive_bsz}
    b = \min \left(
        B,
        \max \left(
            16,
            2^{
                \lfloor
                    \log_2 p \cdot D
                \rfloor
            }
        \right)
    \right),
\end{equation}
where $B$ is the maximum batch size, $D$ is the dataset size, and $p$ is a constant set to $0.0025$ in our experiments.
%The final evaluation results are then presented on the entire  stream, evaluating on the test split of each task using the metrics discussed in Sec.~\ref{sec:stream_metrics}.

\paragraph{Learning} \label{sec:baselines}

%--------------------------------------------------------------------------------------%
% \begin{table}[tb]
%     \begin{center}
%         \begin{tabular}{p{0.15\linewidth}||p{0.2\linewidth}|p{0.2\linewidth}|p{0.2\linewidth}}
%             {\bf Baseline} & {\bf initialization} & {\bf training data} & {\bf h.p. search} \\
%             \hline
%             \hline
%             Independent (Indep) & random & current task & random search \\
%             \hline
%             Fine-tuning (FT) & from a previous task & current task & random search \\
%             \hline
%             Multi-tasking (MT) & from a previous task & multiple tasks & random search \\
%             \hline
%             Pre-training (PT) & from pre-trained model & current task & random search \\
%             \hline
%             PT + FT & from a previous task or a pretrained model & current task & random search \\
%             \hline
%             \art{Bayesian hyperparameter optimization (BHPO)} & random & current task & \art{Gaussian process with upper confidence bound (GP-UCB)} \\
%             %\hline
%         \end{tabular}
%     \end{center}
% \caption{Differentiating baselines across three axes: 1) How the parameter's of the predictor are initialized, 2) What data is used to train the predictor and 3) How hyper-parameter search of each predictor is conducted.}
% \label{tab:metalearners_3fac}
% \end{table}
\begin{table}[tb]
    \begin{center}
        \begin{tabular}{p{0.15\linewidth}p{0.3\linewidth}p{0.2\linewidth}p{0.2\linewidth}}
            \toprule
            {\bf Baseline} & {\bf initialization} & {\bf training data} & {\bf h.p. search} \\
            \midrule
            Independent (Indep) & random & current task & random search \\
            Fine-tuning (FT) & from a previous task & current task & random search \\
            Multi-tasking (MT) & from a previous task & multiple tasks & random search \\
            Pre-training (PT) & from pre-trained model & current task & random search \\
            PT + FT & from a previous task or a pretrained model & current task & random search \\
            \art{Bayesian hyperparameter optimization (BHPO)} & random & current task & \art{Gaussian process with upper confidence bound (GP-UCB)} \\
            \bottomrule
        \end{tabular}
    \end{center}
\caption{Differentiating baselines across three axes: 1) How the parameter's of the predictor are initialized, 2) What data is used to train the predictor and 3) How hyper-parameter search of each predictor is conducted.}
\label{tab:metalearners_3fac}
\end{table}
All baselines we have considered in our empirical evaluation yield one predictor per task, without any parameter sharing across tasks. This is the simplest setting which corresponds to the most widely used design choice in practice.
 There are three independent factors that are used to create each baselines: 1) The choice of initialization, 2) The choice of which data is used to train a given predictor, and 3) The choice of the algorithm used to search in hyper-parameter space. The particular combination of three factors determines the meta-learner $M$ described in Sec.~\ref{sec:protocol}. In this work we have considered the six most widely used combinations of these three factors, as summarized in Tab.~\ref{tab:metalearners_3fac}, namely:

\paragraph{1) Independent (Indep):} The meta-learner initializes the parameters of the predictor for each task at {\em random}, trains using data for the {\em current} task only, and searches over hyper-parameters using {\em random search}. 
This is the the most na\"ive baseline. It is the reference that any other method should beat, as it is the simplest method that does not support any form of transfer learning (the state of the meta-learner $s_k$ is null for all $k$). 

\paragraph{2) Finetuning (FT):} The meta-learner initializes the parameters of the predictor from the parameters of a network trained on {\em a previous task}, trains using data for the current task only, and searches over hyper-parameters using random search. 
In this case the state of the meta-learner $s_k$  consists of the union of the model parameters trained on all tasks observed so far, from $1$ till $k-1$. Knowledge for this learner is the set of model parameters, which correspond to one expert per observed task.

We have considered various criteria to select the previous task from which to finetune: 1) temporal proximity by taking the most recent task (FT-prev), 2) task relatedness by taking the most related past task. For the latter, we have been using as a proxy of task relatedness the performance of a k-nearest neighbor classifier in the feature space produced by the previous predictors, using as training and validation data a small subset of $\mathcal{D}^{tr}$ (up to $10000$ and $5000$ images, respectively). In other words, we use the previous predictors as feature extractors to encode data from the current task, similarly to~\cite{veniat2021efficient}. We have two versions of this. An offline or static version (FT-s) where the features are computed using pretrained independent predictors as in Indep above, and a dynamic version (FT-d) where features are computed online using the actual predictors trained so far (which could have been finetuned themselves on other tasks). Fig.~\ref{fig:ft-d} in Appendix shows an actual example of such learned chain of finetuned models.

\paragraph{3) Multitasking (MT):} 
 The meta-learner initializes the parameters of the predictor using the same approach as in FT-d, trains using data of {\em both the current task and some previous tasks}, and searches over hyper-parameters using random search. Both the selection criterion for what previous task to take for parameter initialization and what previous (auxiliary) tasks to take for additional training data are based on task relatedness using the same k-nearest neighbor classifier score described in the FT-d baseline above. Unlike FT, the multitask baseline can take $k\ge1$ most related auxiliary tasks for additional training data. During training the network is trained in a multitask fashion, weighing the losses of the auxiliary tasks by a single scalar hyper-parameter $\lambda$. This hyper-parameter is subject to hyper-parameter search by the meta-learner, and it sets the relative importance of the auxiliary tasks against the current task. In a single training step, the gradients are accumulated across the mini-batches of current and $k$ auxiliary tasks.
 At test time, the classification heads of the auxiliary tasks are disregarded. With reference to Algorithm~\ref{alg:protocol}, the state $s_k$ of the meta-learner $M$ consists of the set of datasets and predictors trained so far. In this case, knowledge is represented both as the set of model parameters of already observed tasks, as well as the set of datasets encountered so far.
 
\paragraph{4) Pre-training (PT):} The meta-learner initializes the parameters from a {\em pretrained} model, trains using data for the current task only, and searches over hyper-parameters using random search. This is a special case of FT, where all task predictors are finetuned from exactly the same pretrained model. Note that there is no form of knowledge accumulation for this baseline. 

We have considered two pretrained models from which to finetune: 1) A ResNet34 pretrained on ImageNet by supervised learning (PT-ISup) and 2) A Normalizer-Free network (NFNet-F0)~\citep{brock2021high} pretrained on two very large external datasets: ALIGN and 
LTIP~\citep{Flamingo} using CLIP (PT-ext)~\citep{radford2021learning}.
In this case, the state $s_k$ of the meta-learner $M$ is constant over time, as it  merely contains the fixed set of pre-trained parameters.

\paragraph{5) Fine-Tuning with Pre-training (PT + FT):} This variant is exactly the same as FT-d, except that the set of parameters avaiable for finetuning includes not only the parameters of networks trained on previous tasks but also the pretrained model used by PT-ext.  

\paragraph{6) Bayesian Hyper-Parameter Optimization (BHPO):} The meta-learner initializes the parameters of the predictor at random, trains using data for the current task only, and searches over hyper-parameters using a {\em Gaussian Process} to estimate the function value (expected loss at convergence). Instead of running a search in parallel over a set of hyper-parameter configurations like in random search, BHPO runs the search in sequence, using the Upper Confidence Bound acquisition function \citep{srinivas2010gaussian} provided by Google Vizier \citep{vizier} to select the next configuration to search over.

\paragraph{Experimental Setup.} The search space of Indep, FT, PT and PT+FT consists of initial learning rate and label smoothing, which means that for each task we search over the values of these two hyper-parameters. MT adds to the search space also $\lambda$, the weight on the auxiliary tasks. BHPO adds five additional hyper-parameters compared to Indep, which include the choice of learning rate schedule (cosine learning rate, piece-wise constant decay), batch size, choice of architecture (VGG, ResNet34), choice of the two data augmentations (random cropping and flipping).

To vary the compute budget and study the Pareto front of average error rate versus compute, and unless otherwise stated, we vary  the number of hyper-parameter configurations over which the meta-learner searches over at each task (ranging from $2$ to $32$), and the total number of weight updates (ranging from $10000$ to $100000$). Note that different combinations of number of updates and trials per task can lead to the same computational budget, but different performance. 

In the following section we will report results for a total compute budget in the range between $10^{18}$ and $10^{21}$, aiming at models that achieve competitive final performance (for the chosen $64 \times 64$ resolution) on landmark datasets. For instance, the basic Indep baseline achieves 4.2\% error rate on SVHN, 0.7\% on MNIST, 6.8\% on CIFAR10 and 34\% on ImageNet.  In Appendix~\ref{app:cheapo}, we will report results using smaller computational budgets (and hence worse final error rate) for users that have more limited computational resources at their disposal.

\subsection{Findings}\label{sec:main_results}
\begin{figure}[t]
  \centering
  \includegraphics[width=.99\textwidth]{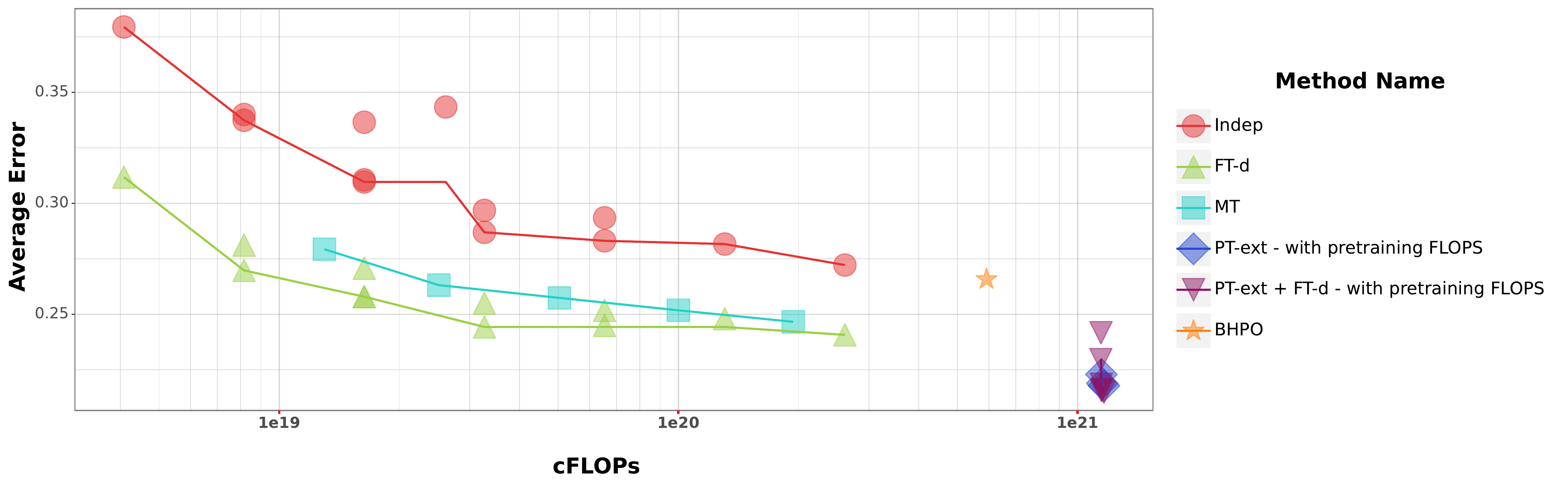}\\
   \includegraphics[width=.99\textwidth]{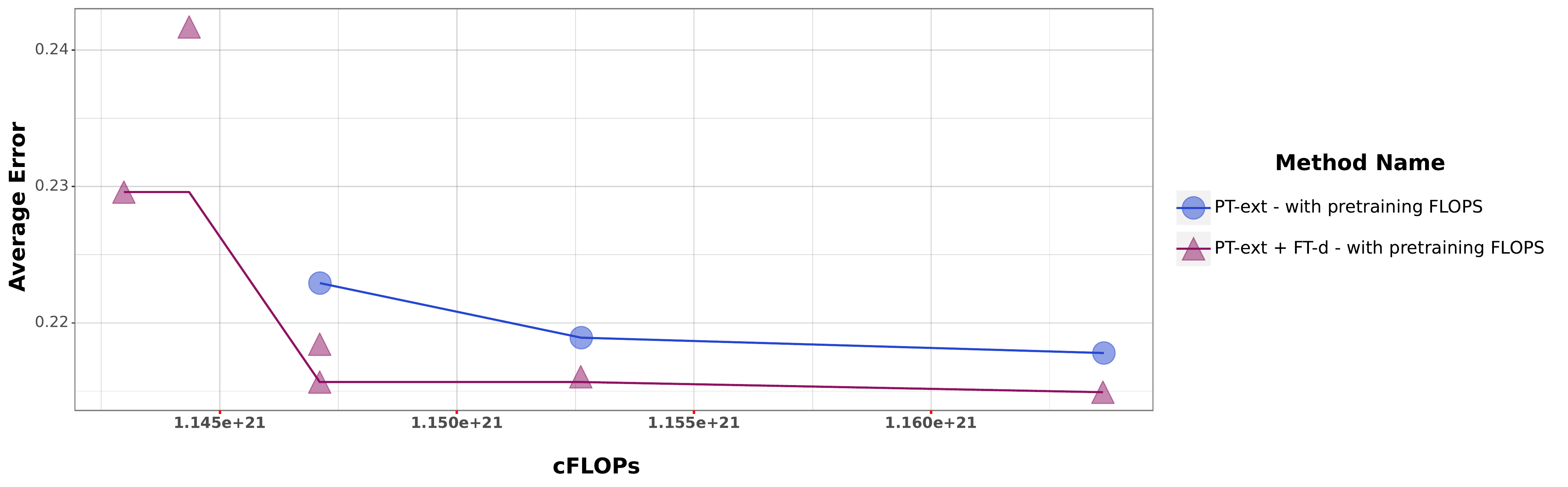}
  \caption{\textbf{Pareto fronts}: Each marker shows the average error rate on $\metateststream$ and the total cFLOP on the entire stream (see Sec.~\ref{sec:protocol}). Since there are \numtasks{} tasks, if each task required $16$ hyper-parameter configurations, a marker is the result of $1696$ experiments. Pareto fronts are created by varying the number of hyper-parameter configurations and the number of gradients steps used to train on each task. The top panel shows a selected baseline from each of the baseline families described in Sec.~\ref{sec:baselines}, and the bottom panel zooms into the pretraining baselines using external data.
  %The point with error bars refers to Indep, where error bars are computed using different seeds for the network parameters. It shows that the gains brought by methods that transfer are statistically significant. %\mr{What does STD mean? } \mr{Comment on cost of pretraining which is currently included.}\yutian{MR: I added full stream BHPO results in Figure 4 and 5. The original plots are commented out. Please have a look if it's useful to include it or not. If not, just revert to the original plots.}
  }
  \label{fig:full_stream-pareto_fronts}
\end{figure}

In this section we report the main results we obtained by applying the previously described baselines to \minerva. A priori, common wisdom would suggest that methods that perform multi-task using all available data would perform the best, while methods that rely on sequential finetuning to grossly underperform~\citep{ash2020}. We would also expect methods based on pretraining to perform the best, but to gain little if anything by combining with other forms of transfer learning as their representations are potentially already general enough. We would also expect Indep to work more poorly on smaller datasets, and that no method is best across the entire spectrum of compute budget. 
Our results show that not all these intuitions find empirical support in \minerva.

We start by reporting the pareto fronts of average error rate versus compute in Fig.~\ref{fig:full_stream-pareto_fronts}. In this figure, we show a selected baseline from each of the families described in Sec.~\ref{sec:baselines}. More results are reported in the Appendix (Sec.~\ref{apx:baselines}). 

\begin{enumerate}
    \item We observe that all methods perform better than learning from scratch (Indep). This shows that indeed there is rich shared structure across tasks in \minerva{} which supports  beneficial transfer. Note that a run with 3 different seeds of the independent baseline (at the second highest cost) yields a standard deviation of $0.003$, suggesting that performance gaps between baselines are significant except for FT-d and MT at the highest computational budget.
    \item BHPO reduces the error compared to Indep, but at the cost of more compute. Overall, BHPO does not improve the pareto front. Given that hyper-parameter search substantially impacts cFLOP (typically by a factor of ten or more), we surmise that there could be more clever and  efficient ways to explore the hyper-parameter space which could lead to a better performance-compute trade-off. 
    \item There is a rather large gap of about 5\% absolute error between the best and the worst method, namely Indep and FT-d if we restrict to training data from \minerva. Given the simplicity of FT-d in terms of its approach to transfer learning and its na\"ive use of compute (random hyper-parameter search and no early stopping), we would expect future approaches to further reduce both compute and error rate.
    \item The performance of FT-d suggests that sequential finetuning, even when applied on chains longer than $10$ steps, as it can be appreciated in Fig.~\ref{fig:ft-d} of Appendix, works remarkably well.
    \item The choice of what to finetune from matters, as FT-prev is significantly worse than FT-d as shown in the Appendix (Sec.~\ref{apx:finetuning}). Therefore, there could be improvements by using better approaches to estimate task relatedness. 
    \item MT works well but the additional compute spent on relearning representations on past tasks does not compensate for the improvements in generalization. Overall, FT-d strikes a better trade-off than MT.
    \item As expected, pretraining improves the performance significantly. Starting with a pretrained network (PT-ext) leads to a significantly lower average error. Notice that PT-ext leverages both a much larger amount of external data and a more powerful architecture. In the Appendix~\ref{apx:pretraining} we also show that pretraining the same architecture used for the other baselines on ImageNet (PT-ISup) reduces the gap between Indep and the best performing baselines (FT-d and MT). 
    \item More surprisingly, using FT-d with a pretrained network (PT-ext + FT-d) lowers the average error further, see bottom plot of Fig.~\ref{fig:full_stream-pareto_fronts}. This demonstrates that leveraging the structure in the stream can improve the already general representations that the pretrained model provides, which opens a new avenue of research on large-scale models that continuously adapt over time. More details on the structure that FT-d  discovers starting from the pretrained model are provided in the Appendix~\ref{apx:pt-ft}.
\end{enumerate}

\begin{figure}[t]
  \centering
    \includegraphics[width=.99\textwidth]{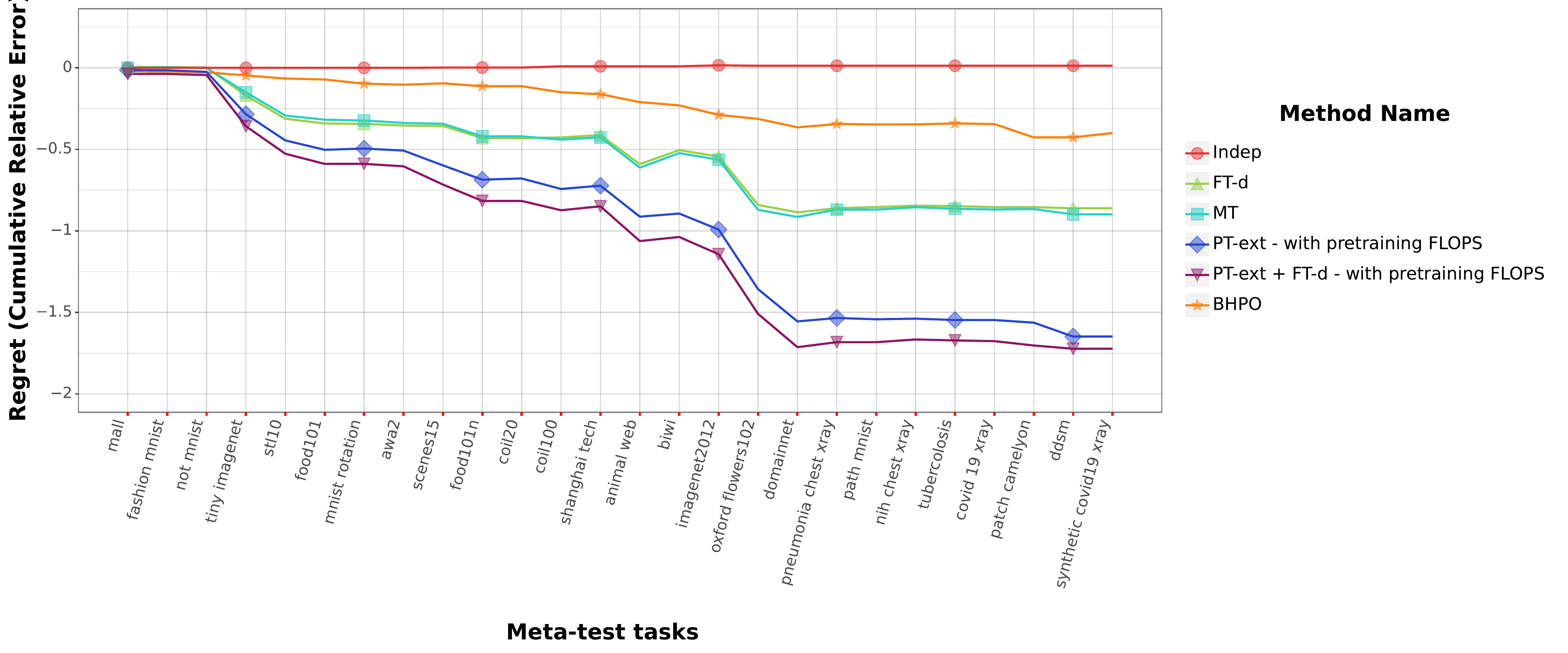}
    \includegraphics[width=.99\textwidth]{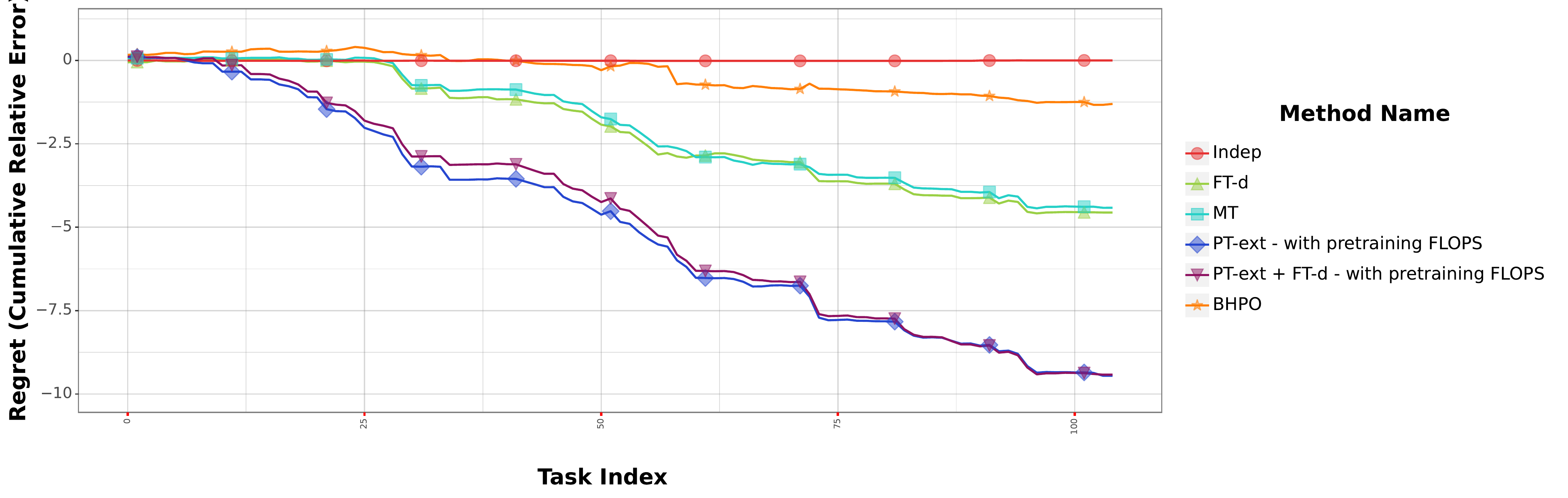}
  \caption{\textbf{Regret plots}: Cumulative error rate relative to Indep. on $\metateststream$ (top) and on the full stream (bottom). 
  % \mr{Can we use the reference as Indep?}
    %\mr{Possible to change line style for different curves?} \art{We should precise that PT-ext and BHPO correspond to a higher compute budget than the other baselines.}
 }
  \label{fig:regret_plots}
\end{figure}

In order to better understand how methods fare on each task, we also present a regret-like plot in Fig.~\ref{fig:regret_plots}. This shows the cumulative error over time relative to Indep, picking hyper-parameters such that all methods use roughly the same amount of compute. When the curve is horizontal it means that a method performs comparably to Indep. When the slope is negative it means that it outperforms Indep, and vice versa.
We observe that no method, including PT variants, transfers well to datasets in the OCR and medical domains. In particular, all regret curves are nearly horizontal over the last nine datasets, which are mostly datasets in a new x-ray domain (Covid-19 related classification tasks from 2021). This shows a clear limitation of current approaches which are not yet capable to a) transfer well to minor domains, and b) accrue knowledge over time, as many of these $9$ last tasks are closely related to each other. 

Finally, the regret plot over the entire stream in the bottom of  Fig.~\ref{fig:regret_plots}  shows that overall methods provide a linear improvement over the Indep baseline. FT-d starts flat as expected (since initially there is nothing to transfer) and later on exhibits a linear gain. However, no method improves over time in the second part of the stream. In other words, none of the baselines we tried was capable to become more accurate as it receives more data and it makes new learning experiences. While this is expected for PT which cannot accrue knowledge by construction, we surmise there could be a more clever variant of FT that actually improves over time.

Overall, these findings indicate that \minerva{} is a good playground for research in never-ending learning. Methods that transfer do significantly better than methods that do not, and yet there seems to be ample room for improvement over the current set of baselines. Unlike our expectations, methods based on multi-tasking did not yield better trade-offs, they might achieve a lower error rate but this does not compensate for the increase in the amount of compute. Instead, sequential FT has worked remarkably well despite the simplicity of the heuristic used to determine task relatedness. This observation still holds when starting from a large pretrained model, opening the question of how to best accumulate knowledge in foundational models.
Finally, we have reported more results using BHPO in Appendix~\ref{app:bhpo}. These show that current BHPO is effective only when increasing the search space. In this setting, BHPO does find better hyper-parameter configurations than random search but the gains are currently too limited relative to the additional compute required.

\subsection{Ablations} \label{sec:ablation}
\begin{figure}[t]
  \centering
  \includegraphics[width=.99\textwidth]{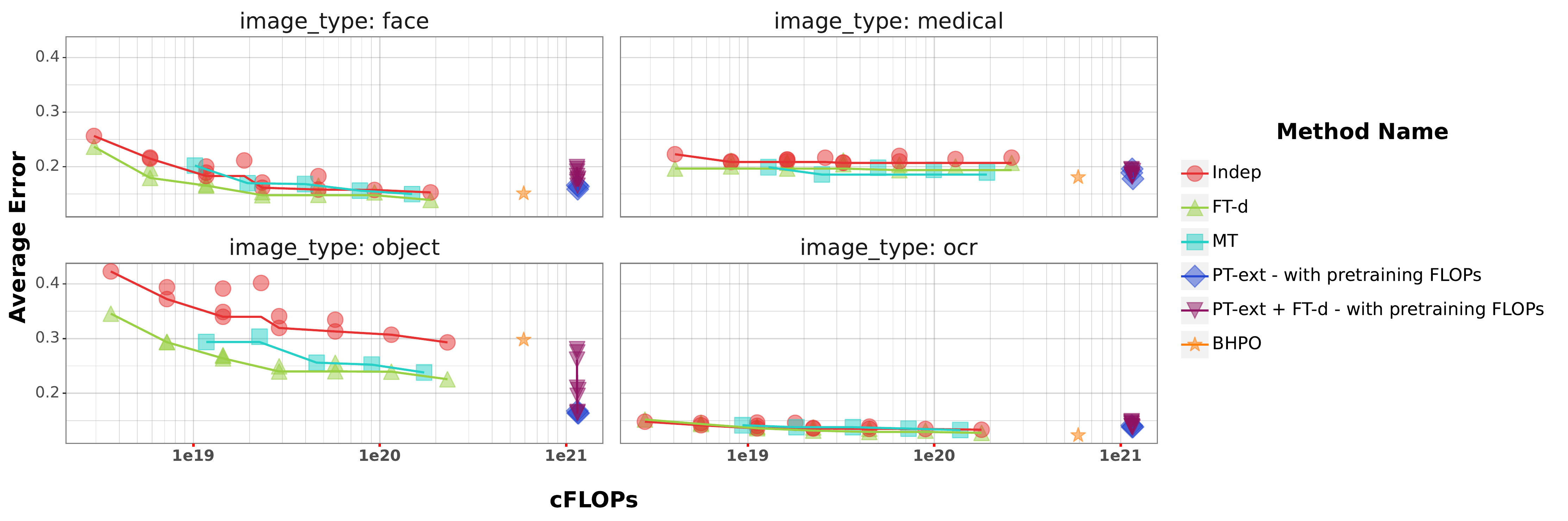}
  \caption{Analysis by domain. Each sub-plot corresponds to a different domain.}
  \label{fig:full_stream-per-domain-ablation}
\end{figure}
In this section we further analyze \minerva{}, studying how results vary by domain, dataset size, task ordering, etc. The goal is to understand which factors affect the performance of the baselines the most, and ultimately, which unique features \minerva{} has to offer relative to existing benchmarks.

\paragraph{Image Domain.} In this experiment, we take baselines which have been trained on the entire stream, and instead of evaluating on all tasks of $\metateststream$, we evaluate on all tasks of $\metatrainstream \cup \metateststream$ but filtering tasks by their domain. For the evaluation we use the test split of each task. We report results on $4$ representative domains, namely OCR, medical, face and object. 
Note that in methods like FT, networks that are trained on a dataset of a certain domain, could be finetuned from a task belonging to a different domain. Moreover, results across domains are not directly comparable since each domain has its own set of tasks. 

Results are shown in Figure~\ref{fig:full_stream-per-domain-ablation}.
We observe a strong dependence on the domain type. The gains over Indep brought by baselines that allow transfer learning are minimal in OCR but substantial for object, for instance. Even more interestingly, the ranking of the various baselines is domain dependent, and there is no winner across all domains. For instance, PT is the best method in the medical domain but the worst in OCR. We conjecture that the reason could be that OCR has relatively large datasets, but it is perhaps the most distant domain relative to the domain used for pretraining. Notice how these insights have been enabled thanks to the richness of domains in \minerva.

\begin{figure}[t]
  \centering
  \includegraphics[width=.99\textwidth]{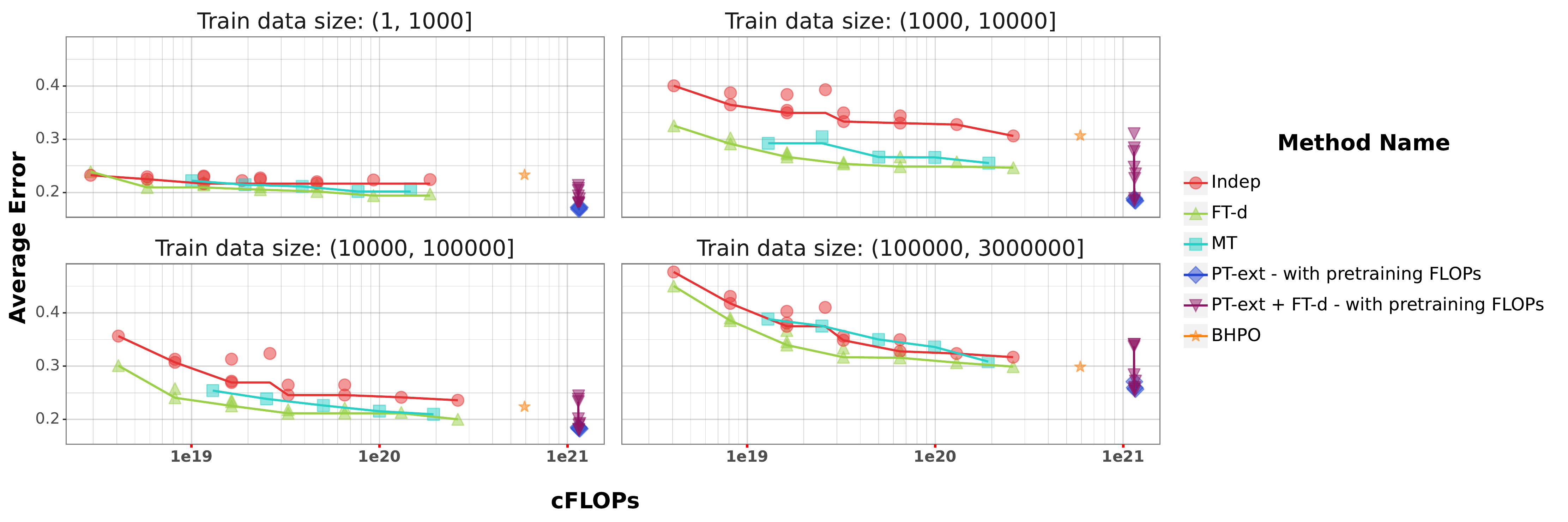}
  \caption{Analysis by dataset size. Each subplot contains the evaluation on datasets of the size indicated on the top.}
  \label{fig:full_stream-per-domain-size}
\end{figure}

\paragraph{Dataset size.} We now study how performance correlates with dataset  size. We use the same methodology used in the previous analysis by domain, but aggregate test results filtering by the size of the datasets. We have defined four groups of increasing size of the training set, namely datasets with a training set with fewer than $1{\small,}000$ examples, datasets with a number of training examples between $1{\small,}000$ and $10{\small,}000$, datasets with a number of training examples between $10{\small,}000$ and $100{\small,}000$ and datasets with more than $100{\small,}000$ examples. Results are shown in Fig.~\ref{fig:full_stream-per-domain-size}. Unsurprisingly, Indep becomes competitive on very large datasets, but surprisingly, methods that support transfer learning (e.g., FT) do not gain much over Indep on very small datasets. The gains are more significant on datasets of intermediate size. Overall, adapting to small datasets is still a challenge for the baselines we have considered. Once again, \minerva{} has enabled this analysis thanks to its diversity of dataset sizes. 

\begin{figure}[t]
  \centering
  \includegraphics[width=.99\textwidth]{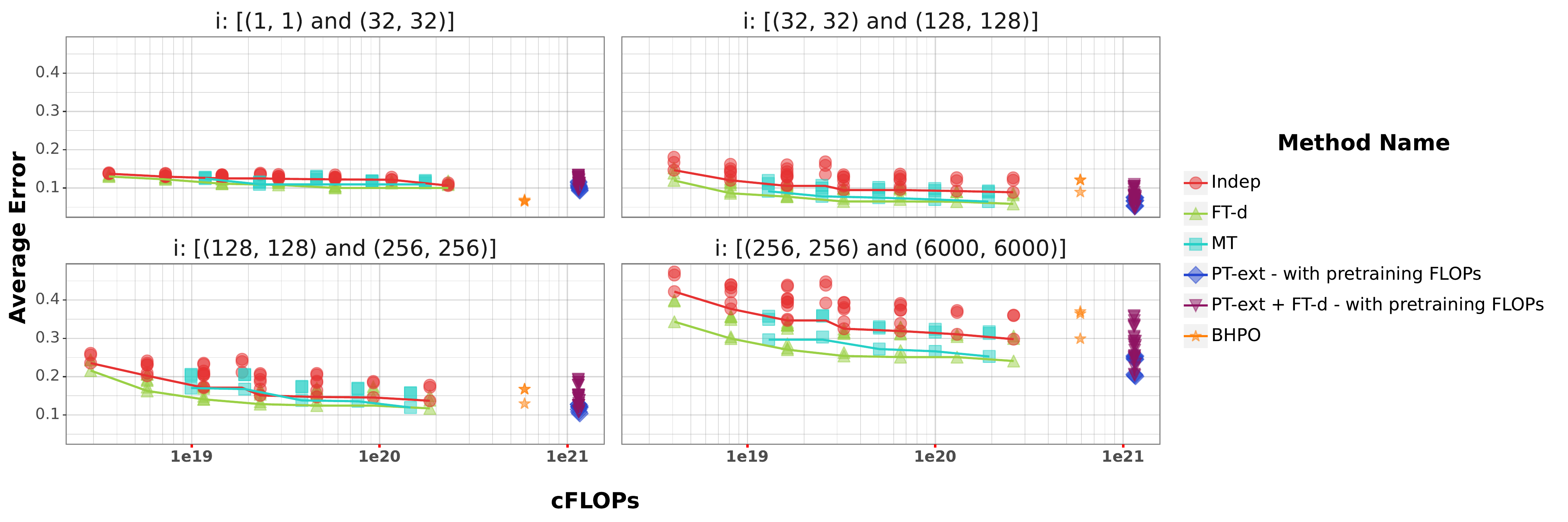}
  \caption{Analysis by the average image resolution of each task. We train on images resized to $64\times64$ pixels but evaluate by selecting tasks with original image resolution within the specified range.}
  \label{fig:full_stream-per-image-resolution}
\end{figure}
\begin{figure}[t]
  \centering
  \includegraphics[width=0.9\textwidth]{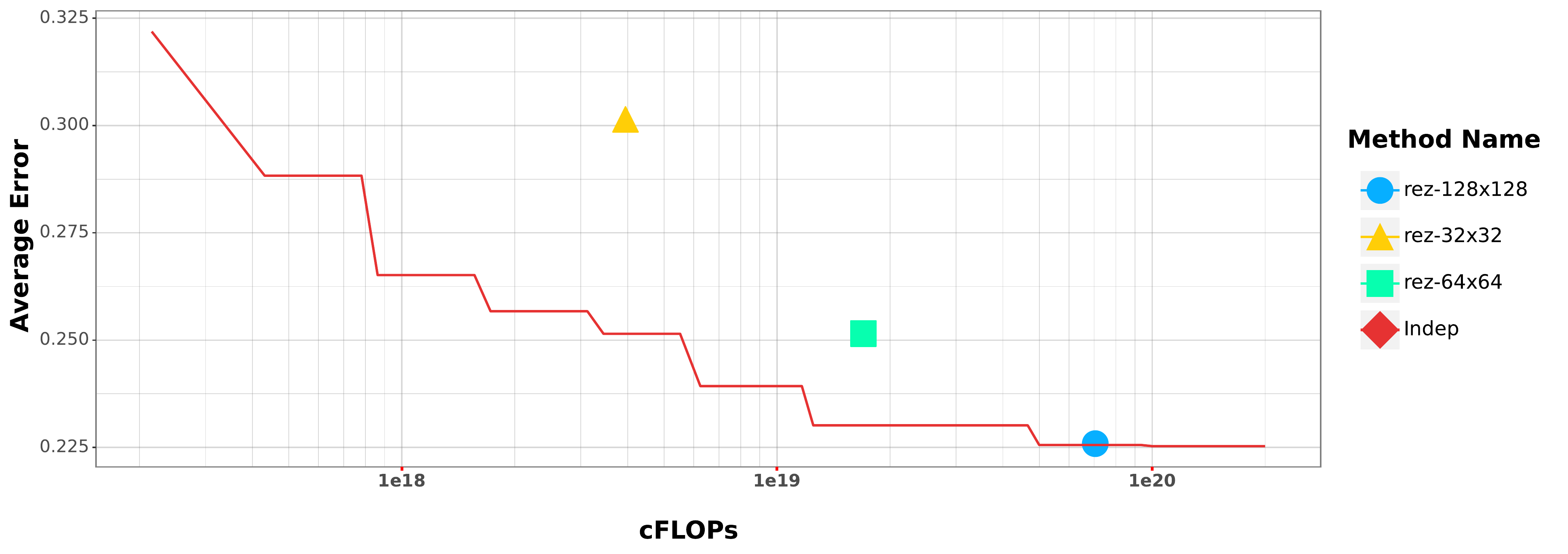}
\caption{Performance of Indep when varying the image resolution  %\jb{\sout{{\em during}} for} 
training and evaluation using ResNet34 with a maximum batch size of $128$, $50{\small,}000$ gradient updates, and $16$ trials per task. The purple line is the Pareto front of the default Indep baseline trained on images of size $64\times 64$ using different models, a maximum batch size of 512, and various combinations of number of updates and trials per task.}
  \label{fig:indep_resolution_sensitivity}
\end{figure}
\paragraph{Image resolution.} With a methodology similar to the one used in the previous experiments, we again train on the full stream (using the default fixed resolution of $64 \times 64$ pixels), evaluate on the test split of each task but report metrics by selecting only tasks with average image resolution within a certain range. Fig.~\ref{fig:full_stream-per-image-resolution} shows that the error rate on datasets with smaller resolution is very low, and on those datasets methods that transfer work comparably to Indep. As the resolution increases we observe a remarkable improvement of FT, PT and MT over Indep. Finally, while we cannot directly compare results across various image resolutions (since each contain a different subset of tasks), we notice that the average error rate on datasets of larger resolution images is much bigger, suggesting that downsizing the resolution to only $64 \times 64$ pixels might deteriorate performance on such tasks, and that architectures that handle variable resolution images might strike better trade-offs.

Alternatively, we also studied how the performance of Indep changes as we vary the choice of the input image resolution used during training and evaluation of each task. This was chosen to be $64 \times 64$ by default in our previous experiments. For the experiment of Fig.~\ref{fig:indep_resolution_sensitivity}, we tried two other spatial resolutions, namely $32 \times 32$ and $128 \times 128$. Note that this experiment is conducted on the short version of the stream. The purple line represents the Pareto front of the Indep baseline, trained on the default image resolution ($64\times 64$) and default maximum batch size ($512$), and varying the model architecture, the number of updates and the number of hyper-parameter configurations over which we search for each task. This frontier is provided as a reference.  We can see that varying the image resolution is yet another way to trade-off error rate versus compute. It is future work to assess whether varying the image resolution or other factors like the architecture size could improve the Pareto fronts.

\paragraph{Task ordering.} 
\begin{wrapfigure}{r}{0.5\textwidth}
  \centering
  \includegraphics[width=.48\textwidth]{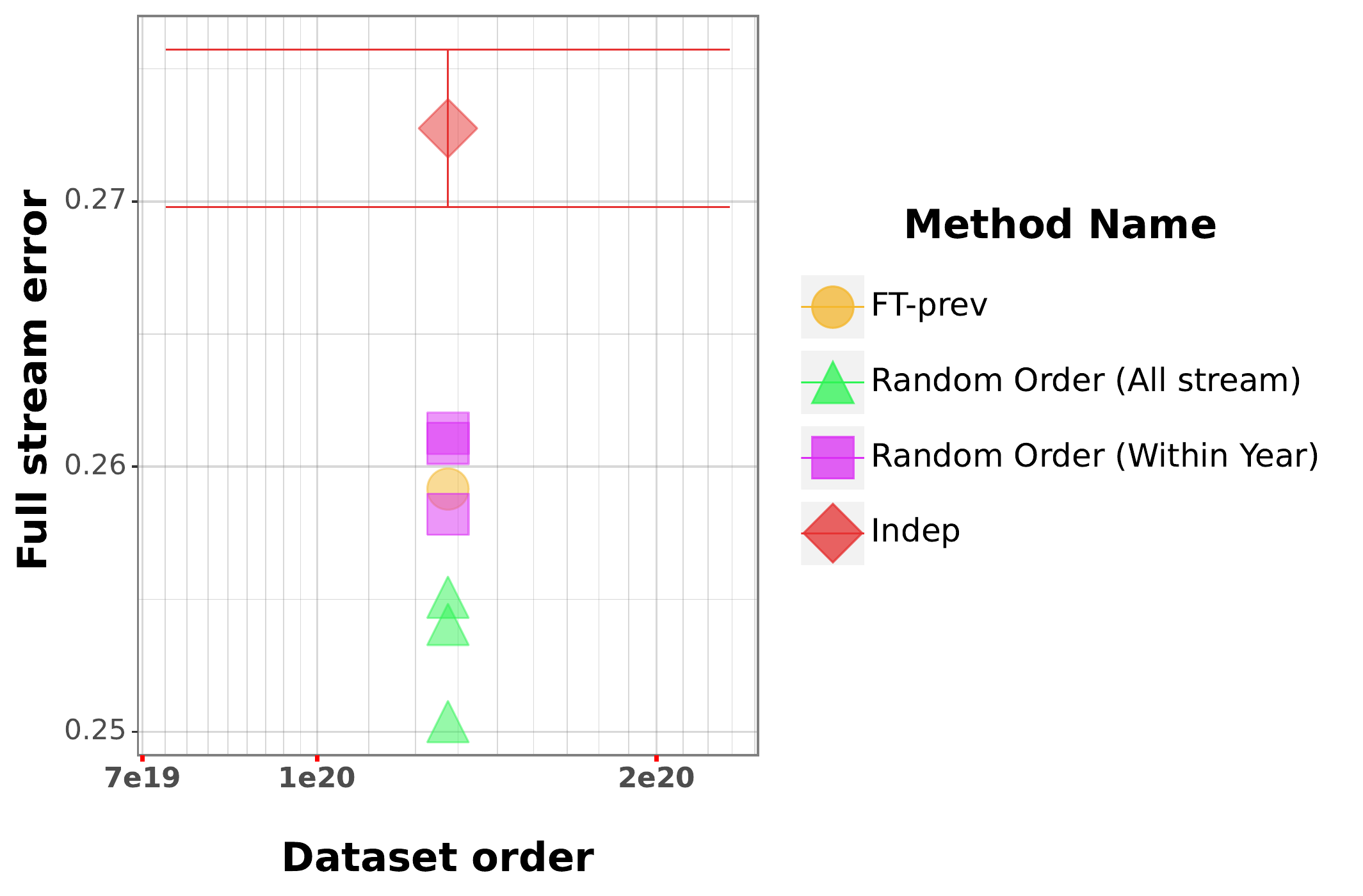}
  \caption{Effect of task ordering. Reference is Indep baseline, showing error bars when varying the seed used to initialize the networks. The other baselines are FT-prev with \art{a) the default ordering (red), b) with within year shuffling of tasks (cyan), and c) with shuffling across the entire stream (green).}}
  \label{fig:-dataset-order-ablation}
\end{wrapfigure}

In this experiment we study whether the performance of baselines is affected by the ordering of the tasks. For this purpose, we create two new variants of \minerva, one where we pick a different shuffling of the datasets within a year (note that in \minerva{} this is already arbitrary), and one where we shuffle the order of the datasets across the entire stream, $\metatrainstream \cup \metateststream$.

Results are showin in Fig.~\ref{fig:-dataset-order-ablation}. We have found that shuffling the task ordering within a year does not affect the error rate in average, meaning that if we average across several stream variants that differ in the within year shuffling we obtain a similar error rate to the default version of \minerva. 

Randomizing the order of the tasks over the {\em entire} stream has instead more dramatic effects. Whenever larger datasets like ImageNet are moved to earlier times, the average error rate is lower. The average error rate over $3$ random orderings is $0.253$, while the average error rate when shuffling withing years is $0.260$ when using FT-prev that accrues knowledge over time. Recall that the standard deviation of the Indep baseline is $0.003$  (also displayed on the figure). Overall this suggests that the order of the tasks does affect performance in \minerva, and that current baselines struggle to transfer from several smaller datasets to bigger datasets.

\iffalse
\begin{table}[t]
\centering
\begin{tabular}{|p{7cm}||c{4cm}|c{3.5cm}|}
 \hline
 \multirow{2}{8em}{\textbf{Baseline Name}} & \multicolumn{2}{|c|}{\textbf{Avg Error}}\\\cline{2-3}
 & \textbf{dev-test split} & \textbf{test split}\\
 \hline
 Full Stream   & 0.2808    & 0.2729\\
 \hline
 Full Stream excluding ImageNet2012   & 0.2904    & 0.3033\\
 \hline
 Large Datasets Only (68 Tasks)   & 0.277    & 0.2699\\
 Random 68 Tasks   & 0.3063    & 0.3035\\
 \hline
 Major Domain Datasets Only (71 Tasks) & 0.2815 & 0.2854 \\
 Random 71 Tasks & 0.3033 & 0.3058 \\
 \hline
 Remove First 30 Tasks & 0.3018 & 0.295 \\
 Remove Last 30 Tasks & 0.2723 & 0.2828 \\
 Remove Random 30 Tasks & 0.276 & 0.2671\\
 \hline
\end{tabular}
\label{tab:cripple_stream_ablation}
\caption{Performance analysis of training with dynamic finetuning strategy on the crippled streams. Keeping important datasets in the stream is suffice to obtain similar results with the full stream. On the other hand, skipping key datasets (e.g. ImageNet) could cause performance degradation. The best results are obtained when we remove the last 30 tasks and remove random 30 tasks on the dev-test split and test split respectively, indicating the importance of filtering out certain tasks.}
\end{table}
\fi

\begin{table}[t]
\centering
% \begin{tabular}{|l||c|c|}
%  \hline
%  \textbf{Baseline Name} & \textbf{Total Tasks in $\metatrainstream$} & \textbf{Avg Error} \\
%  \hline
%  Full Stream & 79  & 0.276\\
%  \hline
%  Full Stream excluding ImageNet & 78 & 0.303\\
%  \hline
%  Large Datasets Only & 40 & 0.270\\
%  Random 40 Tasks & 40 & 0.282 $\pm$ 0.01\\
%  \hline
%  Major Domain Datasets Only & 43 & 0.285 \\
%  Random 43 Tasks & 43 & 0.286 $\pm$ 0.01 \\
%  \hline
%  Remove First 30 Tasks & 49 & 0.295 \\
%  Remove Last 30 Tasks & 49 & 0.283 \\
%  Remove Random 30 Tasks & 49 & 0.283 $\pm$ 0.009\\
%  \hline
% \end{tabular}

\begin{tabular}{lll}
 \toprule
 \textbf{Baseline Name} & \textbf{Total Tasks in $\metatrainstream$} & \textbf{Avg Error} \\
 \midrule 
 Full Stream & 79  & 0.276\\
 Full Stream excluding ImageNet & 78 & 0.303\\ Large Datasets Only & 40 & 0.270\\
 Random 40 Tasks & 40 & 0.282 $\pm$ 0.01\\
 Major Domain Datasets Only & 43 & 0.285 \\
 Random 43 Tasks & 43 & 0.286 $\pm$ 0.01 \\
 Remove First 30 Tasks & 49 & 0.295 \\
 Remove Last 30 Tasks & 49 & 0.283 \\
 Remove Random 30 Tasks & 49 & 0.283 $\pm$ 0.009\\
 \hline
\end{tabular}

\caption{Average test error rate in $\metateststream$ using FT-d. Each row correspond to a different variant of  $\metatrainstream$; the first row is the default version of \minerva. Baselines of random tasks are run with 5 random seeds and we report the mean and std of average test error rate.}
\label{tab:cripple_stream_ablation}
\end{table}

\paragraph{Other Stream Variants.}
In Tab.~\ref{tab:cripple_stream_ablation} we study how tasks in $\metatrainstream$ affect performance of FT-d on $\metateststream$. We picked FT-d since this is a baseline whose performance on $\metateststream$ is expected to depend on what it has learned on $\metatrainstream$. 
The average test error using the default version of $\metatrainstream$ is $0.273$. If we remove ImageNet, the average error rate increases by $3$\%, which is not surprising as the network trained on ImageNet is selected for finetuning by several  subsequent tasks, as shown in Fig.~\ref{fig:ft-d} of Appendix. Shortening $\metatrainstream$ by selecting only tasks belonging to the major domains or larger datasets increases the error rate slightly. FT-d does not seem to leverage well smaller datasets and minor domains. Without improving the transfer ability, this baseline could in fact strike a better trade-off by removing tasks from $\metatrainstream$ as shown when removing the smallest datasets (with less than $10,000$ training examples). 

\mr{Notice that the failure of FT-d to transfer from  several small datasets to a bigger dataset is not a limitation of \minerva, but a limitation of current approaches. \minerva{} detects such deficiency and it enables the discovery of methods that might transfer also in this more difficult (but not so uncommon) condition. It is up to a method to figure out what to leverage when learning on a given task. The data and evaluation protocol provided by \minerva{} are independent from any particular modeling choice, which includes what and how to transfer from.}  

\mr{The last section of tab.~\ref{tab:cripple_stream_ablation} shows what happens when we remove $30$ tasks, either at the beginning, at the end or at random from the meta-train part of the stream.} The first $30$ tasks from the stream contain tasks that many subsequent tasks are finetuned from, including ImageNet, Caltech256, Scene8, etc. Therefore, %\mr{\sout{in this case}} 
removing the first $30$ tasks from the stream deteriorates performance \mr{the most}.  

\paragraph{Forward Transfer.}
\begin{figure}[t]
  \centering
  \includegraphics[trim={0, 0, 0, 1cm}, clip, width=.99\textwidth]{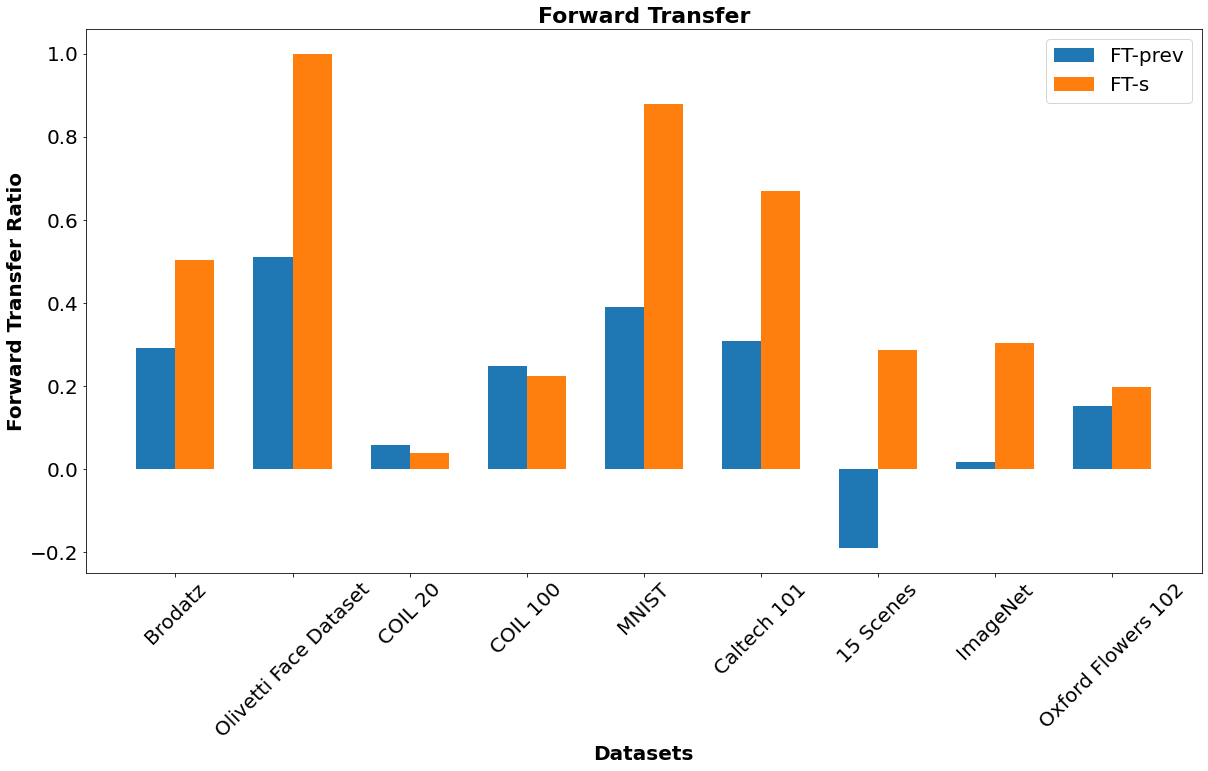}
  \caption{Forward transfer performance of two different fine-tuning strategies.}
  \label{fig:forward_transfer}
\end{figure}

An ideal never-ending learner should be able to learn faster by transferring knowledge from the past to the future. In this study we compare two fine-tuning approaches, namely FT-prev and FT-s, in terms of forward transfer. There are $9$ tasks that are presented twice in the full stream. For each of these tasks, higher forward transfer should imply faster learning when the same task is presented the second time. Notice that the learner has to figure out task relatedness even on duplicate tasks, as every task (including duplicates) is assigned a unique task id and classification head. To measure forward transfer, we adapt the metric proposed in~\citet{ContinualWorld} by computing the normalized difference between the area under the first learning curve and the area under the second learning curve:

\begin{equation}
    \label{eq:forward_transfer}
    \text{\art{FWT}}:=\frac{\text{AUC}_2-\text{AUC}_1}{1-\text{AUC}_1},
\end{equation}
where $\text{AUC}_1$ and $\text{AUC}_2$ are the areas under the accuracy curves on the evaluation dataset when the task was presented for the first and the second time, respectively.  The resulting metric is less or equal to 1, and higher value indicates better forward transfer. Fig.~\ref{fig:forward_transfer} shows that these FT learners do achieve positive forward transfer in average, with FT-s outperforming FT-prev. 
In fact, on some tasks like Olivetti and MNIST the forward transfer of FT-s approaches $1.0$. Since FT-prev finetunes from a possibly interfering task, on the 15 scenes dataset it obtains an even negative transfer, highlighting again the importance of estimating task relatedness. \footnote{\art{FT-d is omitted from this plot because its behavior in terms of transfer is closely related to FT-s. It offers a way to estimate task relatedness on the fly, and the same observations and conclusions as for FT-s hold.}}

\begin{figure}[t]
  \centering
  \includegraphics[width=.99\textwidth]{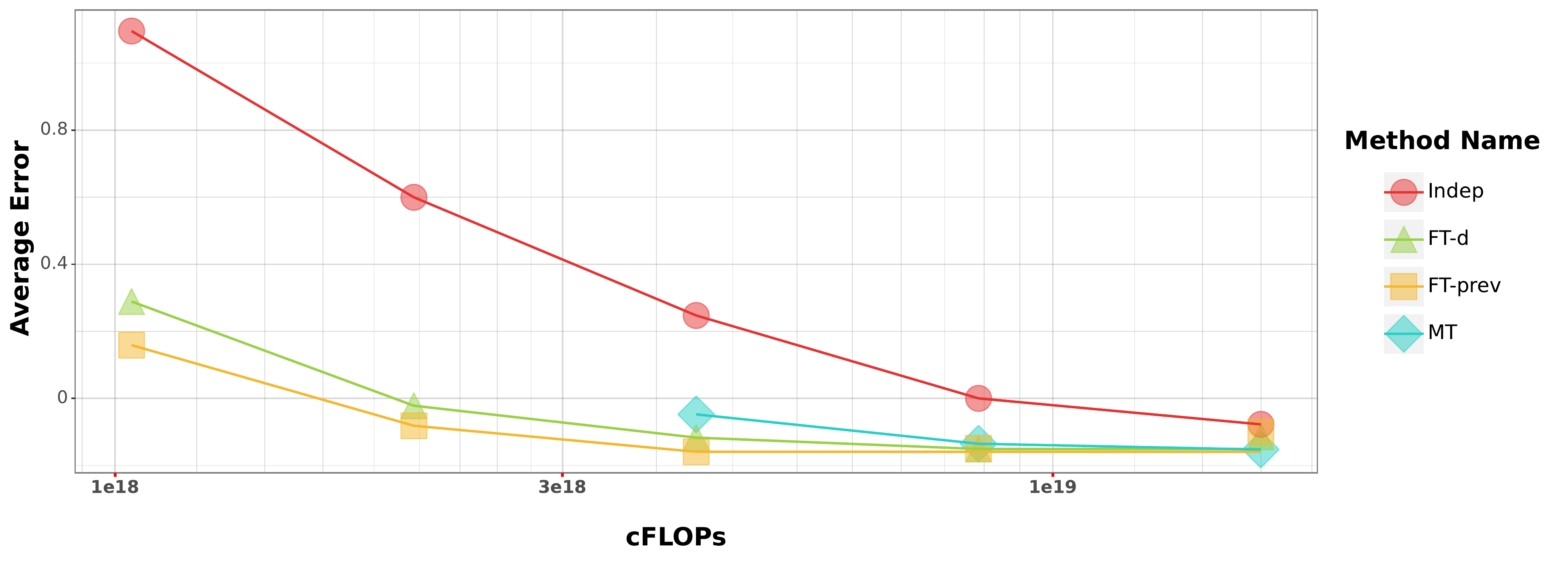}
  \caption{\art{Results on the Split-ImageNet stream with 100 tasks.}}
  \label{fig:split_imagenet_ablation}
\end{figure}

\paragraph{Split-ImageNet.} 
In our last study, we apply the same baselines and training and evaluation protocol to another stream, Split-ImageNet, \citep{rebuffi2017icarl, wu2019large}. Its much smaller variant, Split-MiniImageNet, is one of the most popular large-scale streams used in continual learning research \citep{shim2021online,mai2022online}.
Split-ImageNet is a stream derived from ImageNet, where the original $1000$ classes are partitioned into $100$ disjoint groups, creating a stream of $100$ $10$-way classification tasks. The goal of this study is to assess whether this benchmark yields similar findings as \minerva.

From Fig.~\ref{fig:split_imagenet_ablation}, it can be seen that the difference between the approaches dramatically reduces as the computation budget increases. Moreover, the ranking of the approaches is rather different. On Split-ImageNet FT-prev performs the best. This is not surprising since tasks in Split-ImageNet are highly related and very homogeneous. This however is an artifact of the lack of diversity of the Split-ImageNet stream, and it highlights the usefulness of the proposed \minerva{} benchmark.

\section{Ethical Considerations}
A potential concern about the methodology used to build \minerva{} is the use of relatively old datasets that might suffer even more greatly from issues that our community only recently has started to analyze.
For example, face datasets have been found to lack demographic representation across characteristics such as race and gender, leading to disproportionate lower performance on these subjects~\citep{Prabhu}. Additionally, some machine learning datasets scraped from the internet have been collected without subject consent, prompting to privacy concerns in the collection process~\citep{Paullada}.

 The standards and criteria used to build datasets evolve over time, and therefore, is it sensible to even consider  a stream built upon historical datasets which might not meet today's best practices on issues such as representation and consent? We posed ourselves this question and concluded that the proposed  training/evaluation protocol offers a sufficient mitigation. In particular, the ultimate evaluation is on the meta-test part of the stream, which only includes tasks extracted from the last three most recent years. Moreover, we plan to update the benchmark on a regular basis, to maintain a fresh meta-test stream. This not only alleviates potential model overfitting, but also it enables \minerva{} to track community's standards for what datasets one should consider for evaluation purposes.

A related question is around deprecated datasets. To the best of our knowledge, we have removed any deprecated dataset during the construction of \minerva, or used the most recent non-deprecated version of a dataset (e.g., most recent version of ImageNet). However, it is possible that at some point in the future a dataset currently in \minerva{} will be deprecated. In that case, we will take the responsibility to update the stream by removing any instance of such deprecated dataset in the stream.

From a machine learning point of view, deprecation raises a very interesting question of how to remove knowledge of a particular dataset from a never-ending learning system which presumably has accrued knowledge over time, including from that particular deprecated dataset. While we do not have an answer to this question, we believe that \minerva{} offers an excellent realistic environment to assess whether methods are capable of such manipulation of learned knowledge.

We also reflected on the potentially harmful downstream use cases related to vision tasks, specifically tasks related to facial recognition. Our objective is to enable the development of a core capability of an AI system as opposed to target a particular application, such as face recognition, In particular, our models are trained in a closed world setting, meaning that a classification model trained on a dataset in \minerva{} cannot be used to recognize faces of subjects outside those present in the training set, greatly alleviating concerns related to misuse of models trained on this data for surveillance related applications. Given all these considerations we opted for keeping face datasets that satisfied our requirements on deprecation.

By virtue of the methodology used to construct the stream which relies on existing datasets, we acknowledge the above mentioned  limitations of \minerva. Further progress is certainly required on data collection methods to tackle issues of representation, deprecation, consent, and potential misuse.  

However, we overall believe that the net outcome of this research is positive for our research community and society at large, as \minerva{} encourages the design of more computationally efficient models, that better leverage previous knowledge to learn the next thing more quickly. Given the amount of resources that large-scale models consume, we believe that taking such perspective is very important and it will be even more important in the future as the community further scales up foundation models. While it is still an open question how to effectively learn sequentially while saving compute, the requirement to explicitly measure not just accuracy but also the compute  will encourage researchers to strike better trade-offs than it is otherwise possible today.
\section{Conclusions and Future Work}
In this work, we introduce \minerva{}, a benchmark for evaluating life-long learners on a stream of visual classification tasks. These have been derived by uniformly sampling papers from major computer vision proceedings over the last three decades. 
Since each task is well understood, the main challenge is learning over time to accrue and transfer knowledge. Only by doing so, can learners become more accurate and efficient over time.

\minerva{} comes equipped with a rigorous training and evaluation protocol that is designed to prevent overfitting to the evaluation set. In particular learners are asked to go through the tasks of the last three years only once, and without accessing data of future tasks. Moreover, the evaluation consists of not only an assessment of the classical generalization error, but also compute in terms of FLOPs. Had we not controlled for compute, results would have been different and less revealing, as beating method A with method B would be easier once we provide B with more compute than A. Finally, \minerva{} makes meta-learning a first class citizen, since the assessment of the compute spent while learning {\em includes} the compute spent while doing hyper-parameter search. Therefore, methods that have more efficient meta-learning algorithms will be favored.

In general, \minerva{} is not just about a particular stream, but it is also {\em a process} to build benchmarks. A similar construction method could have been used on other domains, like natural language processing or reinforcement learning, for instance. \minerva{} is open-sourced with scripts to recreate the data stream, the training and evaluation framework and code implementing the most classic baselines.

Our initial results obtained by applying standard baseline approaches to \minerva{} demonstrate the importance of using such a diverse stream. We have found that methods that do transfer perform better than methods that do not, although results vary significantly by domain, image resolution and number of training examples. In particular, we have found that methods that shuffle data to learn generic representations currently strike a worse trade-off between error rate and compute than smarter versions of finetuning. Pre-training approaches perform very well, but they can achieve even superior trade-offs  by adapting their representation over time.

\minerva{} opens up several avenues of future research in never-ending  learning. One direction is towards architectures that support variable resolution inputs like the recent Perceiver~\citep{perceiver} and architectures that support efficient inference despite the large number of parameters, like mixture of experts models~\citep{hard-moe17}. Another direction is on learning algorithms that enable better transfer, model growth over time~\citep{alma22,munet}, and parameter sharing across related tasks~\citep{Rebuffi17}. Another avenue is meta-learning to better answer questions about how to initialize predictors, how to learn more quickly future tasks and how to better shape the search space of architectures, optimizers and learning algorithms. Ultimately, we conjecture that the best method in \minerva{} will need advances at the intersection of continual learning, meta-learning and AutoML, because it will have to adapt to a non-stationary stream of data, leveraging structure across tasks in order to more efficiently tune hyper-parameters for a new predictor.

In the future, we plan to keep evolving \minerva{} over time, by moving the current $\metateststream$ to $\metatrainstream$, and forming a new $\metateststream$ by adding tasks after 2021. This will prevent overfitting and make sure \minerva{} tracks the community interests and standards. In the near future, we are eager to learn how the community uses \minerva{} and we want to understand whether there are ways to improve it.
Eventually, we plan to extend \minerva{} towards a multi-task and multi-modal stream, while retaining the same rigorous training and evaluation protocol we have defined in this work.

% Acknowledgements and Disclosure of Funding should go at the end, before appendices and references
\acks{The authors wish to thank Timothy Nguyen and Joaquin Vanschoren for reviewing this work and providing extensive feedback on how to improve its clarity. The authors also thank Skanda Koppula and Iain Barr for their help with training the pre-trained models. This work has been done at DeepMind without any other source of funding.}


\newpage
\appendix

\section{Individual Contributions} \label{app:authors}
If you wish to contact us, please email us at  \href{mailto:nevis@deepmind.com}{nevis@deepmind.com}. For questions about a specific part of this work, please reach out directly to the relevant authors. Table~\ref{tab:authors_contributions} provides details on each author's contributions.
\begin{table}[h]
    \centering
    \begin{tabular}{p{0.3\textwidth}p{0.6\textwidth}}
    \toprule
    \textbf{Authors} & \textbf{Contributions} \\ 
    \midrule 
    J\"org Bornschein    & conceptualization, methodology, codebase development, FT-* baselines, analysis \\
    Alexandre Galashov &  codebase development, data pipeline, scripts to fetch data, experiments, debugging, analysis, visualizations \\ 
    Ross Hemsley & codebase design and development, baselines, data ingestion libraries, open sourcing, write-up \\
    Amal Rannen-Triki & conceptualization, methodology, stream construction, scripts to fetch data, Indep baseline, analysis, write-up\\ 
    Yutian Chen & scripts to fetch data, BHPO, analysis\\ 
    Arslan Chaudhry & scripts to fetch data, multitasking baselines, analysis, write-up\\ 
     Xu Owen He  & scripts to fetch data, PT-* baselines, forward transfer ablation,  analysis, writing\\ 
     Arthur Douillard & data ingestion libraries, PyTorch codebase, open sourcing\\ 
     Massimo Caccia & ensembling baseline, beta testing\\ 
     Qixuan Feng & SplitImageNet ablation, tensorboard in open-source code\\ 
     Jiajun Shen & ablation on stream variant (Tab.~\ref{tab:cripple_stream_ablation}), memory handling in open source code\\ 
     Sylvestre-Alvise Rebuffi & ViT, discussions \\ 
     Kitty Stacpoole & program management, coordination \\ 
     Diego de las Casas & initial design of codebase\\ 
     Will Hawkins & ethical considerations \& review.\\ 
     Angeliki Lazaridou & discussions \\ 
     Yee Whye Teh & discussions \\ 
     Andrei A. Rusu & conceptualization, methodology, formal analyses, writing  - review \& editing.\\ 
     Razvan Pascanu &  conceptualization, analyzing and discussing results, writing \\
      Marc’Aurelio Ranzato & conceptualization, stream construction, scripts to fetch data, experiments with FT-*,  analysis, writing, team coordination and planning  \\ 
    \end{tabular}
    \caption{Authors contributions.}
    \label{tab:authors_contributions}
\end{table}
%List here the contributions of each co-author.

    %   \name  Marc’Aurelio Ranzato \email ranzato@deepmind.com\\
    %   \addr DeepMind \\
    %   }

\newpage
\clearpage
%\section{Baseline Details}
%\mr{Amal, Ross, Yutian, Arslan, Owen: Can we remove this section? or are there things we need to add here?}

\iffalse
\section{Formalization}
Let us formalize the continual learning problem we are interested in studying. 

Let $\mathcal{D} = \mathcal{D}_x \times \mathcal{D}_y$, where $\mathcal{D}$ is the Cartesian product between $\mathcal{D}_x$, the domain of the observation, and $\mathcal{D}_y$, the domain of targets that we are interested in learning over. Note $\mathcal{D}$ is the domain of all possible tasks. In particular, let $\mathcal{D}_i \in \mathcal{D}$ represent the data for task $i$. I.e. in this formalism we assume that all tasks of interest that we want to learn seuqnetially can be represented with observations in $\mathcal{D}_x$ and outputs in $\mathcal{D}_y$.
\fi

\clearpage

\ag{
\section{High level description of main results and methods used} 
\label{app:overview}
}

In this section we give an overview of results and methods presented in all the Appendix sections. For more detailed discussions, please refer to each of the appendix sections.

In Section~\ref{app:cheapo}, we analyze the behavior of different methods under lower computational budget than in the main paper (see Section~\ref{sec:main_results}). The lower compute budget is achieved by using cheaper arhictectures (ResNet-18 and ResNet-34) as well as using fewer trials in hyperparameters search and fewer updates. Overall, we show that the same conclusion as presented in Section~\ref{sec:main_results}, hold.

In Section~\ref{app:arch}, we present results showing impact of the architecture choice. We consider different backbone architectures, notably: VGG, ResNet34, ResNet50, Vision Transformers (ViT), in particular, ViT-B8. On top of that, for ResNet-34 architecture we provide study of the impact of the input image resolultion, as well as the number of channels in the first residual block. There is a simple compute-performance trade-off for choosing these values, i.e. - larger resolutions lead to better performance and lower compute, and vice-versa.

In Section~\ref{app:transfer}, we explain the method of calculating the transfer matrix which is required for FT-d (see Section~\ref{sec:main_results}) method as well as to understand the task structure, see Figure~\ref{fig:stats_domain}.

In Section~\ref{apx:baselines}, we present additional baseline results. In Section~\ref{apx:finetuning}, we present ablations of different sequential finetuning baselines, i.e., FT-d, FT-s, FT-prev, as well as the impact of using pretrained models (PT-ext). In Figure~\ref{fig:ft-d} and Figure~\ref{fig:pt-ft-d}, we study the qualitative behaviour of FT-d finetuning method without (FT-d) and with (PT-ext + FT-d) pretraining. In Section~\ref{apx:pretraining}, we additionally study the impact of different pre-training and finetuning strategies.

In Section~\ref{app:task_order}, we add quantitative results showing the impact of task orders in the stream. We found that the overall performance is very sensitive on the position of ImageNet dataset in the stream.

In Section~\ref{app:multitask}, we study the performance of multi-task finetuning strategy. We show that it is very important how the network is initialised when it starts to train on each new task in the stream. Moreover, we found that picking all previous task versus a subset did not lead to a very different performance.

In Section~\ref{app:bhpo}, we provide more detailed analysis of Bayesian Hyper-Parameters Optimization (BHPO) method. In particular, we show that the benefit of BHPO increases with higher dimensionality of the search space. In low-dimensional (2-D) search space, BHPO performs similarly to Random Search.

In Section~\ref{app:ensembling}, we present additional results showing that simple ensembling strategy added on top of the reference model could improve results. This, however, comes at the expense of increased inference cost.

In Section~\ref{app:measuring_compute}, we explain the reasonning behind choosing cumulative-FLOPS (cFLOPS) as compute metric as well as alternatives which we considered.

\clearpage

\section{Baselines at a Lower Computational Budget} \label{app:cheapo}
\begin{figure}[th]
    %\centering
    \includegraphics[width=\textwidth]{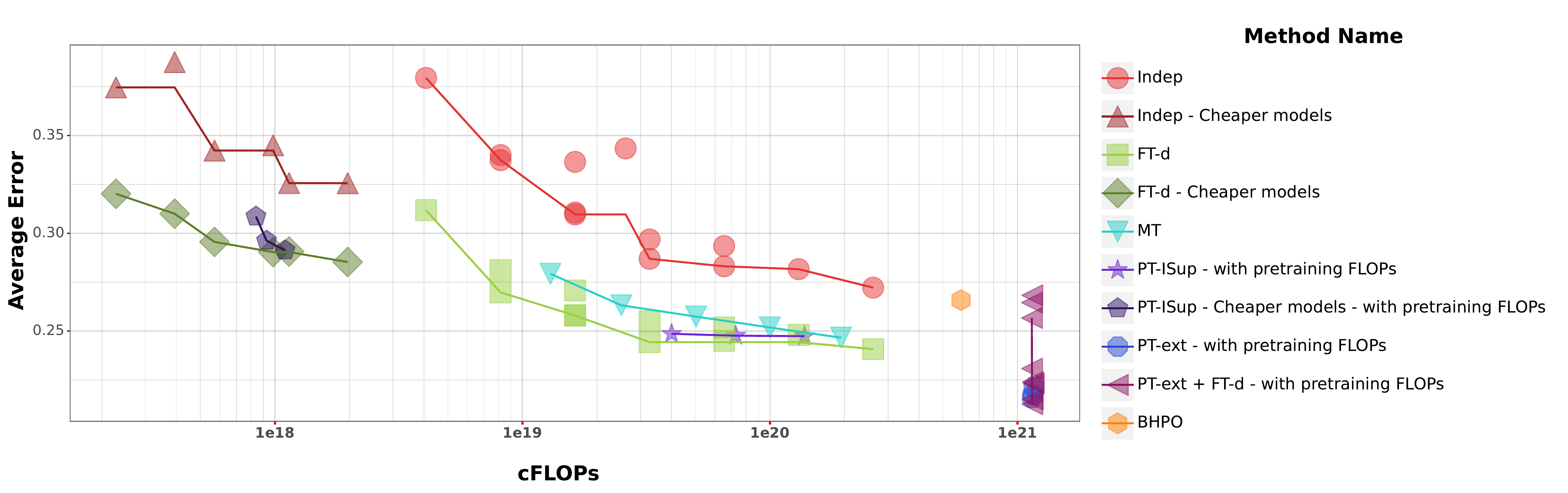}
    \includegraphics[width=0.935\textwidth]{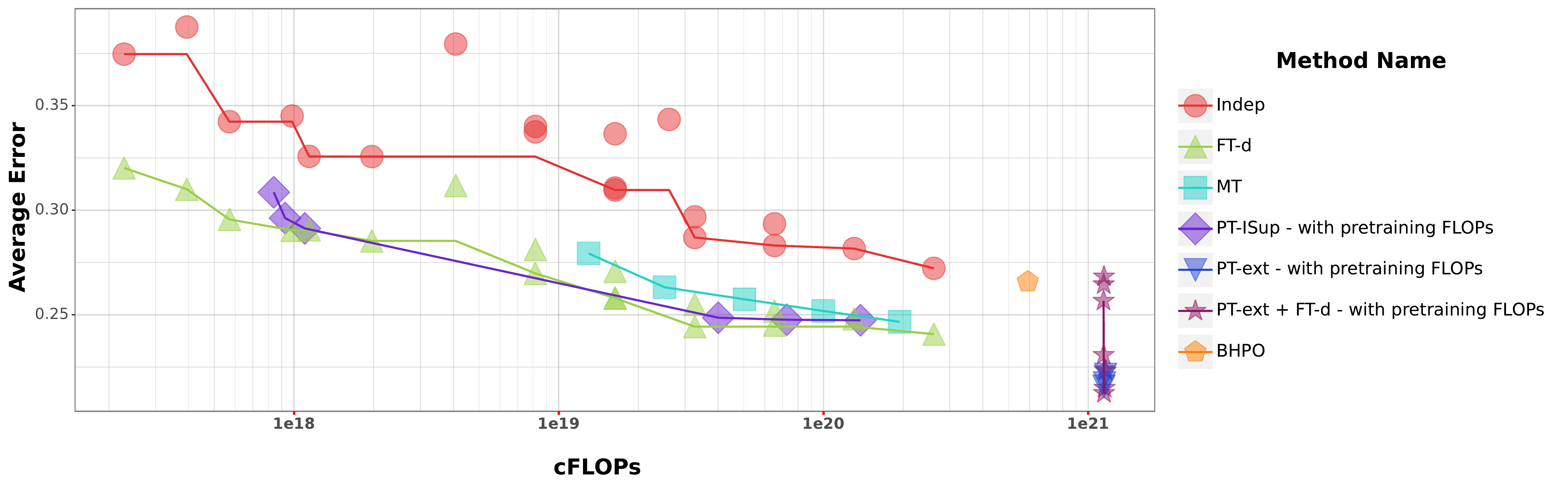}
    \caption{Top: Pareto fronts when considering standard ResNet18 and ResNet34 trained with fewer trials and number of updates (cheaper models). The experiment represented by the third marker of FT-d on the far left ran for 28 hours on the full stream when distributing the hyper-parameter search of the predictor over four devices. On a single worker, the same experiment takes about four days. Bottom: The same pareto front as above, but using a single curve for FT-d, PT-ISUP and Indep.}
    \label{fig:cheapo}
\end{figure}

In the experiment of Fig.~\ref{fig:cheapo} we explore another region of the pareto front: the extremely frugal learners that require between $10^{17}$ and $10^{18}$ to run on the full stream. This amounts to about $4$ days on a single GPU with 16GB of RAM. Unlike what we presented in the main paper where we used a ResNet34 adapted to low-resolution images (with many more channels and without as much spatial sub-sampling), here the network is a default ResNet18 and ResNet34 architecture, which have been designed for higher resolution images. These architectures are cheaper because a) they have at most the same number of blocks, but b) they have many fewer channels. We further reduce compute by fixing the label smoothing parameter to the mid value of our search range, $0.15$, and by searching only over four values of learning rate, namely \{1e-4, 1e-3, 1e-2, 1e-1\}. 

We can see that these cheaper architectures and hyper-parameter strategy extend the Pareto front at lower budgets. This suggests that another possible avenue of future work is the design of more efficient architectures and hyper-parameter search methods. Importantly, this might provide an easy entry point to \minerva{} for researchers who have limited computational resources at their disposal.

\section{Architectures} \label{app:arch}
\begin{figure}[h]
    \centering
    \includegraphics[width=.9\textwidth]{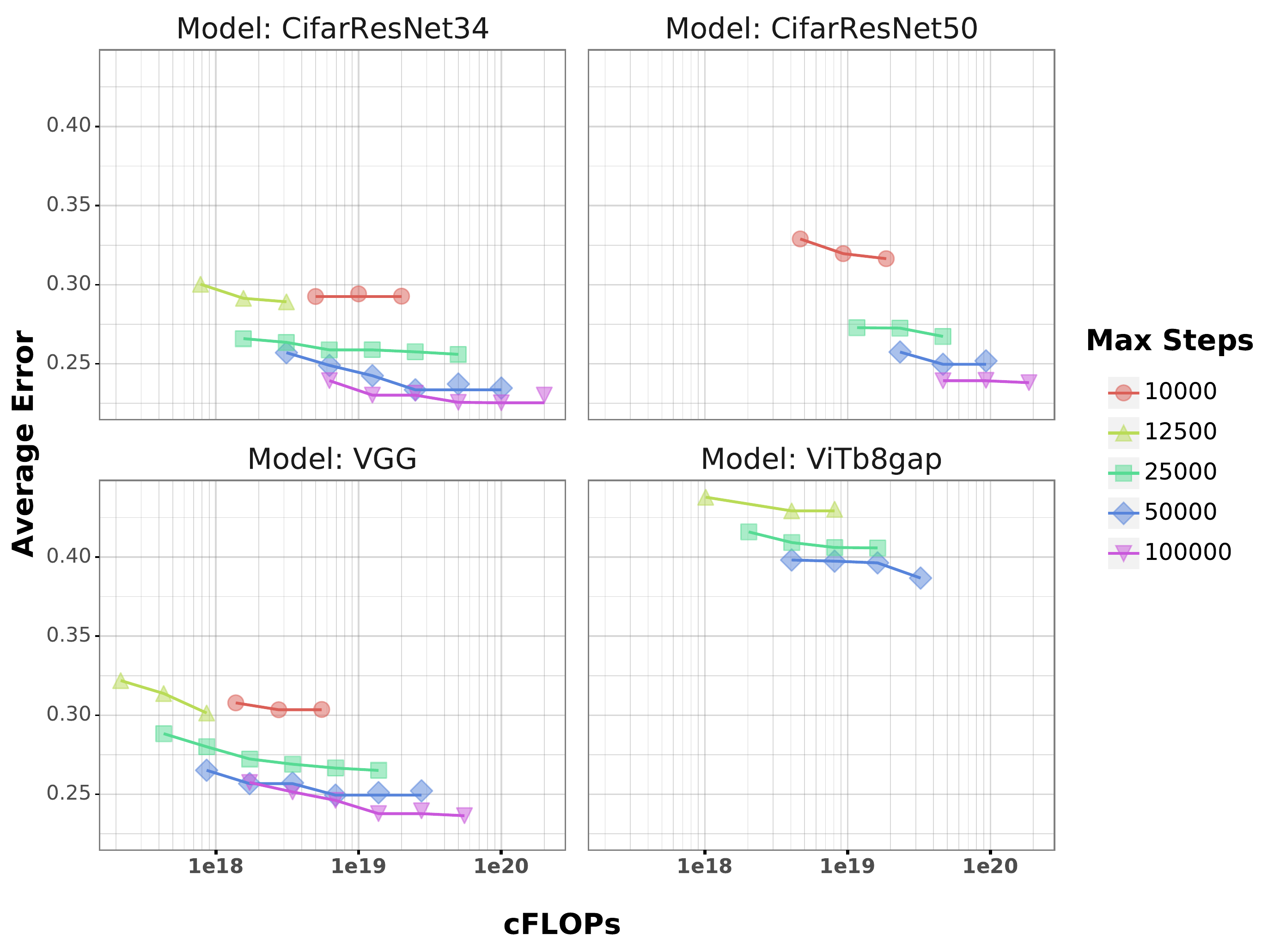}
    \caption{Independent learners with different architecture variants. }
    \label{fig:architectures}
\end{figure}

In this section, we study the impact of the architecture choice. We focus on the independent baseline used as reference in all our experiments, and we conduct two experiments. In the first, we use different architecture variants: VGG, ResNet34, ResNet50 and Vision Transformers (ViT)~\cite{dosovitskiy2020image}. In this manuscript we use a ViT-B8 with the modifications proposed by~\cite{he2022masked} where the classifier is applied after global average pooling over the vision tokens. As all the experiments are conducted using the same resources, the bigger architectures (ResNet50 and ViT) are run with a smaller maximum batch size (see Eq.~\eqref{eq:adaptive_bsz}), and similar search spaces for the number of updates and learning rate. It is therefore expected to see these architectures underperforming, as we observe in Fig.~\ref{fig:architectures}.

\begin{figure}[h]
    \centering
    \includegraphics[width=.9\textwidth]{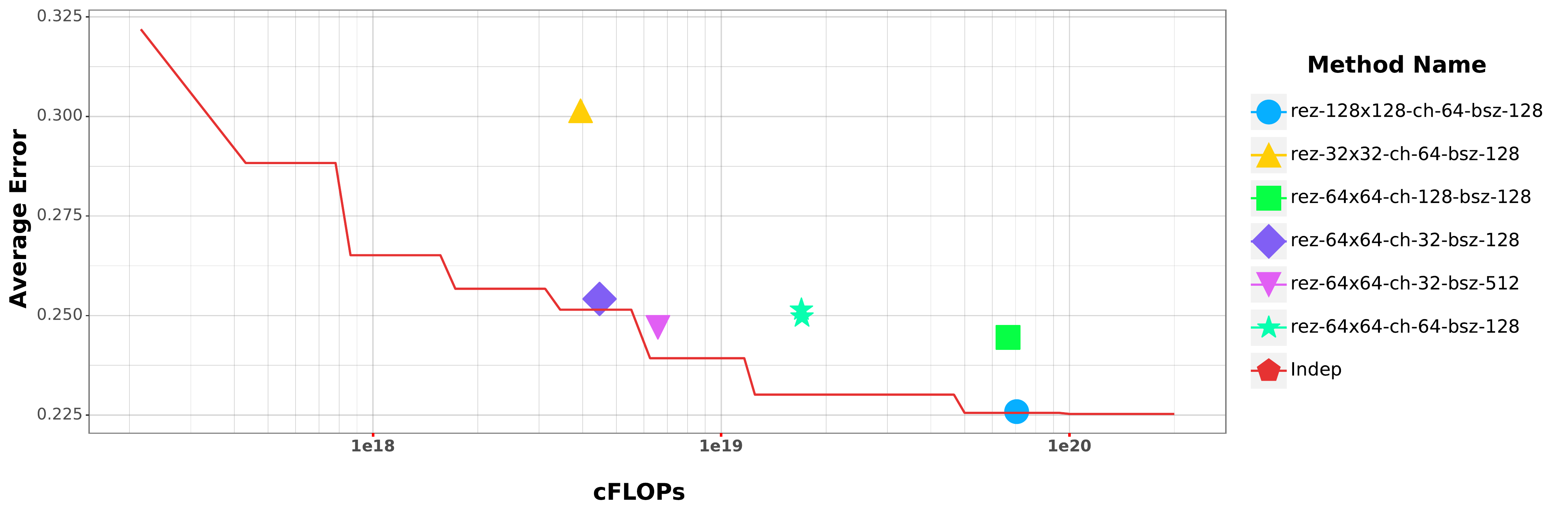}
    \caption{Performance of Indep when varying the input image resolution and the network width. In the legend, ch-$c$ corresponds to the number of channels in the first residual block. The number of channels in the remaining blocks are consecutively doubled.}
    \label{fig:width}
\end{figure}

The second experiment focuses on ResNet34, chosen as default architecture, and varies the number of channels in the different residual blocks. The results of this experiment complete Fig.~\ref{fig:indep_resolution_sensitivity}, providing another axis to trade off error rate versus compute. The full results are shown in Fig.~\ref{fig:width}. In the legend, rez-$r$x$r$ indicates the input resolution, ch-$c$ corresponds to the number of channels in the first residual block (the number of channels in the remaining blocks are consecutively doubled), and bsz-$b$ shows the maximum batch size. We observe that reducing the number of channels saves a significant compute budget, without a significant loss in performance. 

\section{Additional Baseline Results}\label{apx:baselines}
In this section, we report additional  results on the \minerva{} stream. We namely show the results of the finetuning and pretraining families, and complete the experiments reported in Sec.~\ref{sec:main_results}.
\subsection{Finetuning}\label{apx:finetuning}
\begin{figure}[h]
    \centering
    \includegraphics[width=0.99\textwidth]{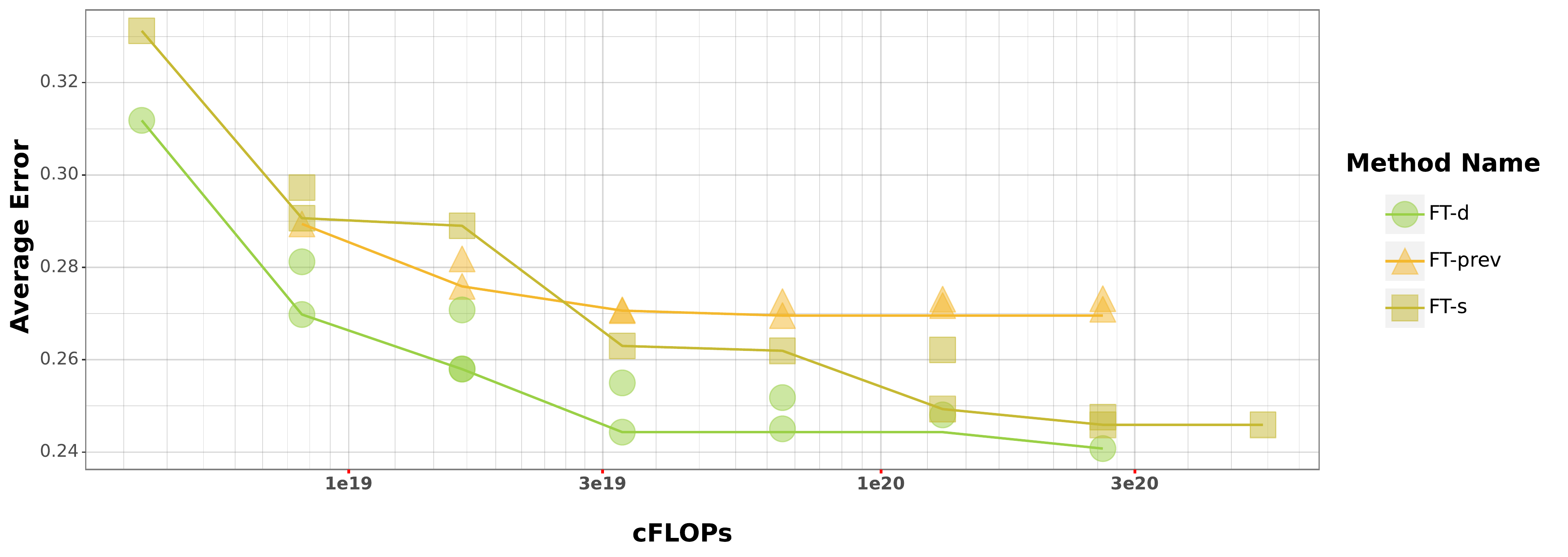}
    \caption{Finetuning baselines: Pareto fronts}
    \label{fig:pareto_finetuning}
\end{figure}

\begin{figure}[h]
    \centering
    \includegraphics[width=0.99\textwidth]{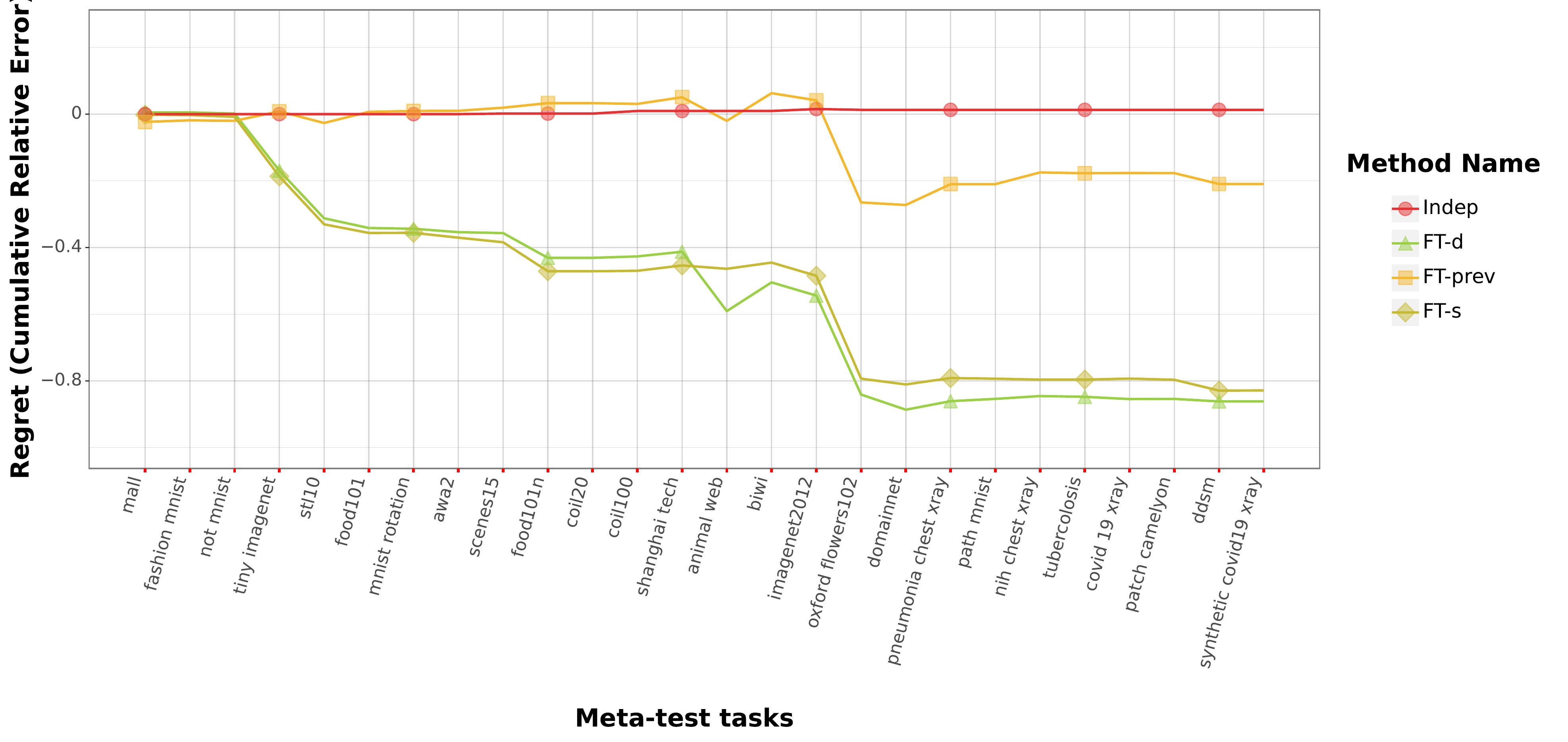}
    \includegraphics[width=0.99\textwidth]{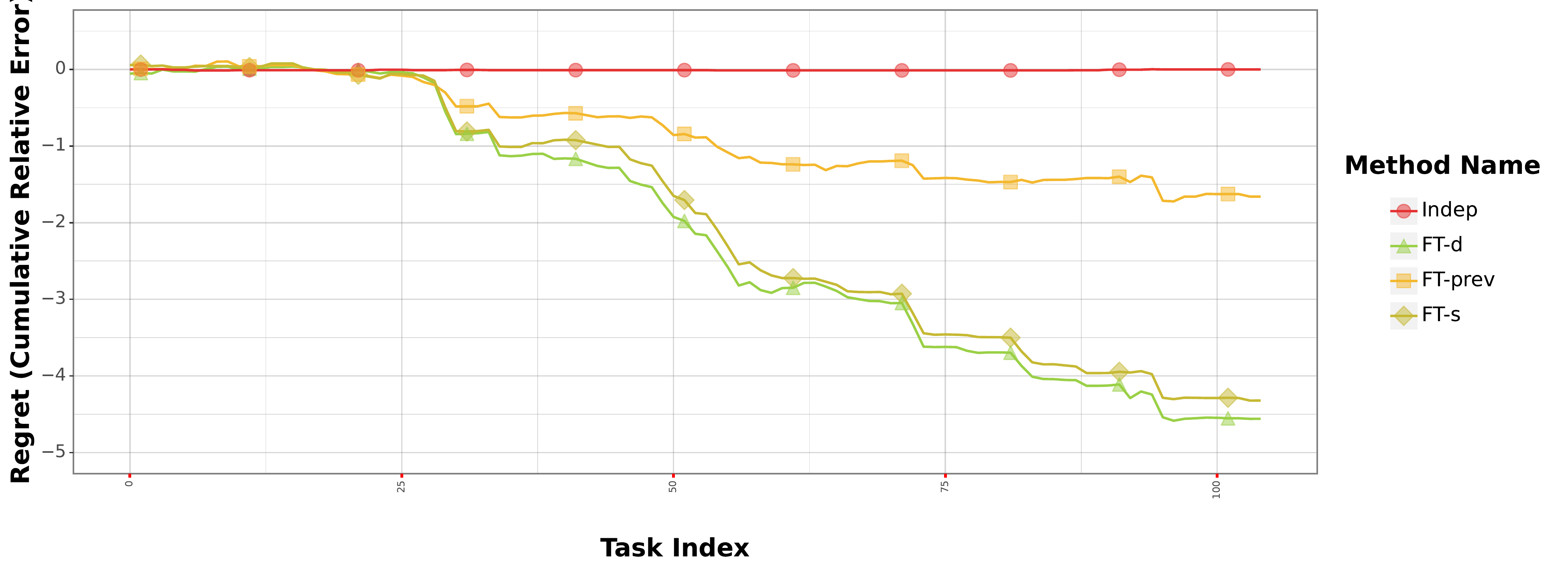}
    \caption{Finetuning baselines - Regret plots: Cumulative error rate relative to Indep. on $\metateststream$ (top) and on the full stream (bottom)}
    \label{fig:regret_finetuning}
\end{figure}

This section focuses on the Finetuning family. In Fig.~\ref{fig:pareto_finetuning}, we show the Pareto fronts of FT-d, FT-prev and FT-s. Fig.~\ref{fig:regret_finetuning} reports the regret plot of the same methods, i.e.\ the cumulative error over time relative to Indep, picking hyper-parameters such that all methods use roughly the same amount of compute (same as in Fig.~\ref{fig:regret_plots}). These results demonstrate the importance of the choice of what to finetune from, as FT-prev is significantly worse than FT-s and FT-d, and FT-s is worse than FT-d suggesting that it is important to make an accurate (up-to-date) estimate of task relatedness. The regret plot further shows that the tested finetuning strategies fail to increase transfer over time (linear slope over the full stream) and to accumulate knowledge over the last 9 tasks of the sequence (horizontal curves).  

%\clearpage

\subsubsection{Analysis of FT-d}
\begin{figure}[h]
  \centering
  \includegraphics[trim={0, 0, 0.3cm 0}, clip, width=.99\textwidth]{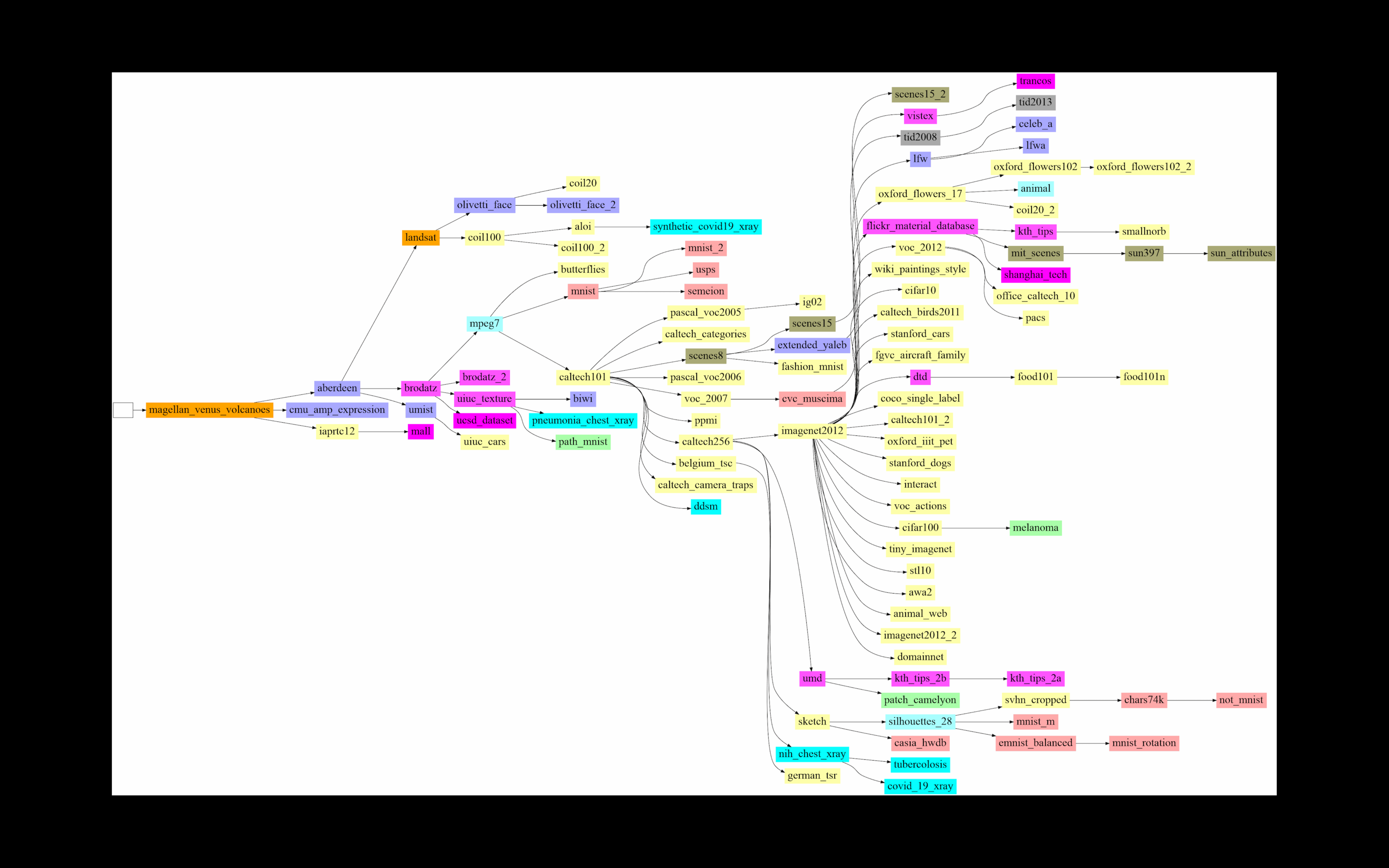}
  \caption{Graph showing the inner working of FT-d. Each box correspond to a task, the color represents the domain. An arrow connecting task $i$ to task $j$ indicates that task $i$ was selected as initialization for the network trained on task $j$.}
  \label{fig:ft-d}
\end{figure}

In Fig.~\ref{fig:ft-d} we report an example of the finetuning sequence learned by FT-d (this is the version which was trained on every task with $50,000$ steps using $16$ trials of hyper-parameter search).
We notice that there are several hubs from which several other models are finetuned from. The biggest hub is ImageNet, followed by Caltech256, MNIST, Caltech101, Scene8, etc. In the graph tasks are mostly organized by visual similarity and domain. For instance, we see two clusters of OCR tasks in red, cluster of medical images in cyan, and a large cluster of generic object recognition tasks in yellow. Finally and perhaps most surprisingly, we observe fairly long chains of finetuning models. It seems that finetuning even more than ten times can produce high performing models.

\subsubsection{Analysis of PT+FT}\label{apx:pt-ft}

\begin{figure}[h]
  \centering
  \includegraphics[width=.9\textwidth]{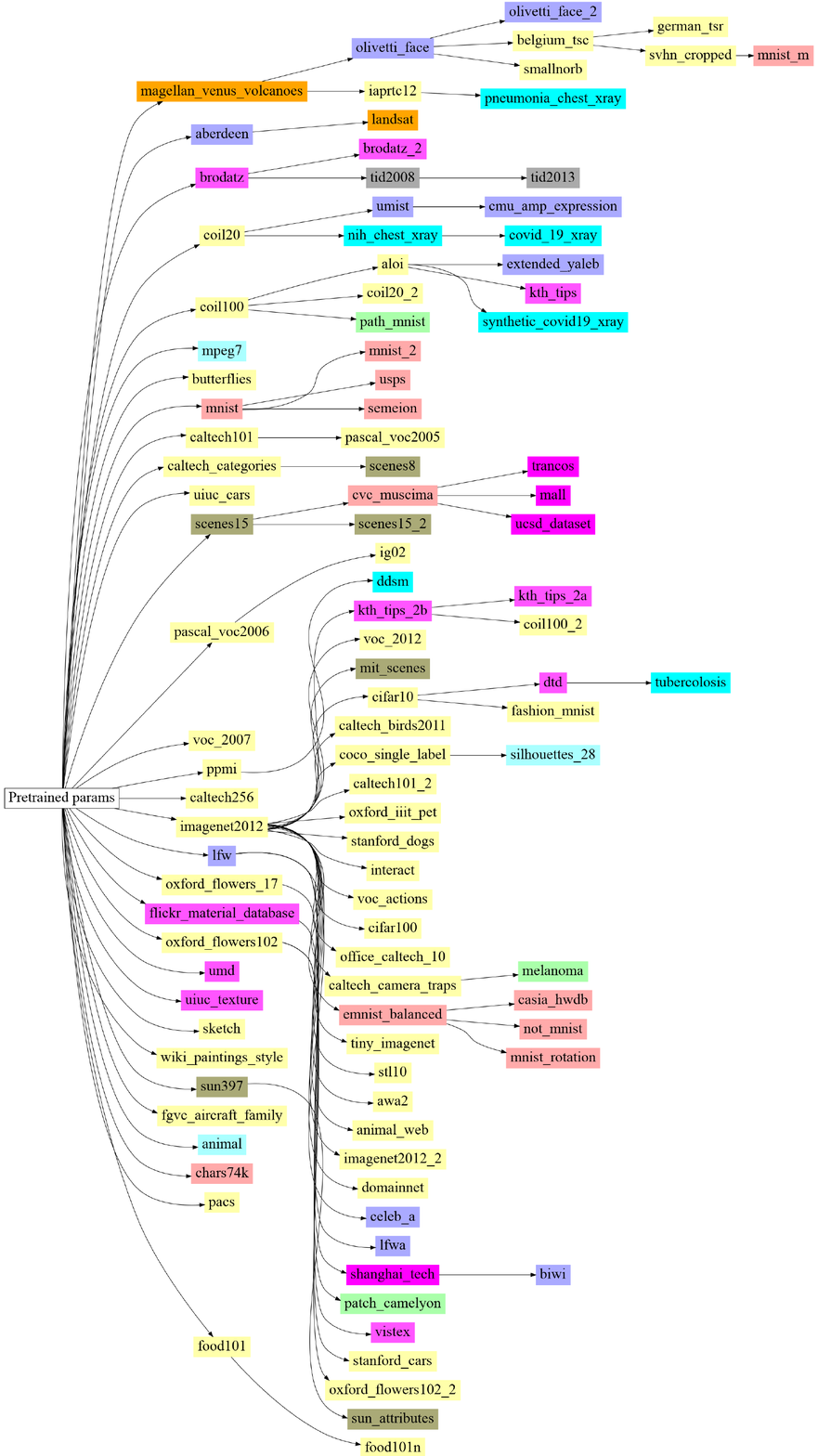}
  \caption{Graph showing the inner working of PT+FT (i.e., FT-d starting from a pretrained model). Each box correspond to a task, the color represents the domain. An arrow connecting task $i$ to task $j$ indicates that task $i$ was selected as initialization for the network trained on task $j$. Compare this graph to the one of FT-d (starting from scratch) in Fig.~\ref{fig:ft-d}.}
  \label{fig:pt-ft-d}
\end{figure}
In Fig.~\ref{fig:pt-ft-d} we report an example of the finetuning sequence learned by PT+FT, i.e., FT-d starting from the model pretrained on external data (PT-ext). We report the version which is reported in the regret plots in Figures~\ref{fig:regret_plots} and~\ref{fig:regret_pretraining}, which corresponds to the best compute-error trade-off in Figures~\ref{fig:full_stream-pareto_fronts} and~\ref{fig:pareto_pretraining}.

Compared to the finetuning sequence learned by FT-d when learning from scratch (see  Fig.~\ref{fig:ft-d}), we observe that while chains are shorter, not all tasks choose to be initialized from the original pre-trained model. ImageNet is still a strong hub, but there are many chains of length two or three that cluster datasets by domain and visual similarity.

\clearpage

\subsection{Pretraining}\label{apx:pretraining}
This section focuses on the Pretraining family. Below, we report results using models pretrained on ImageNet using full supervision (PT-ISup), on the meta-train part of the stream (PT-MT), and on ALIGN and Stock using CLIP (PT-ext). Note that this last variant uses not only a much larger (external) dataset, but also a substantially more powerful architecture. 

\begin{figure}[h]
    \includegraphics[width=0.86\textwidth]{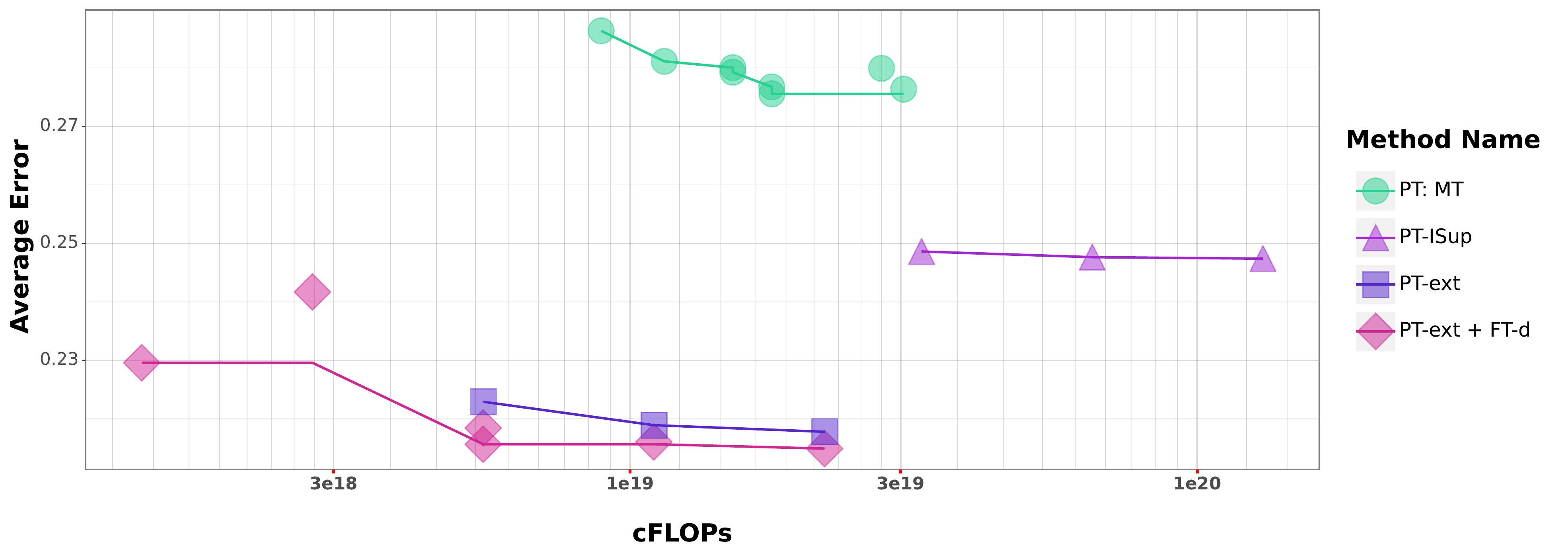}
    \includegraphics[width=0.99\textwidth]{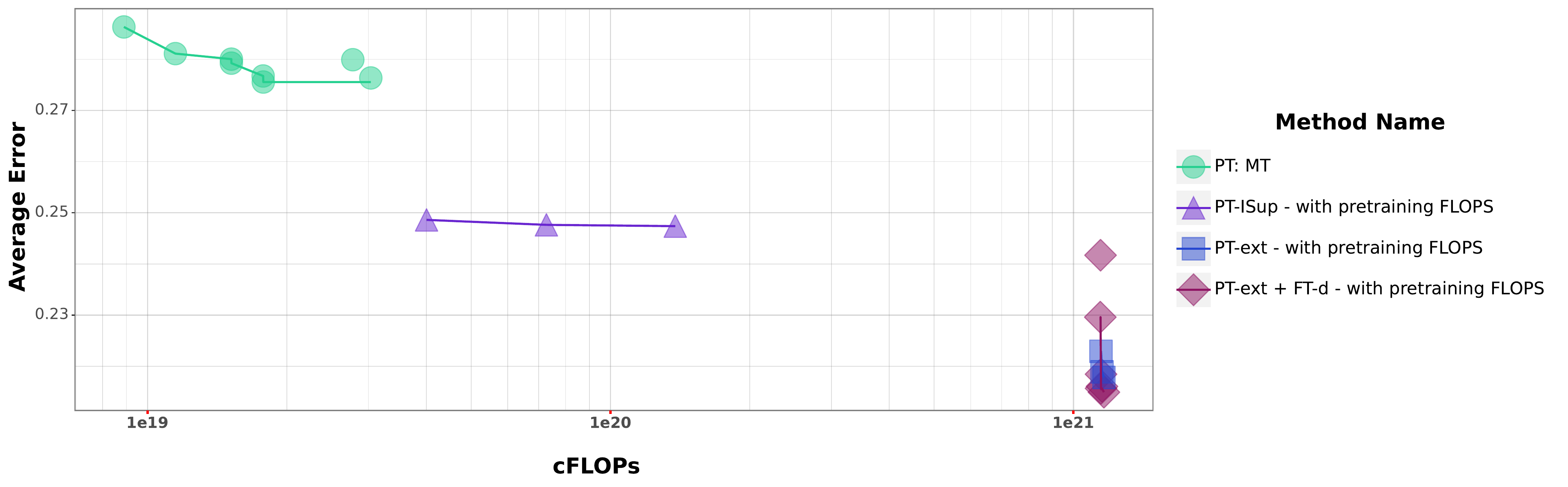}
    \caption{Pretraining baselines: Pareto fronts before (top) and after (bottom) accounting for the flops using during pretraining.}
    \label{fig:pareto_pretraining}
\end{figure}

In Fig.~\ref{fig:pareto_pretraining}, we show the Pareto fronts of these methods, with (bottom) and without (top) the computational cost of pretraining. As expected, we observe that PT-ext leads to a significantly higher performance. However, when the pretraining cost is taken into account, the trade-off between performance and compute is less impressive. We also observe that training on the whole training part of the stream (which includes ImageNet) performs worse than pretraining on ImageNet only. 

\begin{figure}[h]
    \centering
    \includegraphics[width=0.99\textwidth]{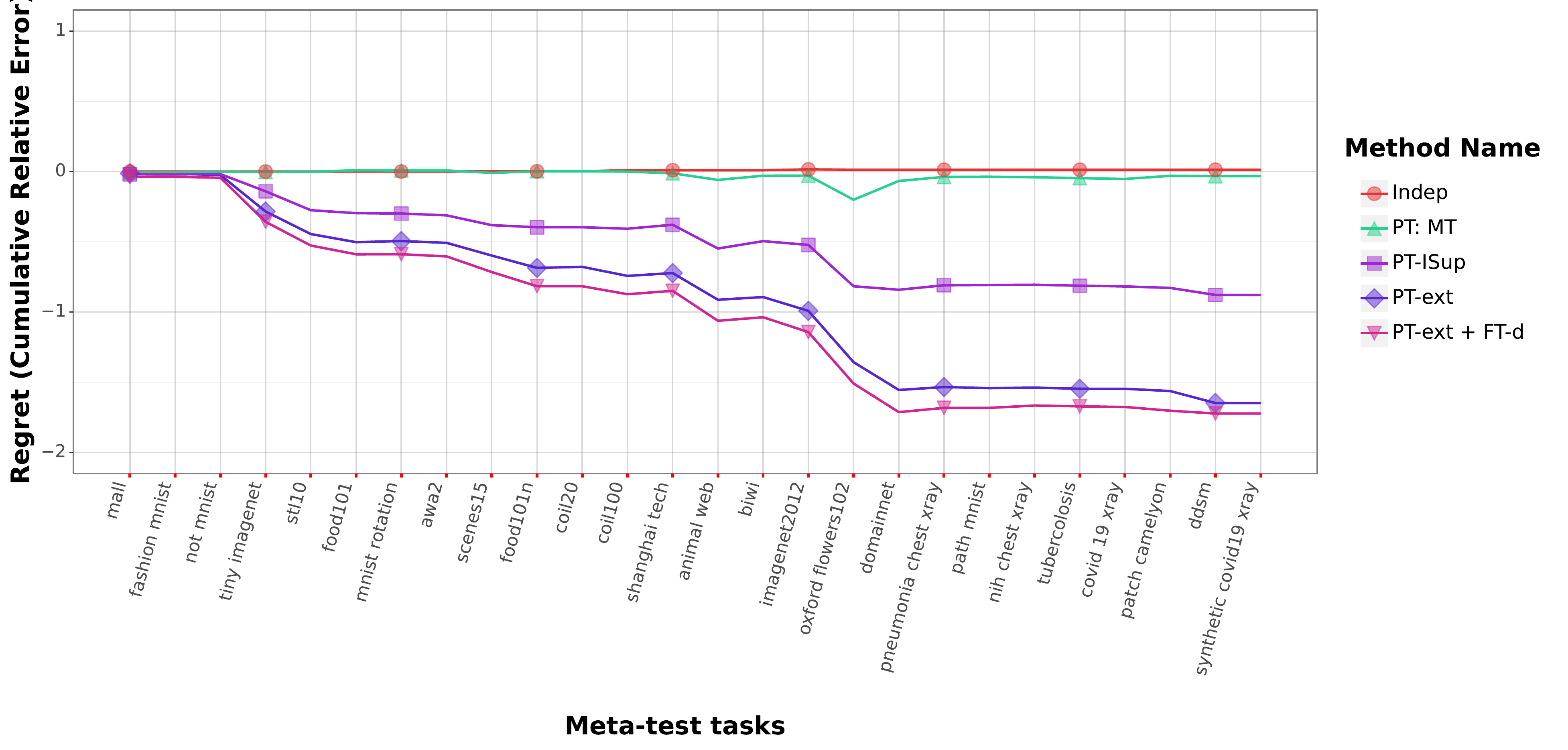}
    \includegraphics[width=0.99\textwidth]{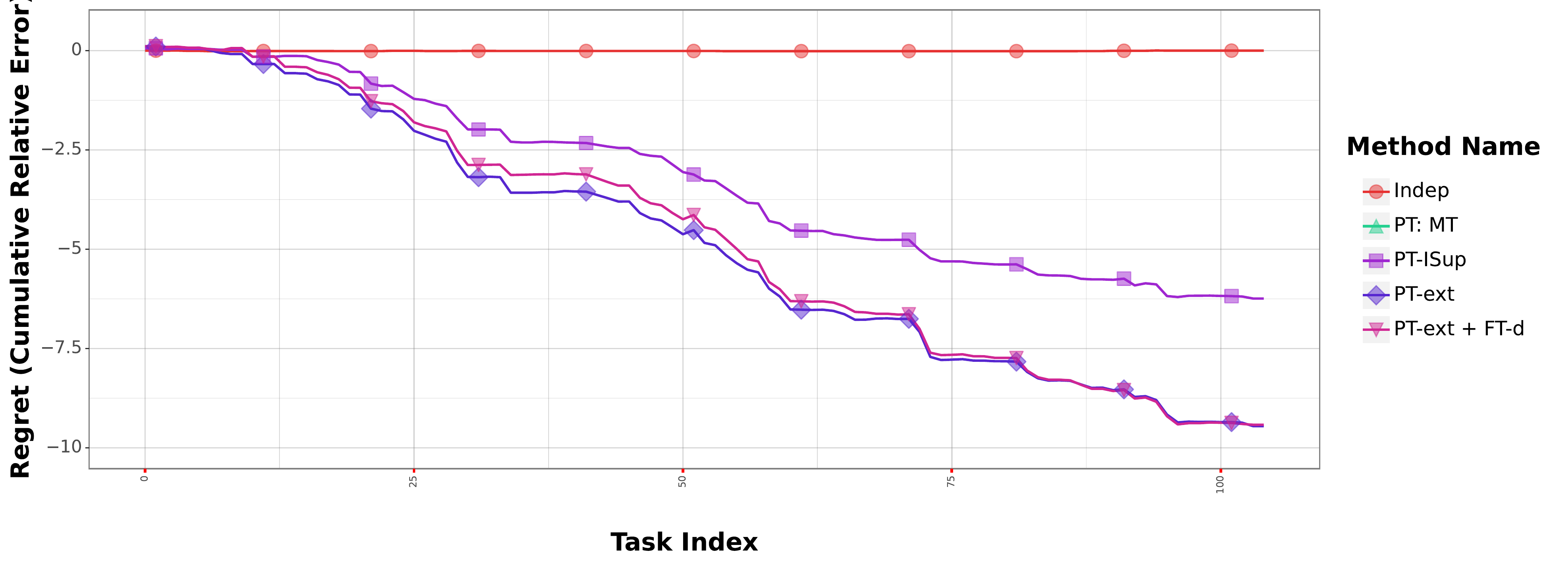}
    \caption{Regret plots: Pretraining baselines}
    \label{fig:regret_pretraining}
\end{figure}

In Fig.~\ref{fig:regret_pretraining}, we report the regret plots of the pretraining methods on the last 3 years (top) and the full stream (bottom) with respect to the Indep baseline. As none of these methods accumulate knowledge, it is not surprising to observe that their transfer does not improve over time. However, it is more surprising to observe that they are on a par with Indep for the last tasks (horizontal curves). This shows that even the pretrained model on the largest dataset (PT-ext) fails to transfer well to the medical datasets that appear towards the end of the stream. 

\section{Task Ordering} \label{app:task_order}

\begin{figure}
    \centering
       \begin{subfigure}[b]{0.8\textwidth}
        \includegraphics[width=\textwidth]{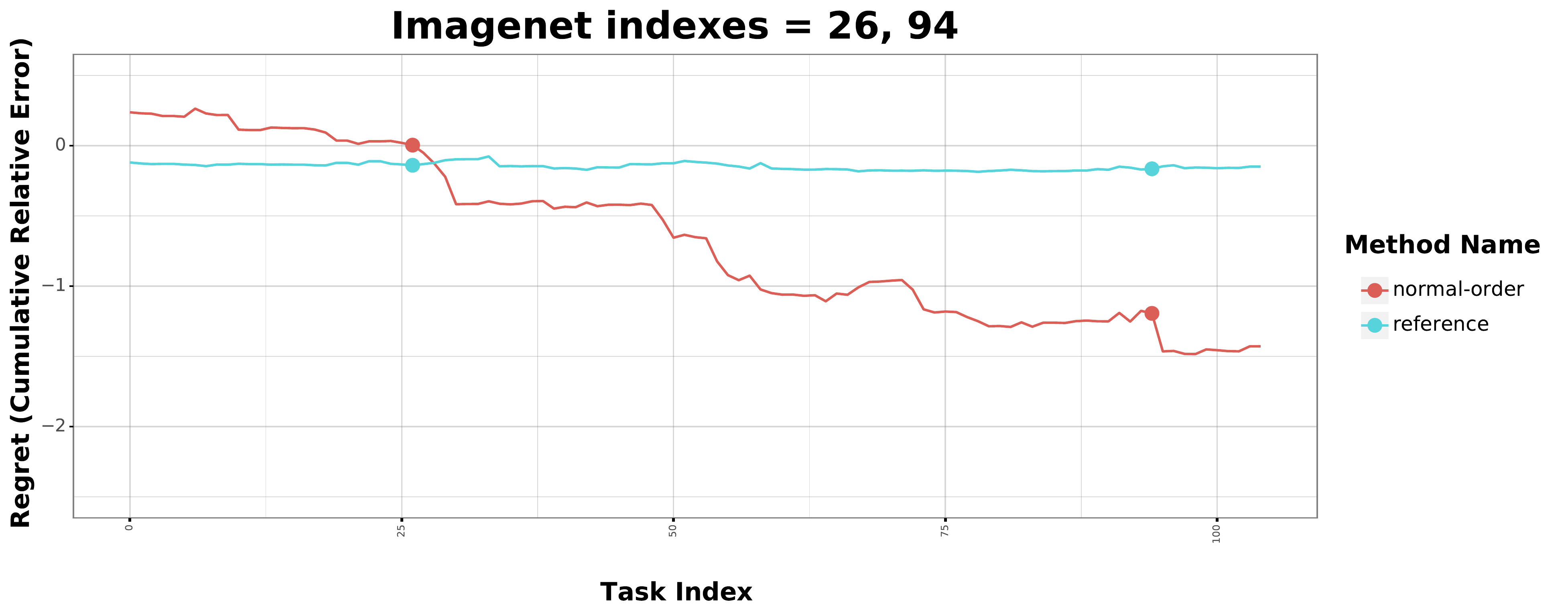}
        \caption{Normal order}
        \label{fig:normal-order}
      \end{subfigure}
      \begin{subfigure}[b]{0.8\textwidth}
        \includegraphics[width=\textwidth]{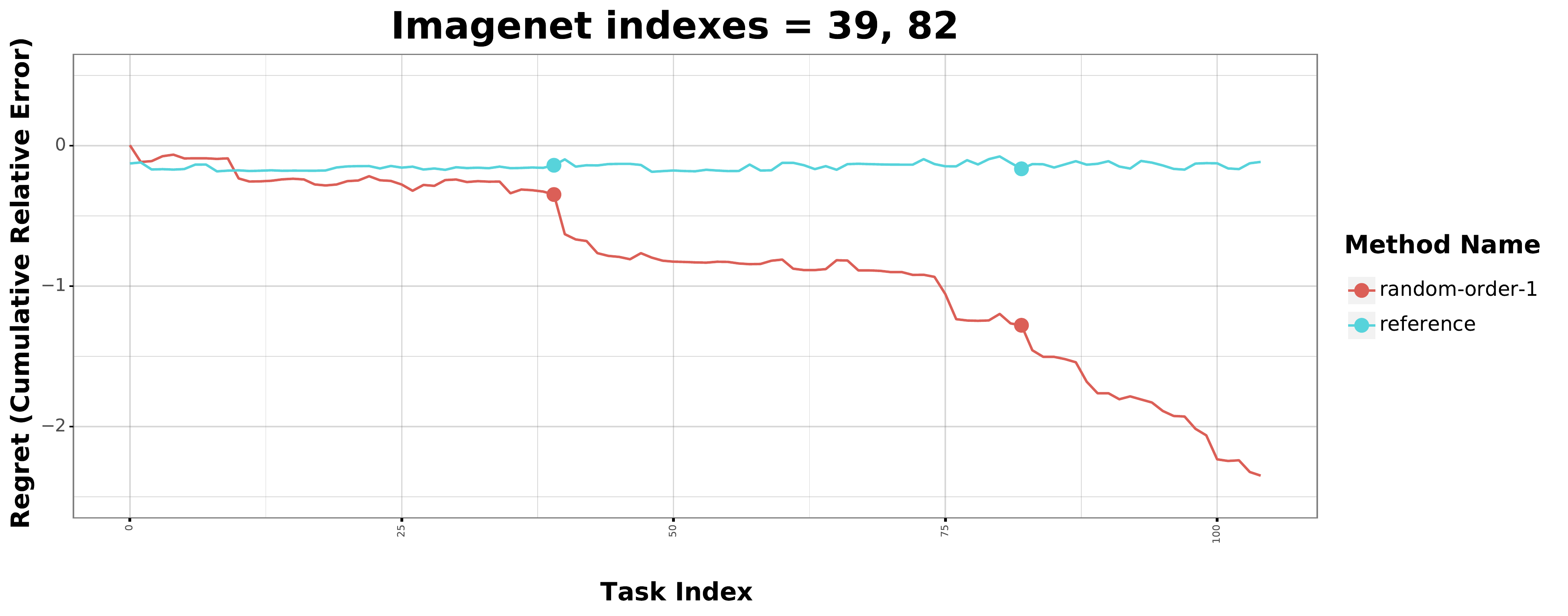}
        \caption{Random order (1)}
        \label{fig:random-order-1}
      \end{subfigure}
     \begin{subfigure}[b]{0.8\textwidth}
        \includegraphics[width=\textwidth]{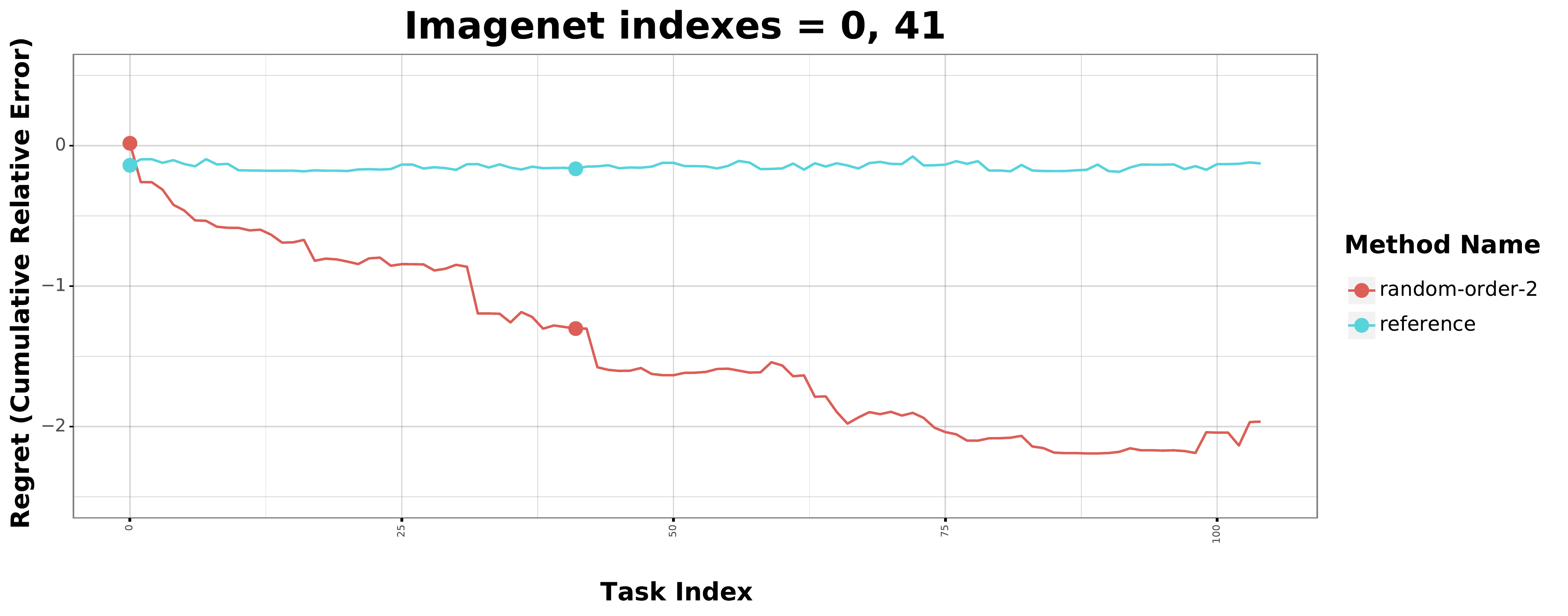}
        \caption{Random order (2)}
        \label{fig:random-order-2}
      \end{subfigure}
     \begin{subfigure}[b]{0.8\textwidth}
        \includegraphics[width=\textwidth]{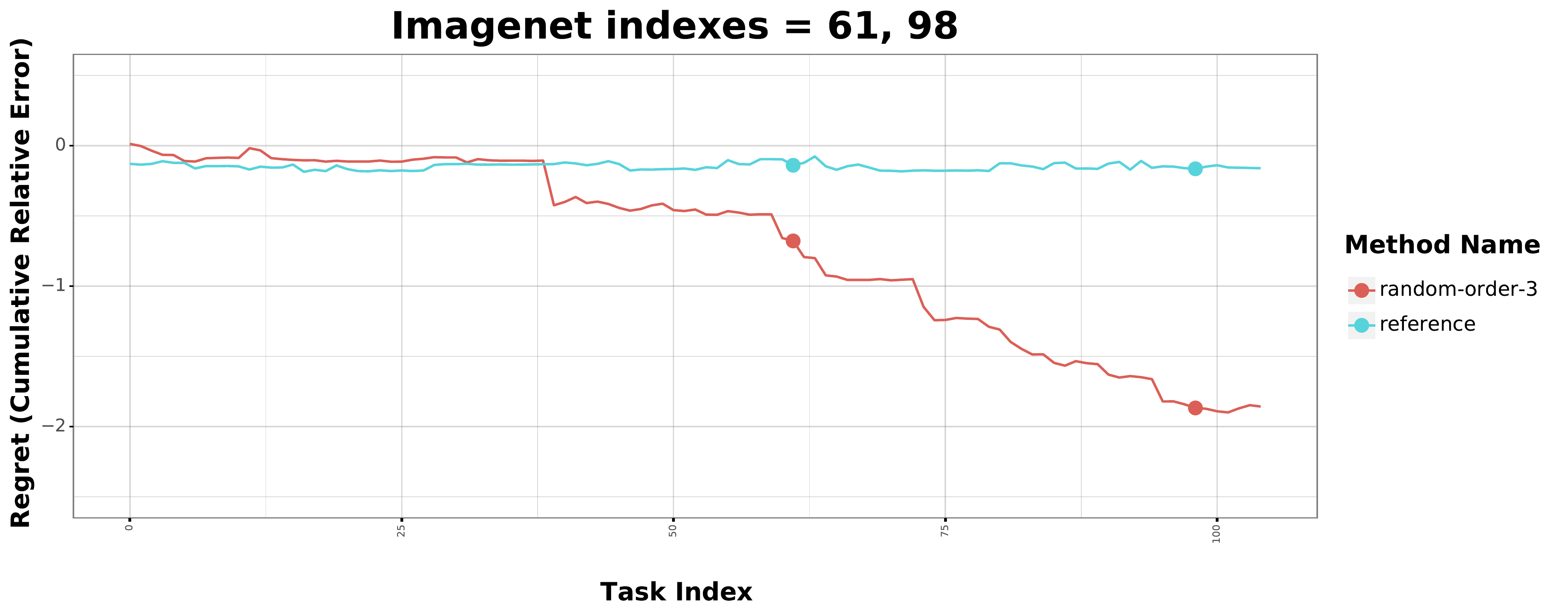}
        \caption{Random order (3)}
        \label{fig:random-order-3}
      \end{subfigure}
    \caption{Regret plots for different stream orderings.}
    \label{fig:order-regret}
\end{figure}

This section completes the results reported in Fig.~\ref{fig:-dataset-order-ablation}. In Fig.~\ref{fig:order-regret}, we show the regret plots with respect to the Indep baseline for different random stream orderings. corresponding to the ``Random Order (All stream)'' experiments in Fig.~\ref{fig:-dataset-order-ablation}. The positions of ImageNet in the stream are displayed on top of each of the panels, and highlighted on the curves. These results show how the position of this dataset, which dominates the stream in terms of size and complexity, influences the performance. The experiment where ImageNet appears at the beginning of the stream corresponds to the best performance reported in Fig.~\ref{fig:-dataset-order-ablation}.

%\clearpage

\section{Multitask Ablation} \label{app:multitask}

\begin{figure}[t]
  \centering
  \includegraphics[width=.7\textwidth]{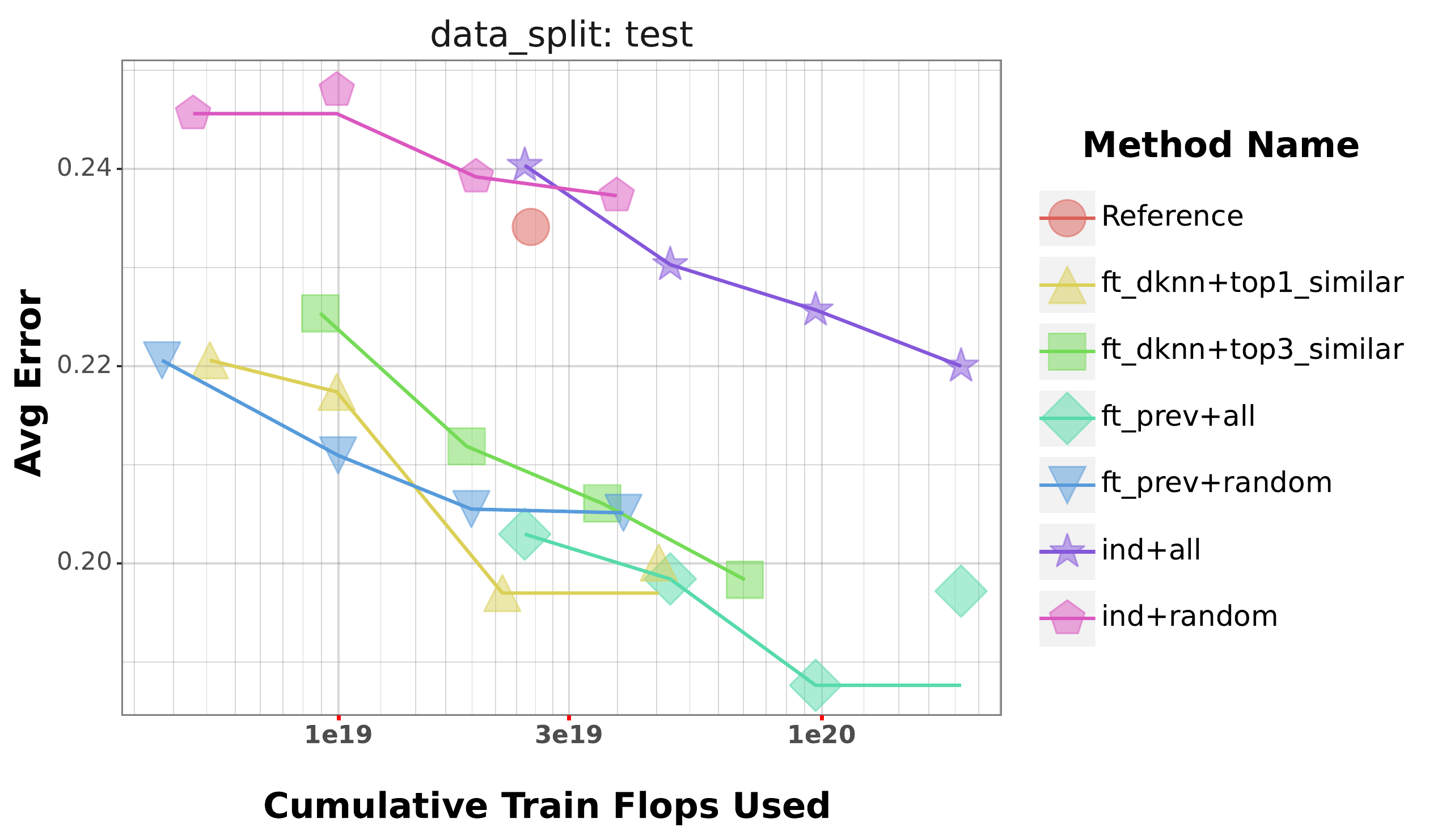}
  \caption{Pareto front of meta-learners using different multitasking strategies on SHORT stream. \texttt{ft\_dknn+top1\_similar} refer to multitask baseline where the meta-learner initializes the predictor of a new task using most similar previous task using dynamic KNN transfer matrix, and use only $k=1$ auxiliary task. Similarly,  \texttt{ft\_dknn+top3\_similar} uses $k=3$ auxiliary tasks.  \texttt{ft\_prev+all} and  \texttt{ft\_prev+random} refer to multitask baselines where the predictor of a new task is initialized form most recent previous task's parameters and meta-learner either uses "all" the previous tasks as auxiliary tasks, or randomly picks one of the previous tasks as an auxiliary task during an SGD update. Finally,  \texttt{ind+all} and  \texttt{ind+random} refer to multitask baselines where the parameter are initialized randomly and either all or one of the previous tasks are used as auxiliary tasks.}
  \label{fig:multitask_ablation}
\end{figure}

In Fig.~\ref{fig:multitask_ablation} we show how various hyper-parameters choices affect the performance of MT.
In particular, we observe that picking all tasks versus sampling a subset of them does not yield a better trade-off, but merely extends the Pareto front towards higher compute regimes. We also observe a very large gap due to how the multitask network is initialized. Despite the multitask learning objective, it is very beneficial to initialize the network, even using the parameters of the most recent previous task. While there is no best initialization across all the compute budgets, i.e., no  initialization dominates all the others, we have found FT-d strategy to work better in the intermediate compute budgets. Further, in that setting, it is empirically best to co-train with the task the network is finetuned from (top-1), as opposed to co-train with other tasks as well (top-3). Hence, we use FT-d (top-1) as the MT strategy while reporting the metrics in the main paper.

On architectures that use BatchNorm~\cite{pmlr-v37-ioffe15}, we found it to be critical for performance to not update the BatchNorm statistics, running means and variances, with the batches of auxiliary tasks. Further, using a small fixed batch size of $64$ for all the auxiliary tasks resulted in a compute-efficient learner without hurting the performance as compared to a learner that uses large variable batch sizes for the auxiliary tasks (see eq.~\ref{eq:adaptive_bsz}).

%\clearpage

\section{BHPO Ablation} \label{app:bhpo}
\begin{figure}
  \centering
  \begin{subfigure}[t]{0.46\textwidth}
    \includegraphics[width=\textwidth, valign=t]{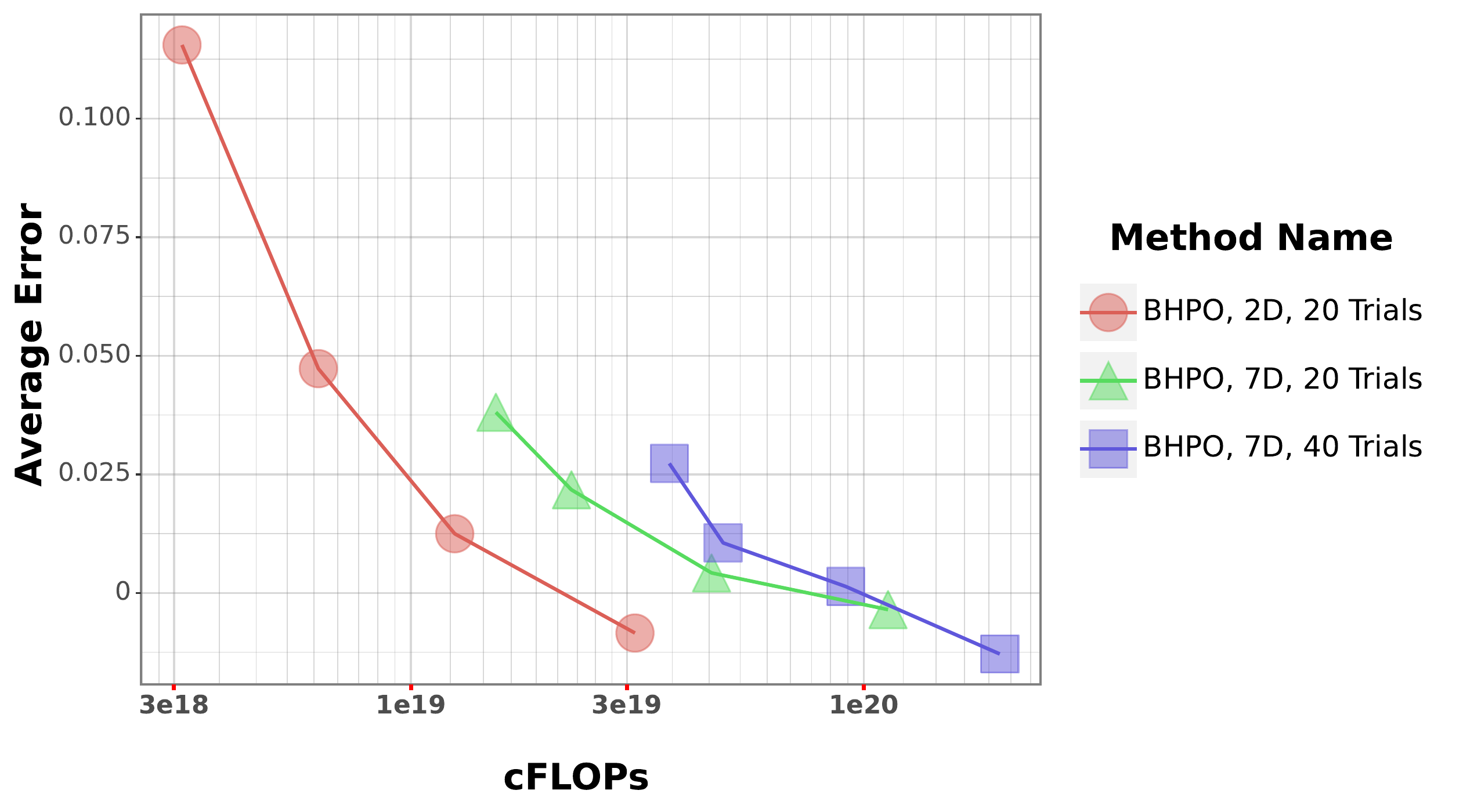}
    \caption{BHPO with different search spaces and trials}
    \label{fig:bhpo-search-space}
  \end{subfigure}
  ~
  \begin{subfigure}[t]{0.5\textwidth}
    \includegraphics[width=\textwidth, valign=t]{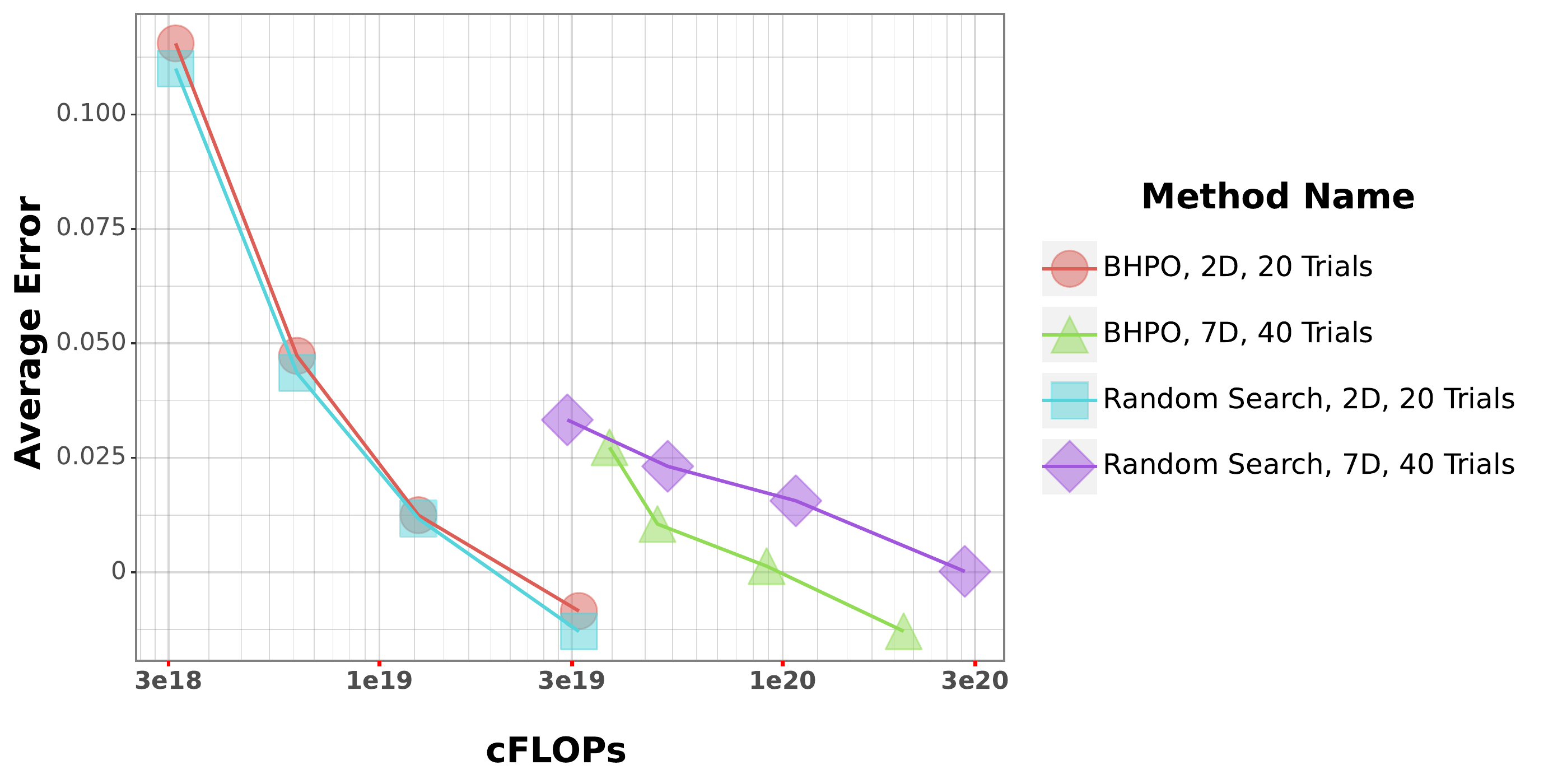}
    \caption{BHPO versus Random Search.}
    \label{fig:bhpo-alg}
  \end{subfigure}
  \caption{Pareto fronts varying the search space and search algorithm. Points along each line corresponds to 5K, 10K, 20K and 50K training steps per trial.
  }\label{fig:bhpo}
\end{figure}
In this section we study the impact of the choice of the search space and optimization algorithm (BHPO versus random search). Using Indep as a case study, we compare two search spaces, a small search space with only two hyper-parameters (learning rate and label smoothing as in the default setting) and a larger space with $7$ hyper-parameters which include also how data is augmented, the network architecture, etc. We also vary the number of training steps per trial (5K, 10K, 20K, 50K) and the number of trials (20, 40). As shown in Fig.~\ref{fig:bhpo-search-space} a large search space allows one to find a smaller error rate given the same training steps. This is mostly evident with a small number of training steps when the training has not converged yet. With a larger number of steps per training, the advantage diminishes. Also, it becomes harder the find the optimal hyper-parameter setting in a larges space given the same number of trials. Moreover, as the large search space includes optimizing the batch size, the computation cost is much higher. In Fig.~\ref{fig:bhpo-alg} we compare random search with the more sophisticated BHPO method. The latter finds a better setting in most cases in the large search space where the optimization is more difficult, but there is no clear difference in the easier 2-parameter space.

%\clearpage
\section{Ensembling} \label{app:ensembling}
In this section, we start by observing that during random hyper-parameter search we perform $N$ trials in parallel, one for each particular configuration of hyper-parameters. Since a hyper-parameter search often yields a set of accurate and diverse models, it seems wasteful to retain only the model that performed the best on the validation set. Instead, we can create an ensemble from these already trained models~\cite{Dietterich2000EnsembleMI}. 

\begin{figure}[t]
  \centering
  \includegraphics[width=.6\textwidth]{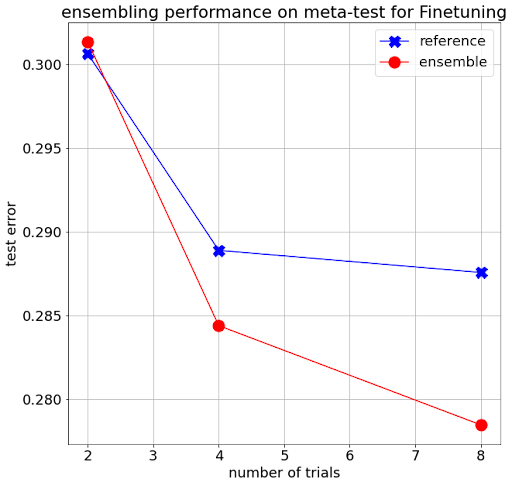}
  \caption{Error rate of ensembling as a function of the number of components. The reference is FT-d using only the best model found during random hyper-parameter search.}
  \label{fig:ensembling}
\end{figure}

Fig.~\ref{fig:ensembling} shows that we can significantly lower the error rate by ensembling, or for the same error rate we can drastically reduce the compute. For instance, by ensembling networks trained on $4$ trials only, we can attain lower error rate at half of the training compute of a (single component) baseline trained with $8$ trials.

In our experiments we have found that the best ensembling approach uses a weighted sum of probability distributions. The weights are the output of a softmax with temperature equal to $0.1$ using as input the top-1 accuracy obtained during cross-validation. This has the effect of weighting more the top performing models.

Notice however that while ensembling reduces the training compute at a given level of error rate, it increases linearly the cost at inference time. 

\section{Experimental protocol for transfer matrix computation}
\label{app:transfer}

\jb{
Figure \ref{fig:stats_domain} (right) shows the transfer matrix for the tasks in the \minerva\, stream. 
Each cell $(i, j)$ displays the test set accuracy for task $i$ after finetuning on task $i$ after training on task $j$ relative to independently training on task $i$ alone. 
To compute the transfer matrix we first pick the best training run from the standard hyperparameter-sweep 
for the {\em Independent} learner on each task. These runs form the reference accuracy 
for transfer to task $j$, and provide the starting point for finetuning. 
For finetuning we take each such run $i$ and finetune it on all tasks $j,\, \forall\; j > i$ with the standard hyperparameter 
sweep for the {\em finetuning} learner. In total we thus performed more than 90,000 training 
runs to compute the transfer matrix ($106 + \frac{106 \cdot 105}{2}$ learner runs with 16 random hyperparameters each).
}

\section{Measuring Compute}
\label{app:measuring_compute}
For the \minerva{} evaluation, we had the key goal of enabling a fair comparison between approaches that {\it efficiently} achieve strong results - such as via knowledge re-use or
carefully applied meta-learning strategies - versus methods that always re-train from scratch over all data, and perform large hyper parameter sweeps to obtain good results.
We explicitly wanted to choose a metric that would favor efficient learners that can scale gracefully as the data and the number of tasks increase.

Some options that we considered include 1) Counting the total number of training examples used (including multiplicity) 2) counting the total number of optimization steps 3)
Estimating the time elapsed during training. Ultimately we found problems with all of these approaches (and other similar approaches). A common issue is that we felt the
measurement should not limit in any way the solution space available to those implementing learners. Counting optimization steps may disadvantage learners that take many very small steps,
counting the number of examples accessed has the the problem that, in-reality, accessing data that already exists is a relatively inexpensive operation - and it's easy to imagine
that this could penalize learners that make use of replay buffers or caches that ultimately increase learner efficiency.

Counting floating point operations also suffers from challenges - notably since most hardware accelerator devices have a fixed maximum {\it throughput} of floating point operations per second.
Achieving this maximum throughput is key to training learners quickly, and making best use of available resources. Learners that use very sparse compute sparingly may ultimately
take a lot longer (in wall clock time) to train than learners that can efficiently use the dense compute available in modern accelerators (such as dense matrix multiplications).
Extending this further, learners that can efficiently be trained in parallel can require far less wall-clock time to train compared to learners that are inherently sequential.
Many real-world applications would prefer learners that make use of large amounts of compute if they are highly parallelizable.
Furthermore, the actual number of floating point operations performed is dependent on the underlying hardware platform (for example due to padding vectorized computations), and the levels of optimization performed on the linear algebra primitives themselves.

When considering our approach, we ultimately kept in mind the following core constraints,
1) the approach should not penalize or limit any particular learner implementation,
2) the approach should reasonably approximate cost of training the learner in a way that maps to reality,
3) storage and memory are typically cheaper than overall numerical compute, and so can be ignored without affecting the ordering of results,
4) it should be practically feasible for users of the benchmark to compute comparable values for the resources used, independently of the hardware they have available.
In these respects, we feel that the cumulative floating point operations used offers a fair compromise.
\jb{We obtained the FLOP counts reported in this study with the {\em cost analysis} API provided by the JAX deep learning framework when targeting execution on a CPU. The counts therefore take attained compiler optimizations into account, and include only minimal overhead due to padding for wide SIMD hardware. We confirmed that the reported counts closely match manually estimated counts for various model architectures and image sizes.} 

\clearpage

%%%%%%%%%%%%%%%%%%%%%%%%%%%%%%%%%%%%%%%%%%%%%%%%%%%%%%%%%%%%%%%%%%%%%%%%%%%%%%%%%%%%%%%%%%%%%%%%%%%%%%%%%%%%%%%%%%%%%%%%%
%% Full List 

\section{List of Datasets in \minerva} \label{app:ds_list}
\art{The following tables list all the datasets included in our benchmark, in the order of use. For each dataset, we indicate the year in which it appeared in our sampling procedure, the sampled paper that uses it, the task type(classification: C or multilabel classification: M)  , the image domain, the number of samples in the dataset (size) and the average input resolution. }

{
\footnotesize

\begin{longtable}{l p{4cm} p{3.5cm} p{0.8cm} p{1.1cm} p{0.9cm} p{1.5cm}}
% \caption{List of \minerva{} tasks}
\toprule
Year & Dataset Name & Sampled Paper & Type & Domain & Size  & Avg. res. \\
\midrule
\endhead
1992 & Aberdeen face database. \cite{aberdeen_face_database._introducing} & \cite{Craw1992FaceRB} & C & face & 468 & (519, 417) \\
1992 & Magellan Venus Volcanoes \cite{magellan_venus_volcanoes_introducing},\cite{Dua:2019} & \cite{1641019} & C & satellite & 102 & (1024, 1024) \\
1998 & Brodatz \cite{brodatz1966textures},\cite{brodatz_introducing} & \cite{BMVC.12.27} & C & texture & 672 & (213, 213) \\
1998 & LandSat UCI repo \cite{landsat_uci_repo_introducing},\cite{Dua:2019} & \cite{jones1998genetic} & C & satellite & 3764 & (3, 3) \\
1998 & Olivetti Face Dataset \cite{samaria1994parameterisation},\cite{olivetti_face_dataset_introducing} & \cite{Hall1998IncrementalEF} & C & face & 288 & (80, 70) \\
2000 & COIL 20 \cite{Nene1996},\cite{coil_20_introducing} & \cite{Matas2000ObjectRU} & C & object & 973 & (128, 128) \\
2001 & MPEG-7 \cite{mpeg-7_introducing} & \cite{12509cdafa7b4a0abb979d53664cb463} & C & shape & 943 & (341, 388) \\
2001 & COIL 100 \cite{Nayar1996ColumbiaOI},\cite{coil_100_introducing} & \cite{12509cdafa7b4a0abb979d53664cb463} & C & object & 6120 & (128, 128) \\
2004 & Butterfly dataset \cite{lazebnik:inria-00548542},\cite{butterfly_dataset_introducing} & \cite{lazebnik:inria-00548542} & C & object & 460 & (335, 431) \\
2004 & MNIST \cite{lecun1998gradient},\cite{mnist_introducing} & \cite{brown2004non} & C & ocr & 51000 & (28, 28) \\
2005 & UMIST \cite{graham1998characterising},\cite{umist_introducing} & \cite{kong2005discriminant} & C & face & 686 & (217, 199) \\
2005 & CMU AMP expression \cite{cmu_amp_expression_introducing} & \cite{kong2005discriminant} & C & face & 650 & (64, 64) \\
2005 & Caltech 101 \cite{li_andreeto_ranzato_perona_2022},\cite{caltech_101_introducing} & \cite{bart2005single} & C & object & 2601 & (244, 301) \\
2006 & Caltech Categories \cite{fergus2003object},\cite{caltech_categories_introducing} & \cite{gill2006single},\cite{hegerath2006patch} & C & object & 996 & (341, 514) \\
2006 & ALOI \cite{geusebroek2005amsterdam},\cite{aloi_introducing} & \cite{geusebroek2006compact} & C & object & 71494 & (144, 192) \\
2006 & UIUC cars \cite{agarwal2004learning},\cite{uiuc_cars_introducing} & \cite{gill2006single} & C & object & 823 & (50, 112) \\
2001 & COIL 100 \cite{Nayar1996ColumbiaOI},\cite{coil_100_introducing} & \cite{12509cdafa7b4a0abb979d53664cb463} & C & object & 6120 & (128, 128) \\
2004 & MNIST \cite{lecun1998gradient},\cite{mnist_introducing} & \cite{brown2004non} & C & ocr & 51000 & (28, 28) \\
2006 & Caltech Categories \cite{fergus2003object},\cite{caltech_categories_introducing} & \cite{gill2006single},\cite{hegerath2006patch} & C & object & 996 & (341, 514) \\
2006 & UIUC cars \cite{agarwal2004learning},\cite{uiuc_cars_introducing} & \cite{gill2006single} & C & object & 823 & (50, 112) \\
2008 & Graz-02 \cite{ig02},\cite{graz-02_introducing},\cite{marszalek2007accurate} & \cite{BMVC.24.47:abbreviated} & C & object & 747 & (497, 622) \\
2010 & Pascal 2007 \cite{pascal_2007_introducing} & \cite{BMVC.24.26:abbreviated} & M & object & 2501 & (382, 471) \\
2010 & PPMI \cite{yao2010grouplet},\cite{ppmi_introducing} & \cite{delaitre2010recognizing} & C & object & 2023 & (258, 258) \\
2010 & Olivetti Face Dataset \cite{olivetti_face_dataset_introducing} & \cite{BMVC.24.5:abbreviated} & C & face & 288 & (80, 70) \\
2011 & Caltech 256 \cite{griffin_holub_perona_2022},\cite{caltech_256_introducing} & \cite{kim2011hierarchical} & C & object & 20696 & (325, 371) \\
2011 & Oxford Flowers \cite{nilsback2006visual},\cite{oxford_flowers_introducing} & \cite{yang2011learning} & C & object & 680 & (555, 583) \\
2011 & Flicker Material Dataset \cite{Sharan-JoV-14},\cite{flicker_material_dataset_introducing} & \cite{hu2011toward} & C & texture & 676 & (384, 512) \\
2011 & LFW \cite{LFWTech},\cite{lfw_introducing} & \cite{kan2011side} & C & face & 11248 & (250, 250) \\
2011 & German Traffic Sign Recognition Benchmark \cite{Stallkamp2012},\cite{german_traffic_sign_recognition_benchmark_introducing} & \cite{timofte2011sparse} & C & object & 33392 & (50, 50) \\
2011 & VisTex \cite{vistex_introducing} & \cite{dahl2011learning} & C & texture & 125 & (512, 512) \\
2011 & Belgium Traffic Sign Dataset \cite{5403121},\cite{belgium_traffic_sign_dataset_introducing},\cite{mathias2013traffic},\cite{timofte2014multi} & \cite{timofte2011sparse} & C & object & 3887 & (118, 105) \\
2011 & Oxford Flowers 102 \cite{nilsback2008automated},\cite{oxford_flowers_102_introducing} & \cite{awais2011augmented} & C & object & 1020 & (534, 630) \\
2011 & Brodatz \cite{brodatz_introducing} & \cite{dahl2011learning} & C & texture & 672 & (213, 213) \\
2011 & ImageNet \cite{5206848},\cite{imagenet_introducing} & \cite{kim2011hierarchical} & C & object & 1281167 & (406, 473) \\
2012 & KTH-TIPS2-b \cite{kth-tips2-b_introducing} & \cite{timofte2012training} & C & texture & 3191 & (199, 198) \\
2012 & UMD \cite{xu2006projective},\cite{umd_introducing},\cite{xu2009viewpoint},\cite{xu2009combining},\cite{xu2012scale} & \cite{timofte2012training} & C & texture & 676 & (960, 1280) \\
2012 & CVC-MUSCIMA \cite{fornes2012cvc},\cite{cvc-muscima_introducing} & \cite{timofte2012training} & C & ocr & 419 & (1848, 3465) \\
2012 & KTH-TIPS \cite{kth-tips_introducing} & \cite{timofte2012training} & C & texture & 554 & (196, 200) \\
2012 & UIUC texture \cite{lazebnik2005sparse},\cite{uiuc_texture_introducing} & \cite{timofte2012training} & C & texture & 676 & (480, 640) \\
2013 & NORB \cite{lecun2004learning},\cite{norb_introducing} & \cite{wu2013scale} & C & object & 20655 & (96, 96) \\
2013 & KTH-TIPS2-a \cite{kth-tips2-a_introducing} & \cite{qi2013multi} & C & texture & 3097 & (200, 198) \\
2013 & IAPRTC-12 \cite{ESCALANTE2010419},\cite{iaprtc-12_introducing} & \cite{verma2013exploring} & M & object & 10830 & (395, 439) \\
2013 & sketch dataset \cite{eitz2012hdhso},\cite{sketch_dataset_introducing} & \cite{li2013sketch} & C & object & 13536 & (1111, 1111) \\
2014 & Wikipaintings \cite{artgan2018},\cite{wikipaintings_introducing} & \cite{Karayev_2014} & C & object & 48576 & (223, 221) \\
2014 & Pascal 2012 \cite{pascal_2012_introducing} & \cite{https://doi.org/10.48550/arxiv.1405.3531} & C & object & 5717 & (386, 470) \\
2015 & MNIST \cite{lecun1998gradient},\cite{mnist_introducing} & \cite{srinivas2015data} & C & ocr & 51000 & (28, 28) \\
2015 & SUN 397 \cite{xiao2010sun},\cite{sun_397_introducing},\cite{xiao2016sun} & \cite{li2015dictionary} & M & scene & 76128 & (291, 353) \\
2015 & CIFAR 10 \cite{krizhevsky2009learning},\cite{cifar_10_introducing} & \cite{he2016deep} & C & object & 42500 & (32, 32) \\
2015 & MIT Scenes \cite{mit_scenes_introducing} & \cite{Li_2015_CVPR} & C & scene & 4554 & (413, 501) \\
2016 & Chars74K \cite{deCampos09},\cite{chars74k_introducing} & \cite{aljundi2017expert} & C & ocr & 45740 & (158, 168) \\
2016 & DTD \cite{cimpoi14describing},\cite{dtd_introducing} & \cite{shahriari2016learning} & C & texture & 1880 & (451, 496) \\
2016 & Stanford Dogs \cite{KhoslaYaoJayadevaprakashFeiFei_FGVC2011},\cite{stanford_dogs_introducing} & \cite{kobayashi2016learning} & C & object & 10200 & (385, 442) \\
2016 & SVHN \cite{netzer2011reading},\cite{svhn_introducing} & \cite{aljundi2017expert} & C & object & 62268 & (32, 32) \\
2016 & VOC Actions \cite{everingham2010pascal},\cite{voc_actions_introducing} & \cite{aljundi2017expert} & M & object & 4234 & (226, 150) \\
2016 & ANIMAL \cite{bai2009integrating},\cite{animal_introducing} & \cite{li2016learning} & C & shape & 1346 & (474, 581) \\
2016 & Interact \cite{ Antol2014},\cite{interact_introducing} & \cite{sharmanska2016learning} & C & object & 2090 & (475, 582) \\
2016 & MS COCO \cite{https://doi.org/10.48550/arxiv.1405.0312},\cite{ms_coco_introducing} & \cite{zhao2016regional} & C & object & 82783 & (130, 127) \\
2016 & Stanford Cars \cite{KrauseStarkDengFei-Fei_3DRR2013},\cite{stanford_cars_introducing} & \cite{moghimi2016boosted} & C & object & 6937 & (483, 700) \\
2016 & CUB 200 \cite{WahCUB_200_2011},\cite{cub_200_introducing} & \cite{moghimi2016boosted} & C & object & 5094 & (386, 467) \\
2016 & Oxford IIIT Pets \cite{parkhi12a},\cite{oxford_iiit_pets_introducing} & \cite{kobayashi2016learning} & C & object & 3128 & (391, 437) \\
2016 & Caltech 101 \cite{li_andreeto_ranzato_perona_2022},\cite{caltech_101_introducing} & \cite{guo2016bag} & C & object & 2601 & (244, 301) \\
2016 & FGVC Aircraft \cite{maji13fine-grained},\cite{fgvc_aircraft_introducing} & \cite{moghimi2016boosted} & C & object & 5683 & (747, 1099) \\
2016 & Caltech 101 Silhouettes \cite{marlin2010inductive},\cite{caltech_101_silhouettes_introducing} & \cite{li2016learning} & C & shape & 4100 & (28, 28) \\2017 & SUN Attribute \cite{Patterson2012SunAttributes},\cite{sun_attribute_introducing},\cite{patterson2014sun} & \cite{kodirov2017semantic} & M & scene & 9692 & (478, 580) \\
2017 & CelebA \cite{liu2015faceattributes},\cite{celeba_introducing} & \cite{kalayeh2017improving} & C & face & 162770 & (218, 178) \\
2017 & CIFAR 100 \cite{Krizhevsky09learningmultiple},\cite{cifar_100_introducing} & \cite{rebuffi2017icarl} & C & object & 42500 & (32, 32) \\
2017 & LFWA \cite{Taigman2009MultipleOF},\cite{lfwa_introducing} & \cite{kalayeh2017improving} & C & face & 4716 & (250, 250) \\
2018 & TID2008 \cite{ponomarenko2009tid2008},\cite{tid2008_introducing} & \cite{lin2018self} & C & quality & 1149 & (384, 512) \\
2018 & TID2013 \cite{ponomarenko2015image},\cite{tid2013_introducing} & \cite{lin2018self} & C & quality & 1973 & (384, 512) \\
2018 & USPS \cite{hull1994database},\cite{usps_introducing} & \cite{liu2018detach} & C & ocr & 6207 & (16, 16) \\
2018 & Semeion \cite{doi:10.3109/10826089809115875},\cite{semeion_introducing},\cite{Dua:2019} & \cite{liu2018detach} & C & ocr & 1074 & (16, 16) \\
2018 & Office Caltech \cite{gong2012geodesic},\cite{office_caltech_introducing} & \cite{mancini2018boosting} & C & object & 1410 & (360, 373) \\
2018 & PACS \cite{Li2017dg},\cite{pacs_introducing} & \cite{mancini2018boosting} & C & object & 6062 & (227, 227) \\
2018 & MNIST-m \cite{ganin2015unsupervised},\cite{mnist-m_introducing} & \cite{mancini2018boosting} & C & ocr & 46111 & (32, 32) \\
2018 & ISBI-ISIC 2017 melanoma classification challenge \cite{codella2018skin},\cite{isbi-isic_2017_melanoma_classification_challenge_introducing} & \cite{radhakrishnan2018patchnet} & C & medical & 2000 & (2228, 3281) \\
2018 & CASIA-HWDB1.1 \cite{liu2011casia},\cite{casia-hwdb1.1_introducing} & \cite{jayaraman2018quadtree} & C & ocr & 797600 & (81, 70) \\
2018 & Caltech Camera Traps \cite{DBLP:conf/eccv/BeeryHP18},\cite{caltech_camera_traps_introducing} & \cite{beery2018recognition} & C & object & 13553 & (748, 1024) \\
2018 & EMNIST Balanced \cite{cohen_afshar_tapson_schaik_2017},\cite{emnist_balanced_introducing} & \cite{jayaraman2018quadtree} & C & ocr & 95880 & (28, 28) \\
2019 & Food 101 \cite{bossard14},\cite{food_101_introducing} & \cite{tan2019mixconv} & C & object & 75750 & (475, 495) \\
2019 & Food 101n \cite{lee2017cleannet},\cite{food_101_n_introducing} & \cite{8953770} & C & object & 45032 & (361, 394) \\
2019 & Trancos \cite{TRANCOSdataset_IbPRIA2015},\cite{trancos_introducing} & \cite{hossain2019one} & C & counting & 403 & (480, 640) \\
2019 & UCSD dataset \cite{chan2008privacy},\cite{ucsd_dataset_introducing},\cite{chan2008modeling},\cite{chan2009analysis} & \cite{hossain2019one} & C & counting & 800 & (158, 238) \\
2019 & AWA2 \cite{xian2018zero},\cite{awa2_introducing} & \cite{elhoseiny2019creativity} & M & object & 25827 & (192, 245) \\
2019 & Mall dataset \cite{chen2012feature},\cite{mall_dataset_introducing},\cite{change2013semi},\cite{chen2013cumulative},\cite{loy2013crowd} & \cite{hossain2019one} & C & counting & 680 & (480, 640) \\
2019 & NotMNIST \cite{bulatov2011notmnist},\cite{notmnist_introducing} & \cite{shafaei2019less} & C & ocr & 12345 & (28, 28) \\
2019 & STL10 \cite{coates2011analysis},\cite{stl10_introducing} & \cite{shafaei2019less} & C & object & 4250 & (96, 96) \\
2019 & MNIST-rot \cite{larochelle2007empirical},\cite{mnist-rot_introducing} & \cite{murugan2019so} & C & ocr & 51104 & (28, 28) \\
2019 & Fashion MNIST \cite{xiao2017/online},\cite{fashion_mnist_introducing} & \cite{shafaei2019less} & C & object & 51000 & (28, 28) \\
2019 & 15 Scenes \cite{lazebnik2006beyond},\cite{15_scenes_introducing} & \cite{jiang2019semi} & C & scene & 3021 & (244, 273) \\
2019 & COIL 100 \cite{sameer1996columbiacoil100},\cite{coil_100_introducing} & \cite{zhang2019self} & C & object & 6120 & (128, 128) \\
2019 & COIL 20 \cite{sameer1996columbia},\cite{coil_20_introducing} & \cite{zhang2019self} & C & object & 973 & (128, 128) \\
2019 & Tiny Imagenet \cite{tinyimagenet},\cite{tiny_imagenet_introducing} & \cite{shafaei2019less} & C & object & 85099 & (64, 64) \\
2020 & AnimalWeb \cite{khan2020animalweb},\cite{animalweb_introducing} & \cite{khan2020animalweb} & C & object & 12909 & (1154, 1461) \\
2020 & BIWI \cite{fanelli_IJCV},\cite{biwi_introducing} & \cite{pan2020self} & M & face & 11325 & (480, 640) \\
2020 & ShanghaiTech \cite{zhang2016single},\cite{shanghaitech_introducing} & \cite{duansofa} & C & counting & 595 & (696, 961) \\
2021 & PatchCamelyon \cite{Veeling2018-qh},\cite{patchcamelyon_introducing} & \cite{yang2021deep} & C & medical & 262144 & (96, 96) \\
2021 & DomainNet-Real \cite{peng2019moment},\cite{domainnet-real_introducing} & \cite{peng2019moment} & C & object & 102770 & (467, 472) \\
2021 & Path MNIST \cite{kather_2016_53169},\cite{path_mnist_introducing} & \cite{kothawade2022clinical} & C & medical & 3356 & (28, 28) \\
2021 & Pneumonia Chest X-ray \cite{kermany2018identifying},\cite{pneumonia_chest_x-ray_introducing} & \cite{kothawade2022clinical} & C & xray & 5216 & (970, 1327) \\
2021 & Oxford Flowers 102 \cite{nilsback2008automated},\cite{oxford_flowers_102_introducing} & \cite{pi2021searching} & C & object & 1020 & (534, 630) \\
2021 & Synthetic COVID-19 Chest X-ray Dataset \cite{zunair2021synthesis},\cite{synthetic_covid-19_chest_x-ray_dataset_introducing} & \cite{zunair2021synthesis} & C & xray & 14410 & (256, 256) \\
2021 & ImageNet \cite{5206848},\cite{imagenet_introducing} & \cite{DBLP:conf/bmvc/PiWLLY21} & C & object & 1281167 & (406, 473) \\
2021 & NIH Chest X-ray \cite{wang2017hospital},\cite{nih_chest_x-ray_introducing} & \cite{tetteh2021multi} & M & xray & 73638 & (1024, 1024) \\
2021 & covid-19 x-ray \cite{9144185},\cite{covid-19_x-ray_introducing},\cite{RAHMAN2021104319} & \cite{9892393} & C & xray & 14281 & (299, 299) \\
2021 & Tubercolosis \cite{9224622},\cite{tubercolosis_introducing} & \cite{chertieffect} & C & xray & 2809 & (512, 512) 
\end{longtable}

}

%%%%%%%%%%%%%%%%%%%%%%%%%%%%%%%%%%%%%%%%%%%%%%%%%%%%%%%%%%%%%%%%%%%%%%%%%%%%%%%%%%%%%%%%%%%%%%%%%%%%%%%%%%%%%%%%%%%%%%%%%
%% Short stream list 

\section{List of Datasets in the \SHORT version of \minerva} 
\label{app:ds_short_list}
\label{tab:short_datasetlist}
% \caption{List of \minerva{} \SHORT tasks}

{
\footnotesize
\begin{longtable}{l p{6cm} p{1cm} p{1.5cm} p{1.5cm} p{2.0cm}}
\toprule
Year & Dataset Name & Type & Domain & Size  & Avg. res. \\
\midrule
2004 & COIL 100 \cite{Nayar1996ColumbiaOI},\cite{coil_100_introducing} & C & object & 6120 & (128, 128) \\
2004 & MNIST \cite{lecun1998gradient},\cite{mnist_introducing} & C & ocr & 51000 & (28, 28) \\
2006 & Pascal 2005 \cite{inproceedings},\cite{pascal_2005_introducing} & C & object & 881 & (430, 553) \\
2006 & Caltech Categories \cite{fergus2003object},\cite{caltech_categories_introducing} & C & object & 996 & (341, 514) \\
2006 & UIUC cars \cite{agarwal2004learning},\cite{uiuc_cars_introducing} & C & object & 823 & (50, 112) \\
2009 & Pascal 2006 \cite{Everingham2006ThePV},\cite{pascal_2006_introducing} & C & object & 2211 & (420, 524) \\
2010 & Caltech 101 \cite{li_andreeto_ranzato_perona_2022},\cite{caltech_101_introducing} & C & object & 2601 & (244, 301) \\
2011 & Graz-02 \cite{ig02},\cite{graz-02_introducing},\cite{marszalek2007accurate} & C & object & 747 & (497, 622) \\
2011 & 15 Scenes \cite{lazebnik2006beyond},\cite{15_scenes_introducing} & C & scene & 3021 & (244, 273) \\
2011 & Pascal 2007 \cite{pascal_2007_introducing} & M & object & 2501 & (382, 471) \\
2011 & LFW \cite{LFWTech},\cite{lfw_introducing} & C & face & 11248 & (250, 250) \\
2013 & sketch dataset \cite{eitz2012hdhso},\cite{sketch_dataset_introducing} & C & object & 13536 & (1111, 1111) \\
2013 & Brodatz \cite{brodatz_introducing} & C & texture & 672 & (213, 213) \\
2014 & ImageNet \cite{5206848},\cite{imagenet_introducing} & C & object & 1281167 & (406, 473) \\
2014 & Pascal 2012 \cite{pascal_2012_introducing} & C & object & 5717 & (386, 470) \\
2014 & Caltech 256 \cite{griffin_holub_perona_2022},\cite{caltech_256_introducing} & C & object & 20696 & (325, 371) \\
2018 & CIFAR 100 \cite{Krizhevsky09learningmultiple},\cite{cifar_100_introducing} & C & object & 42500 & (32, 32) \\
2018 & CIFAR 10 \cite{krizhevsky2009learning},\cite{cifar_10_introducing} & C & object & 42500 & (32, 32) \\
2018 & USPS \cite{hull1994database},\cite{usps_introducing} & C & ocr & 6207 & (16, 16) \\
2018 & MNIST \cite{lecun1998gradient},\cite{mnist_introducing} & C & ocr & 51000 & (28, 28) \\
2018 & MNIST-m \cite{ganin2015unsupervised},\cite{mnist-m_introducing} & C & ocr & 46111 & (32, 32) \\
2018 & Office Caltech \cite{gong2012geodesic},\cite{office_caltech_introducing} & C & object & 1410 & (360, 373) \\
2018 & PACS \cite{Li2017dg},\cite{pacs_introducing} & C & object & 6062 & (227, 227) \\
2018 & ISBI-ISIC 2017 melanoma classification challenge \cite{codella2018skin},\cite{isbi-isic_2017_melanoma_classification_challenge_introducing} & C & medical & 2000 & (2228, 3281) \\
2019 & Fashion MNIST \cite{xiao2017/online},\cite{fashion_mnist_introducing} & C & object & 51000 & (28, 28) \\
2020 & Stanford Dogs \cite{KhoslaYaoJayadevaprakashFeiFei_FGVC2011},\cite{stanford_dogs_introducing} & C & object & 10200 & (385, 442) \\
2020 & CUB 200 \cite{WahCUB_200_2011},\cite{cub_200_introducing} & C & object & 5094 & (386, 467) \\
2020 & Stanford Cars \cite{KrauseStarkDengFei-Fei_3DRR2013},\cite{stanford_cars_introducing} & C & object & 6937 & (483, 700) \\
2020 & FGVC Aircraft \cite{maji13fine-grained},\cite{fgvc_aircraft_introducing} & C & object & 5683 & (747, 1099) \\
\end{longtable}
}

\newpage
\bibliography{nevis2022}

\end{document}